\pdfoutput=1

\documentclass[journal]{IEEEtran}

\usepackage{supertabular}

\usepackage{amsthm}
\usepackage{amsmath,amsfonts}
\usepackage{algorithmic}
\usepackage{algorithm}
\usepackage{array}
\usepackage{amssymb}
\usepackage{bbding}
\usepackage{booktabs}
\usepackage{colortbl}
\usepackage{cite}
\usepackage{float}
\usepackage{fontawesome}
\usepackage{graphicx}
\usepackage[pagebackref=false,breaklinks=true,colorlinks,citecolor=blue,linkcolor=blue,bookmarks=false]{hyperref}

\makeatletter
\def\@cite#1#2{\textcolor{blue}{[#1\if@tempswa , #2\fi]}}
\makeatother
\usepackage{capt-of}
\usepackage{multirow}
\usepackage{multicol}
\usepackage{stfloats}
\usepackage{subcaption}
\usepackage{tabularx}
\usepackage[table]{xcolor}
\usepackage{tikz}
\usepackage{textcomp}
\usepackage{pifont}
\usepackage{pgfplotstable}
\usepackage{url}
\usepackage{utfsym}
\usepackage{verbatim}
\usepackage{xcolor}
\usepackage{wasysym}
\usepackage{wrapfig}
\usepackage{caption}
\usepackage[export]{adjustbox}

\usepackage{arydshln} 

\newcolumntype{L}[1]{>{\raggedright\arraybackslash}p{#1}}
\newcommand{\ie}{\textit{i.e.},\@\xspace}

\makeatletter
\DeclareRobustCommand\onedot{\futurelet\@let@token\@onedot}
\def\@onedot{\ifx\@let@token.\else.\null\fi\xspace}

\def\eg{\emph{e.g}\onedot} 
\def\ie{\emph{i.e}\onedot} 
 
\def\etc{\emph{etc}\onedot} 
\def\wrt{w.r.t\onedot} 

\def\etal{\emph{et al}\onedot}

\makeatother

\usepackage[autostyle=true]{csquotes}

\usepackage[most]{tcolorbox}
\usepackage{xcolor}

\definecolor{TIFS_Navy}{RGB}{20, 75, 140}       
\definecolor{TIFS_Background}{RGB}{248, 250, 252} 
\definecolor{TIFS_Accent}{RGB}{0, 102, 51}    
\definecolor{TIFS_Warning}{RGB}{180, 0, 0}     

\definecolor{MyGray}{gray}{0.95}

\definecolor{Score10}{RGB}{0, 128, 0}         
\definecolor{Score89}{RGB}{100, 180, 0}       
\definecolor{Score57}{RGB}{255, 165, 0}       
\definecolor{Score24}{RGB}{220, 20, 60}       
\definecolor{Score1}{RGB}{139, 0, 0}          

\newtcolorbox{promptbox}[1]{
    enhanced,
    sharp corners,                
    boxrule=0.6pt,                
    colframe=black,           
    colback=MyGray,      
    colbacktitle=black,       
    coltitle=white,               
    fonttitle=\footnotesize\bfseries\rmfamily, 
    fontupper=\footnotesize\rmfamily,
    title={#1},
    toptitle=1.2mm,               
    bottomtitle=1.2mm,            
    lefttitle=8pt,                
    left=8pt, right=8pt,          
    top=8pt, bottom=8pt,          
    before skip=15pt, after skip=15pt
}

\newtheorem{theorem}{Theorem}
\newtheorem{definition}[theorem]{Definition}

\usepackage{xspace}

\definecolor{myblue}{rgb}{0.93, 0.95, 1.0}

\newcommand{\xmark}{\ding{55}}
\newcommand{\cmark}{\ding{51}}

\definecolor{cvprblue}{rgb}{0.21,0.49,0.74}
\definecolor{resnet50}{RGB}{15, 82, 186}
\definecolor{incv3}{RGB}{46, 139, 87}  
\definecolor{mobi}{RGB}{63, 72, 204}     
\definecolor{dense}{RGB}{30, 144, 255} 
\definecolor{conv}{RGB}{70, 130, 180}  
\definecolor{vit}{RGB}{220, 20, 60}    
\definecolor{pit}{RGB}{255, 140, 0}    
\definecolor{swin}{RGB}{205, 92, 92}   
\definecolor{deit}{RGB}{160, 82, 45}       
\definecolor{visformer}{RGB}{240, 128, 128}
\definecolor{Gray}{gray}{0.95}
\definecolor{TableGreen}{RGB}{155, 205, 124}

\definecolor{color_blue}{RGB}{135, 206, 235}
\definecolor{color_orig}{RGB}{255, 217, 178}
\definecolor{green1}{rgb}{0.013,0.545,0.341}
\definecolor{purple1}{rgb}{0.5,0,0.5}
\definecolor{yellow1}{rgb}{1,0.647,0}
\definecolor{Blue}{HTML}{377EB8}
\definecolor{Orange}{HTML}{FF7F00}
\definecolor{Green}{HTML}{4DAF4A}
\definecolor{Red}{HTML}{E41A1C}
\definecolor{Purple}{HTML}{984EA3}
\definecolor{Brown}{HTML}{A65628}
\definecolor{Pink}{HTML}{F781BF}

\definecolor{mibase}{HTML}{4DBBD5}  
\definecolor{mi300}{HTML}{00A087}
\definecolor{liboost}{HTML}{E64B35}

\definecolor{blueperturb}{HTML}{4C72B0}
\definecolor{greenperturb}{HTML}{55A868}

\begin{document}

\title{Boosting the Local Invariance for Better Adversarial Transferability}

\author{Bohan Liu, Xiaosen Wang
\thanks{Corresponding Author: Xiaosen Wang.}
\thanks{Bohan Liu is with the School of Computer Science and Technology, Xidian University, Xi'an 710126, China (e-mail: bhliuricardo@stu.xidian.edu.cn).}
\thanks{Xiaosen Wang is with the School of Computer Science and Technology, Huazhong University of Science and Technology, Wuhan 430074, China (e-mail: xswanghuster@gmail.com).}
\thanks{This article has supplementary downloadable material available at http://ieeexplore.ieee.org, provided by the authors. The material includes a PDF file containing additional experimental details and results. Contact xswanghuster@gmail.com for further
questions about this work.}
}
%

\markboth{IEEE TRANSACTIONS ON INFORMATION FORENSICS AND SECURITY}%
{Shell \MakeLowercase{\textit{et al.}}: A Sample Article Using IEEEtran.cls for IEEE Journals}

\maketitle
\begin{abstract}
Transfer-based attacks pose a significant threat to real-world applications by directly targeting victim models with adversarial examples generated on surrogate models. While numerous approaches have been proposed to enhance adversarial transferability, existing works often overlook the intrinsic relationship between adversarial perturbations and input images. In this work, we find that the adversarial perturbations often exhibit poor translation invariance for a given clean image and model, which is attributed to local invariance. Through empirical analysis, we demonstrate a positive correlation between the local invariance of adversarial perturbations \wrt the input image and their transferability across models. Based on this finding, we propose a general adversarial transferability boosting technique called the Local Invariance Boosting approach (LI-Boost). Extensive experiments on the standard ImageNet dataset demonstrate that LI-Boost significantly enhances five categories of transfer-based attacks, \ie, gradient-based, input transformation-based, model-related, advanced objective function, and ensemble attacks. The improvements hold not only on conventional CNNs, ViTs, and defense mechanisms, but also on real-world commercial vision API systems and vision-language models. Our approach provides a promising direction for future research on improving adversarial transferability across models. Our code is available
at \url{https://github.com/Trustworthy-AI-Group/TransferAttack}.
\end{abstract}

\vspace{-.2cm}
\section{Introduction}
\label{sec:intro}
\IEEEPARstart{D}{eep} Neural Networks (DNNs)~\cite{he2016deep,krizhevsky2012imagenet,vaswani2017attention} have achieved substantial success across various deep learning tasks, \eg, image recognition~\cite{2016inceptionv3,huang2017densenet,2021vit}, image generation~\cite{rombach2022high,ramesh2022hierarchical}, and large language model~\cite{mann2020language,touvron2023llama}, \etc. However, researchers have shown that DNNs are vulnerable to adversarial examples~\cite{szegedy2014intriguing,goodfellow2015FGSM}, in which small, often imperceptible perturbations can deceive the model into making incorrect predictions. This vulnerability poses a serious risk to real-world DNN-based applications, particularly in security-sensitive domains such as face verification~\cite{sharif2016accessorize} and autonomous driving~\cite{eykholt2018robust}. Consequently, adversarial attacks~\cite{goodfellow2015FGSM,moosavi2016deepfool,kurakin2017IFGSM,wang2019gan,wang2026devling} and defenses~\cite{madry2018towards,Ali2019AT,cohen2019RS,naseer2020NRP} have attracted extensive research interest. 
One of the intriguing characteristics of adversarial examples is their transferability across different models, where adversarial examples generated on a surrogate model can deceive previously unseen victim models~\cite{liu2017delving,dong2018mifgsm}. Unlike other attacks, transfer-based attacks do not necessitate access to the information of victim models, making them a particularly practical and serious threat to real-world DNN applications. Given these potential risks, extensive research has been conducted to enhance the transferability of adversarial attacks~\cite{Lin2020NesterovSim,wang2021vmi,wu2020sgm,wang2021fia,xie2019dim}.

Existing transfer-based attacks can be broadly categorized into five types~\cite{wang2026devling}: 1) \textbf{Gradient-based attacks}~\cite{dong2018mifgsm,Lin2020NesterovSim,wang2021vmi}, which typically incorporate various momentum techniques to stabilize the optimization process, thus improving convergence. 2) \textbf{Input transformation-based attacks}~\cite{xie2019dim,wang2021admix,wang2024bsr,wang2023rethinkingb}, which apply transformations to the input image to enhance the diversity of gradients for more effective optimization. 3) \textbf{Model-related attacks}~\cite{wu2020sgm,guo2020linbp,wang2023bpa}, which introduce model-specific modifications during the forward or backward propagation stages. 4) \textbf{Advanced objective functions}~\cite{wang2021fia,huang2019ila,Li2023ILPD,luo2025disrupting}, which design novel objective functions using the mid-layer features. 5) \textbf{Ensemble attacks}~\cite{liu2017delving,dong2018mifgsm,xiong2022stochastic,cao2025ViT}, which target multiple models simultaneously to increase adversarial transferability. Notably, these approaches directly optimize the perturbation \wrt the input image, without accounting for the inherent relationship between the perturbation and the input itself.

\begin{figure}[tb]
    \centering \includegraphics[width=.9\linewidth]{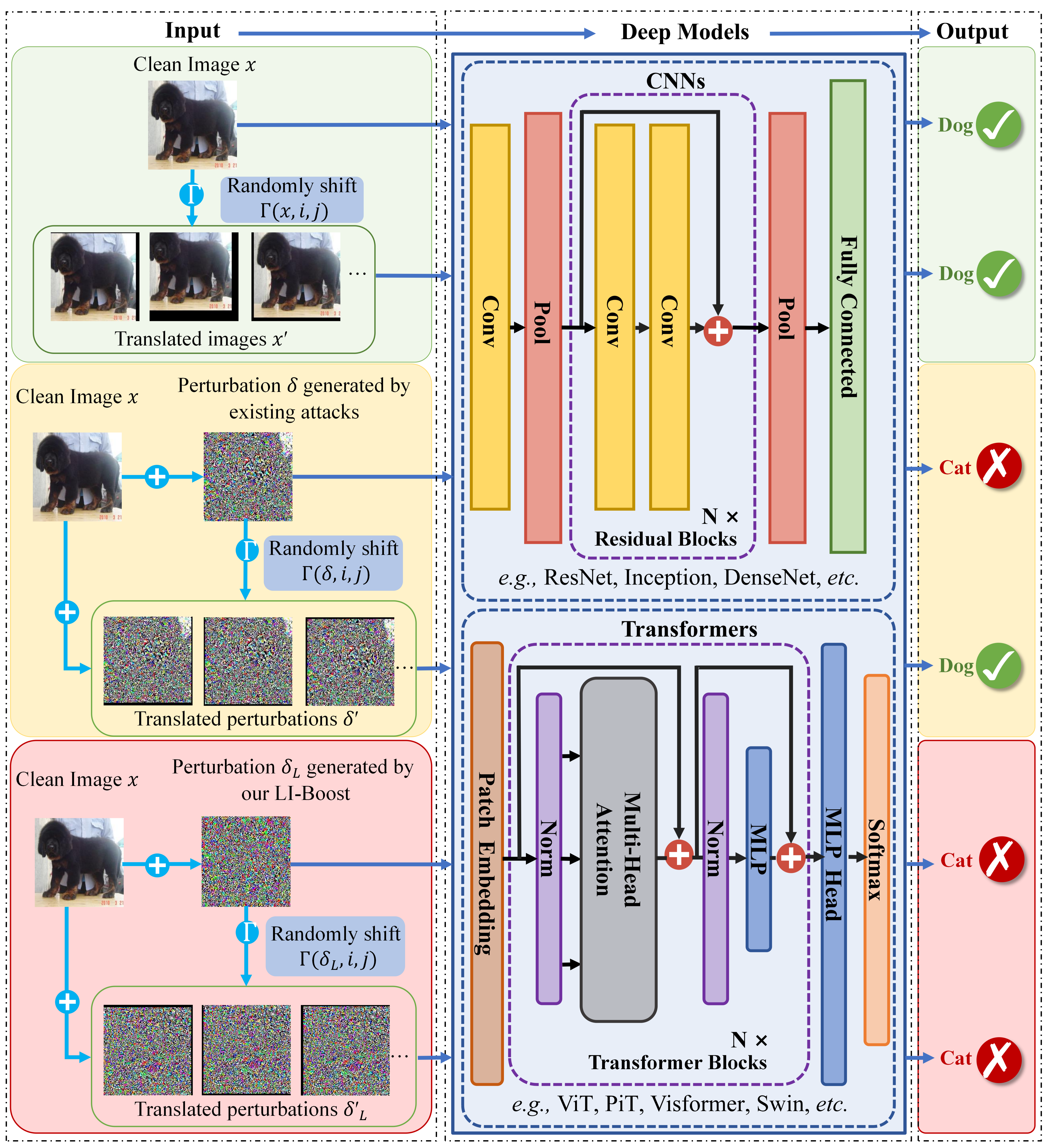}
    \caption{The impact of translation invariance of clean image and adversarial perturbation. The translated clean image can be correctly recognized by deep models (either CNNs or ViTs), whereas the slightly translated adversarial perturbations cannot consistently fool these models, and our LI-Boost enhanced perturbation can induce them to make incorrect predictions.}
\label{fig:translation_impact}
\vspace{-.7cm}
\end{figure}

It is widely known that clean images are consistently and accurately classified by various deep learning models and exhibit robust translation invariance on the same model. As shown in Fig.~\ref{fig:translation_impact}, however, we find that adversarial perturbations exhibit significantly weaker translation invariance for the same clean image and model. This observation is counterintuitive, given the inherent similarity among the local regions of clean images. We hypothesize and empirically validate that the local invariance of adversarial perturbation \wrt the clean image for a given model is positively correlated with its adversarial transferability across different models. Building on this insight, we introduce a novel and generalizable framework to enhance the transferability of various transfer-based attacks. Our contributions are summarized as follows:


\begin{enumerate}
    \item{We introduce the concept of local invariance for adversarial perturbations and reveal the underlying relationship between local invariance in the surrogate model and adversarial transferability across different models, providing new insights to enhance adversarial transferability across various models.}
    \item{We propose a novel and general boosting approach called LI-Boost to enhance adversarial transferability. Specifically, at each iteration, LI-Boost optimizes the adversarial perturbation using the gradient of adversarial examples with several translated perturbations to enhance the local invariance.}
    \item{Extensive experiments on the ImageNet dataset demonstrate that LI-Boost can effectively boost various types of transfer-based attacks on CNNs, ViTs, and defense mechanisms, as well as applications on commercial vision API systems, vision-language models, showing its generality and superiority in various scenarios.} 
\end{enumerate}

\vspace{-.23cm}
\section{Related Work}
\label{sec:formatting}
In this section, we provide a brief overview of existing adversarial attack and defense approaches.
\vspace{-.28cm}
\subsection{Adversarial Attacks}
After identifying the vulnerability of DNNs against adversarial examples~\cite{szegedy2014intriguing}, numerous adversarial attacks have emerged~\cite{moosavi2016deepfool,madry2018towards,wang2019gan}. White-box attacks~\cite{goodfellow2015FGSM,kurakin2017IFGSM,croce2020reliable}, which have full access to the target model (\eg, gradients, architectures, and logits), are widely used to assess the robustness of DNNs. In contrast, black-box attacks, which have limited access to the target model, pose more severe threats to real-world DNN-deployed applications. Black-box attacks can be further categorized into score-based attacks~\cite{uesato2018adversarial,guo2019simple,andriushchenko2020square}, decision-based attacks~\cite{li2020qeba,wang2022triangle,maho2021surfree} and transfer-based attacks~\cite{liu2017delving,wang2021vmi}. Among these, transfer-based attacks, where the adversarial examples generated on surrogate models are used to attack the target model without any direct access, have garnered significant research \mbox{interest}~\cite{xie2019dim,gao2020patch,zhang2023improving,wang2023diversifying,zhang2024bag,naseer2022improving,Li2023ILPD,zhang2022naa,Zhang2023VITtoken,zhang2024VirtualDense,zhu2025improving,zhu2025reinforcing}.

\textbf{Gradient-based attacks} are popular white-box attacks (\eg, FGSM~\cite{goodfellow2015FGSM}, I-FGSM~\cite{kurakin2017IFGSM}) that exhibit superior white-box attack performance but poor transferability. To boost adversarial transferability, various approaches integrate momentum to stabilize the optimization~\cite{Lin2020NesterovSim,qin2022boosting,wang2021boosting}. For instance, MI-FGSM~\cite{dong2018mifgsm} first integrates the momentum into I-FGSM and achieves much higher transferability. VMI-FGSM~\cite{wang2021vmi} further refines gradient variance to stabilize the update direction. PGN~\cite{ge2023boosting} introduces a penalized gradient norm to the original loss function, producing adversarial examples in flatter local regions with improved transferability across models.  MUMODIG~\cite{ren2025mumodig} improves transferability through generating integration paths using diverse baseline samples and enforcing the monotonicity of each path.

Numerous \textbf{input transformation-based attacks} have been proposed to boost adversarial transferability~\cite{zou2020improving,wang2021admix,wang2024bsr}. DIM~\cite{xie2019dim} improves transferability by randomly resizing and padding the input image before the gradient calculation. \textit{Admix}~\cite{wang2021admix} enhances diversity by combining the original image with a second image from a distinct category to generate more diverse perturbations. SIA~\cite{Wang2023SIA} applies various transformations to the blocks of the input image while maintaining its structural integrity. BSR~\cite{wang2024bsr} splits the image into blocks, then shuffles and randomly rotates them. SID~\cite{zhou2025sid} leverages spatial invariance by fusing global image into localized blocks through linear or frequency domains, followed by multi-scale downsampling and random positioning.

Additionally, \textbf{model-related attacks} modify the architecture of the surrogate model for enhanced transferability. For example, Linbp~\cite{guo2020linbp} modifies the backward propagation process by setting the gradient of the ReLU activation function to a constant value and scaling the gradients of residual blocks. SGM~\cite{wu2020sgm} prioritizes the gradients from skip connections over those from residual modules to improve transferability. BPA~\cite{wang2023bpa} introduces a non-monotonic function as the derivative of ReLU and integrates a temperature-controlled softmax function to activate the truncated gradient for better transferability. VDC~\cite{zhang2024VirtualDense} imports virtual dense connections for dense gradient back-propagation in attention maps and MLP blocks based on the forward propagation for vision transformers. FPR~\cite{ren2025fpr} refines the forward propagation by diversifying the attention map and accumulating the output token embedding using momentum.

\textbf{Advanced objective functions} often perturb mid-layer features to improve transferability~\cite{wang2023diversifying,zhang2022naa,zhang2022enhancing}. For instance, ILA~\cite{huang2019ila} enhances the similarity of feature differences between an adversarial example and its benign counterpart on a pre-specified layer of the source model. FIA~\cite{wang2021fia} disrupts object-aware features that significantly influence model decisions to calculate the aggregated gradients for updating the perturbation. ILPD~\cite{Li2023ILPD} amplifies the magnitude of perturbations in the adversarial direction within intermediate layers by incorporating perturbation decay in a single-stage optimization framework. BFA~\cite{WANG2024bfa} partitions image features into white-box and black-box components, generates a fitted image by perturbing the white-box features, and then computes the fitted gradient across images with varying degrees.

\textbf{Ensemble attacks} simultaneously generate adversarial examples on multiple surrogate models to enhance adversarial transferability. Dong \etal~\cite{dong2018mifgsm} aggregate the logits from all surrogate models to generate adversarial examples. SVRE~\cite{xiong2022stochastic} adopts stochastic variance to reduce gradient variance between various models. CWA~\cite{chen2024rethinkingensemble} identifies shared vulnerabilities across an ensemble of models to improve adversarial transferability. SMER~\cite{tang2024ensemble} stochastically selects a single surrogate model in each internal loop for mini-batch perturbation and employs a reinforcement learning framework to dynamically optimize ensemble weights.

Besides, with the development of vision-language models (VLMs), adversarial attacks on VLMs have also gained great research attention. AttackVLM~\cite{zhao2023attackvlm} evaluates the robustness of open-source VLMs through transfer-based attacks and query-based strategies. Dong \etal~\cite{dong2023ssacwa} combine SSA~\cite{long2022ssa} and CWA~\cite{chen2024rethinkingensemble} to improve transferability on commercial VLMs, \eg Google's Bard. M-Attack~\cite{li2025mattack} utilizes random cropping and resizing to enhance targeted adversarial transferability against closed-source VLMs. FOA-Attack~\cite{jia2025foa} incorporates local feature alignment and dynamically adjusts each model's weight, thereby yielding a substantial boost in adversarial transferability on VLMs. MPC-Attack~\cite{li2026mpc} employs multiple vision encoders with different paradigms, concatenates their feature representations, and jointly optimizes the perturbation across the combined feature space.
\vspace{-.15cm}
\subsection{Adversarial Defense}
Numerous defenses have been proposed to mitigate the threat of adversarial examples. Adversarial training~\cite{goodfellow2015FGSM,tramer2018ensemble,madry2018towards,Ali2019AT} incorporates adversarial examples into the training process, which has proven to be one of the most effective methods for improving the model's robustness. 
Fast-AT~\cite{Wong2020ATdefense} adopts a single iteration to generate adversarial examples for training, which can significantly boost adversarial robustness. Guo~\etal~\cite{guo2018countering} employed various image transformations (\eg, JPEG compression, etc.) to preprocess inputs before feeding them into the models. Liao~\etal~\cite{liao2018defense} propose the high-level representation guided denoiser (HGD) by minimizing the difference between the model's outputs on clean and denoised images. Naseer~\etal~\cite{naseer2020NRP} developed a Neural Representation Purifier (NRP) trained using a self-supervised adversarial training method to purify input images. Several certified defense methods offer verifiable defense capabilities, such as randomized smoothing (RS)~\cite{cohen2019RS}. Diffusion models for purification (DiffPure)~\cite{nie2022diffpure} exhibit excellent potential for adversarial defense. Besides, Xu~\etal~\cite{xu2026mimir} proposes a self-supervised AT method specific to ViTs, named MIMIR, utilizing masked image modeling and mutual information, which ranks top-1 on the standardised evaluation RobustBench~\cite{croce2020robustbench}.

\vspace{-.05cm}
\section{Methodology}
\vspace{-.1cm}
\subsection{Preliminaries}
Given a victim model $f$ with parameters $\theta$ and a clean image $x \in \mathcal{X}$ with ground-truth label $y$, where $x$ is in $d$ dimensions and $\mathcal{X}$ denotes all the legitimate images, adversarial attacks seek to identify an adversarial example $x+\delta$ $ \in \mathcal{X}$ such that:
\vspace{-.1cm}
\begin{equation}
    f(x;\theta) \neq f(x+\delta;\theta) \quad \text{ s.t. } \quad \|\delta\|_p \leq \epsilon.
\end{equation}
Here $\epsilon$ represents the perturbation budget, $\delta$ is the perturbation of $x$, and $\|\cdot\|_p$ is the $\ell_{p}$-norm distance. In this work, we adopt  $\ell_{\infty}$ distance to align with existing works. To generate such a perturbation, the adversary typically maximizes the loss function $J$ (\eg, cross-entropy loss) of the target model, which can be formalized as:
\begin{equation}
    \delta = \mathop{\arg\max}\limits_{\|\delta\|_p \leq \epsilon} J(x+\delta,y;\theta).
    \label{eq:goal}
\end{equation}
The transferability of adversarial examples generated on the surrogate model when applied to the victim model $f$ can be evaluated by the attack success rate (ASR) as follows:
\begin{equation}
    ASR = \frac{1}{|\mathcal{X}|}\sum_{x\in \mathcal{X}} \mathbb{I}[f(x) \neq f(x+\delta)],
    \vspace{-.1cm}
\end{equation}
where $\delta$ is generated on surrogate model $f_s$ w.r.t the input image $x$ and $\mathbb{I}(\cdot)$ is the indicator function.

\vspace{-.25cm}
\subsection{Motivation}
\label{motivation}
DNNs with different architectures often exhibit the ability to consistently recognize the same image, demonstrating the model-independent semantic consistency of clean images. In addition, DNNs are known for their strong translation invariance property~\cite{jaderberg2015spatial,kauderer2017quantifying}, wherein they reliably produce accurate predictions across translated versions of an image. This behavior mirrors the human visual system that the slightly translated images can still be correctly recognized, as translation does not fundamentally alter the images' semantic content.

Adversarial transferability refers to the ability of adversarial examples generated on the surrogate model to successfully deceive other models. This concept parallels the observation that clean images are often classified correctly by various models. However, existing adversarial examples often exhibit weak transferability across different models, particularly between CNNs and ViTs. Besides, as shown in Fig.~\ref{fig:translation_impact}, we observe that adversarial perturbations also exhibit poor translation invariance for a given clean image and DNN. This observation contradicts human intuition, which suggests that local regions of an image should retain consistent semantic features. In contrast, the corresponding adversarial perturbations vary significantly. For example, while the pixels of a dog’s ear are visually similar, the associated perturbations vary substantially. This indicates that the perturbations not only overfit the victim model but also become highly sensitive to pixel positions within the image.

This finding inspires us to hypothesize that translation invariance may be beneficial for enhancing the adversarial transferability. To validate this assumption, we first define the local invariance of adversarial perturbation $\delta$ to quantify translation invariance  mathematically as follows:
\vspace{-.1cm}
\begin{definition}[Local Invariance]
    Given an adversarial perturbation $\delta$ for the input image $x \in \mathcal{X}$ and surrogate model $f_s$, the local invariance of  perturbation is quantified as:
    \begin{gather*}
       \mathcal{I} (x,\delta, k) = \frac{\sum_{-k \leq{i,j}\leq k} \mathbb{I}\left[f_s(x) \neq f_s(x + \Gamma(\delta, i, j))\right]}{(2k+1)^2}, 
    \end{gather*}
    where $\Gamma(\delta, i, j)$ denotes the translation operator that translates $\delta$ by $i$ pixels horizontally and $j$ pixels vertically, and $k$ represents the upper bound of translated pixels. Boundary regions introduced are handled via zero-padding.
\end{definition}
Intuitively, the local invariance represents the expected probability that the adversarial perturbation retains its fooling ability under small spatial translation. To quantify this property across different methodologies, we calculated the average local invariance of adversarial perturbations generated by various transfer-based attacks. As shown in Fig.~\ref{fig:local_invariance_scatter}, we observe that the improved adversarial transferability is often associated with better local invariance. Based on this observation, we conclude that the local invariance of adversarial perturbations serves as an indicator of their transferability across different models. Furthermore, enhancing local invariance appears to positively influence the adversarial transferability through maximizing the expected attack success probability of adversarial examples within a small translated local neighborhood of the corresponding perturbation.

\subsection{Local Invariance Boosting Approach}%

Building on the above analysis, we propose a novel attack framework called Local Invariance Boosting approach (LI-Boost), which enhances the local invariance of adversarial perturbations w.r.t the clean image to improve transferability across various models. Specifically, we can formulate the problem as follows:
\begin{equation}
    \delta^* = \mathop{\arg\max}\limits_{\|\delta\|_p \leq \epsilon} \left[\min_{\delta' \in \mathcal{N}_k(\delta)} J(x+\delta', y; \theta)\right],
    \label{eq:obj}
\end{equation}
where $k$ represents the maximum number of pixels by which the perturbation is translated, and the local neighborhood for the perturbation $\delta$ is as in Eq.\ref{eq:sampledelta}:
\begin{equation}
\mathcal{N}_k(\delta)=\{\Gamma (\delta,i,j) \mid -k \leq i,j \leq k\},
\label{eq:sampledelta}
\end{equation}
\begin{figure}[tb]
    \centering
    \captionsetup[subfigure]{justification=raggedright, font=footnotesize, labelfont=normalfont,textfont=normalfont, singlelinecheck=false}
    \hspace{-.1cm}
    \begin{subfigure}[t]{0.47\linewidth}
        \centering
        \hspace{-.6cm}
        \begin{tikzpicture}[clip]
            \begin{axis}[
                width=1.3\linewidth,
                height=1.22\linewidth,
                grid=both,
                minor grid style={gray!25, dashed},
                major grid style={gray!25, dashed},
                xlabel={Local Invariance},
                ylabel={Attack success rates (\%)},
                ylabel style={font=\scriptsize, yshift=-18pt},
                xlabel style={font=\scriptsize, yshift=5pt},
                tick label style={font=\tiny},
                xmin=0.12, xmax=0.69,
                ymax=100,ymin=32
            ]
                \addplot[only marks, mark=*, magenta, opacity=0.8, mark size=1.2pt] coordinates {(0.25,35.6)}; 
                \addplot[only marks, mark=*, magenta, opacity=0.8, mark size=5.0pt] coordinates {(0.31,54.6)}; 
                \addplot[only marks, mark=*, magenta, opacity=0.8, mark size=5.0pt] coordinates {(0.48,80.3)}; 
                \addplot[only marks, mark=*, magenta, opacity=0.8, mark size=2.4pt] coordinates {(0.44,71.2)}; 
                \addplot[only marks, mark=*, cyan, opacity=0.9, mark size=1.2pt] coordinates {(0.32,51.4)};    
                \addplot[only marks, mark=*, cyan, opacity=0.9, mark size=4.0pt] coordinates {(0.35,59.9)};    
                \addplot[only marks, mark=*, cyan, opacity=0.9, mark size=5.0pt] coordinates {(0.47,77.9)};    
                \addplot[only marks, mark=*, cyan, opacity=0.9, mark size=5.0pt] coordinates {(0.51,83.8)};    
                \addplot[only marks, mark=*, cyan, opacity=0.9, mark size=5.0pt] coordinates {(0.625,88.6)};   
                
                \addplot[only marks, mark=*, green, opacity=0.6, mark size=1.2pt] coordinates {(0.48,36.5)};
                \addplot[only marks, mark=*, green, opacity=0.6, mark size=4.0pt] coordinates {(0.55,36.5)};
                \addplot[only marks, mark=*, green, opacity=0.6, mark size=5.0pt] coordinates {(0.62,36.5)};

                \node [above] at (axis cs: 0.195, 34.5) {{\tiny MI-FGSM}};
                \node [above] at (axis cs: 0.24, 55.9) {{\tiny VMI-FGSM}};
                \node [above] at (axis cs: 0.44, 81.3) {{\tiny PGN}};
                \node [above] at (axis cs: 0.36,69.2) {{\tiny MUMODIG}};
                \node [above] at (axis cs: 0.35, 45.8) {{\tiny DIM}};
                \node [above] at (axis cs: 0.39, 53.0) {{\tiny Admix}};
                \node [above] at (axis cs: 0.5, 70.2) {{\tiny SIA}};
                \node [above] at (axis cs: 0.54, 75.99) {{\tiny BSR}};
                \node [above] at (axis cs: 0.65, 80.25) {{\tiny SID}};
                \node [above] at (axis cs: 0.55, 42) {{\tiny {\# Propagations}}};
                \node [above] at (axis cs: 0.48, 38) {{\tiny 1}};
                \node [above] at (axis cs: 0.55, 38) {{\tiny 15}};
                \node [above] at (axis cs: 0.62, 38) {{\tiny 20}};
            \end{axis}
        \end{tikzpicture}
        \caption{Local invariance ($k = 5$) on ResNet-50 and the average attack success rates (\%) on nine normally trained models of various attacks.}
        \label{fig:local_invariance_scatter}
    \end{subfigure}
     \hspace{-.1cm}
    \begin{subfigure}[t]{0.47\linewidth}
        \centering
        \begin{tikzpicture}[clip]
            \begin{axis}[
                width=1.1\linewidth,
                height=1.2\linewidth,
                grid=both,
                minor grid style={gray!25, dashed},
                major grid style={gray!25, dashed},
                xlabel={Upper bound of translated pixels $k$},
                ylabel={Attack success rates (\%)},
                ylabel style={font=\scriptsize, yshift=-20pt},
                xlabel style={font=\scriptsize, yshift=5pt},
                tick label style={font=\tiny},
                xtick={0,1,2,3,4,5},
                ymin=20, ymax=100,
                axis y line*=left,
                legend style={at={(0., 0.62)}, anchor=west, align=center, font=\tiny, minimum width=0.02cm, column sep=0.05ex, inner sep=0.15ex, fill opacity=0.5}, 
                legend image post style={scale=0.5},
            ]
                \addplot[line width=0.7pt, solid, mark=star, color=magenta, mark options={mark size=1pt}] 
                    coordinates {(0,94.9) (1,93.7) (2,90.8) (3,90.6) (4,88.4) (5,87.1)};
                \addlegendentry{White-box}

                \addplot[line width=0.7pt, solid, mark=square*, color=cyan, mark options={mark size=1pt}] 
                    coordinates {(0,28.2) (1,35.8) (2,34.9) (3,36.6) (4,37.2) (5,37.3)};
                \addlegendentry{Black-box}

                \addplot[line width=0.7pt, dashed, mark=triangle*, color=red, mark options={mark size=1pt}] 
                    coordinates {(0,0) (1,0) (2,0) (3,0) (4,0) (5,0)};
                \addlegendentry{Computation cost}
            \end{axis}
            \begin{axis}[
                width=1.1\linewidth,
                height=1.2\linewidth,
                axis y line*=right, 
                axis x line=none, 
                ylabel={\# Propagations},
                ylabel style={font=\scriptsize, yshift=-135pt, rotate=180},
                tick label style={font=\tiny},
                ymin=0, ymax=125,
                xticklabels={,,},
            ]
                \addplot[line width=0.6pt, dashed, mark=triangle*, color=red, mark options={mark size=1pt}] 
                    coordinates {(0,1) (1,9) (2,25) (3,49) (4,81) (5,121)};
            \end{axis}            
        \end{tikzpicture}
        \caption{Attack success rates (\%) and number of forward and backward propagations of MI-FGSM with Eq.~\eqref{eq:obj} using various $k$.}
        \label{fig:mifgsm_k_performance_brute}
    \end{subfigure}
    \caption{The local invariance of various transfer-based attacks (left) and the attack performance of MI-FGSM with enhanced local invariance (right).}
    \vspace{-.6cm}
\end{figure}
To assess the effectiveness of this approach, we employ MI-FGSM to solve Eq.~\eqref{eq:obj} using various $k$. We choose \mbox{ResNet-50} as the white-box backbone and the other eight normally trained models illustrated in Sec.~\ref{sec:setup} as the black-box settings. As shown in Fig.~\ref{fig:mifgsm_k_performance_brute}, adversarial transferability consistently improves when enhancing the local invariance, providing empirical support for our hypothesis.

\begin{algorithm}[tb]
\caption{LI-Boost-MI}
\label{alg:algorithm}
\textbf{Input}: Victim model $f$ with the loss function $J$; a raw image $x$ with ground-truth label $y$; perturbation budget $\epsilon$; decay factor $\mu$; number of iterations $T$; number of \mbox{sampled} perturbations $N$;  upper bound of translated pixels $k$.\\
\textbf{Parameter}: $\alpha =\epsilon / T; g_0 = 0; \delta_0 = 0$.\\
\textbf{Output}: Perturbation $\delta$.
\begin{algorithmic}[1] 
    \FOR{$t = 1$ to $T$}
        \STATE Calculate the gradient $\bar{g_t}$ w.r.t $\delta_t$ using Eq.~\eqref{gradient_calculate},
        \STATE Update the momentum: $g_{t}=\mu\cdot g_{t-1} + \frac{\bar{g}_{t}}{||\bar{g}_{t}||_{\infty}}$,
        \STATE Update the adversarial perturbation:\\ $\delta_t = \mathrm{clamp}(\delta_{t-1} + \alpha \cdot \mathrm{sign} (g_t), -\epsilon, \epsilon)$,
    \ENDFOR
\STATE \textbf{return} $\delta = \delta_T$.
\end{algorithmic}
\end{algorithm}
While a min-max formulation could theoretically enforce robustness against the worst-case spatial shift, such an objective is computationally intractable to optimize in practice due to the continuous perturbation space. Besides, optimizing against a single worst-case does not guarantee generalization across the shift distribution, as the local optima in adversarial learning are known to generalize poorly~\cite{rice2020overfitting}. To mitigate this issue, we relax the min-max objective into an expected loss maximization task as in Eq.\ref{eq:exception}. This relaxation is well-motivated by the fact that a perturbation that performs well in expectation across diverse shifts generalizes more effectively to the transfer gap between surrogate and target models.  
\begin{equation}
    \delta^* = \mathop{\arg\max}\limits_{\|\delta\|_{p}\leq \epsilon} \mathbb{E}_{\delta' \in \mathcal{N}_k(\delta)} \left[J(x + \delta', y; \theta)\right],
    \label{eq:exception}
\end{equation}
However, computing the exact expectation is computationally prohibitive, as the required propagations scale quadratically with the increase of translation bound $k$ (\eg, from 9 at $k$ = 1 to 49 at $k$ = 3). This quadratic growth in the number of required gradient evaluations, $(2k+1)^2$, results in prohibitive training cost for large $k$, as shown in the runtime and backpropagation analysis in Tab.~\ref{tab:time_complexity}.


\begin{table}[htb]
\centering
\renewcommand{\arraystretch}{1}
\newcolumntype{Y}{>{\centering\arraybackslash}X}
\caption{Time consumption and computation complexity w.r.t. various \textit{k}. We report average runtime(seconds/image) and computational cost(Backpropagations/BP).} 
\label{tab:time_complexity}
\begin{tabularx}{\linewidth}{c *{7}{Y}} 
\toprule
$k$ & 0 & 1 & 2 & 3 & 4 & 5 & 6 \\ 
\midrule
\textbf{Time (s/img)} & 0.07 & 0.69 & 1.92 & 3.78 & 6.26 & 9.35 & 13.07 \\ 
\midrule
\textbf{BP (count)} & 1 & 9 & 25 & 49 & 81 & 121 & 169 \\ 
\bottomrule
\end{tabularx}
\end{table}

To enhance computational efficiency without sacrificing the attack effectiveness, we employ a Monte Carlo sampling strategy~\cite{1949monte} to approximate the gradient by randomly drawing multiple shifted perturbations in each update. Specifically, the gradient is estimated as:
\begin{equation}
\begin{aligned}
    \bar{g} &= \nabla_{\delta} \mathbb{E}_{\delta' \in \mathcal{N}_k(\delta)} [J(x + \delta', y; \theta)] \\
    &= \mathbb{E}_{\delta' \in \mathcal{N}_k(\delta)} \nabla_{\delta} [J(x + \delta', y; \theta)] \\
    &\approx \frac{1}{N} \sum_{n=1}^{N} \nabla_{\delta} J(x + \delta'_n, y; \theta),
\end{aligned}
\label{gradient_calculate}
\end{equation}
where $i, j$ in $\mathcal{N}_k(\delta)$ are randomly sampled from $[-k,k]$ with $k$ being a predefined parameter of the upper bound of translated pixels. $N$ denotes the total number of sampled perturbations. The selection of an appropriate $N$ is crucial for balancing the trade-off between attack efficiency and \mbox{effectiveness}. It is important to note that LI-Boost is a general boosting technique applicable to a variety of attacks. As an example, we incorporate LI-Boost into the MI-FGSM, denoted as LI-Boost-MI. The details are summarized in Algorithm~\ref{alg:algorithm} for clarity.

\begin{table*}[!t]
\centering
\captionsetup{labelsep=newline, justification=centering}
\caption{
 Attack success rates (\%) on nine normally trained models and nine defense mechanisms of various \textbf{gradient-based attacks} w/wo LI-Boost. The adversarial examples are crafted on RN-50. * indicates the white-box model. The best results are  \textbf{bold}, while methods incorporating LI-Boost are highlighted in \colorbox{gray!10}{gray}.
}
  \setlength{\tabcolsep}{1.5pt} 
\setlength{\extrarowheight}{1.2pt}
\resizebox{\linewidth}{!}{
\begin{tabular}{c|ccccc|cccc|ccccccccc|c}
      \hline
      \multirow{2}{*}{\parbox{2.85cm}{\centering \textbf{Gradient-based \\ Attacks}}}& \multicolumn{5}{c|}{\textbf{CNNs}} & \multicolumn{4}{c|}{\textbf{ViTs}} & \multicolumn{9}{c|}{\textbf{Defenses}} &\multirow{2}{*}{\textbf{Avg.}}\\
    
    & \textbf{RN-50} &\textbf{Inc-v3}&\textbf{MN-v3}&\textbf{DN-121} &\textbf{FSNet} &\textbf{ViT} &\textbf{PiT} &\textbf{Visformer} &\textbf{Swin} & \textbf{Inc-v3$_{ens3}$}& \textbf{Inc-v3$_{ens4}$} & \textbf{IncRes-v2$_{ens}$}& \textbf{AT} & \textbf{HGD} & \textbf{RS} & \textbf{NRP} & \textbf{DiffPure} & \textbf{MIMIR}\\
    \hline
  MI-FGSM & 94.9* & 34.5 & 40.6 & 45.9 & 25.2 &10.5 & 18.0 & 23.1 & 27.8 & 25.3& 25.6& 21.0&33.7 & 19.2& 21.9 &25.1 & 13.8 & 6.7& 28.5\\
   \rowcolor{gray!10}\textbf{LI-Boost-MI} & \textbf{97.0*} & \textbf{45.3} & \textbf{55.2} & \textbf{62.0} &\textbf{41.8} & \textbf{19.1} & \textbf{29.1} & \textbf{38.7} & \textbf{41.9}  & \textbf{35.5}& \textbf{34.8}& \textbf{30.3}&\textbf{34.3} & \textbf{33.3}& \textbf{23.8}&\textbf{32.2}&\textbf{21.4} & \textbf{7.2}&\textbf{37.9}\\
    \hline
    VMI-FGSM & 97.5* & 54.4 & 58.0 & 66.0 & 51.2 &28.6 & 40.5 & 46.5 & 49.4 & 44.5& 43.5& 38.9& 36.0& 44.4& 25.3& 44.1& 21.9& 8.2&44.4\\
       \rowcolor{gray!10}\textbf{LI-Boost-VMI} & \textbf{99.3*} & \textbf{67.0} & \textbf{71.4} & \textbf{79.3} & \textbf{65.9} & \textbf{39.7} & \textbf{53.7} & \textbf{61.4} & \textbf{62.6} & \textbf{57.8}& \textbf{57.1}& \textbf{52.8} & \textbf{37.9} & \textbf{59.1}&\textbf{30.4} &\textbf{60.4}& \textbf{36.8} & \textbf{9.0}&\textbf{55.6}\\
    \hline
    PGN & \textbf{99.1*} & 84.2 & 86.4  & 91.6 & 81.5& 54.8 & 69.7 &77.0& 78.8 & 77.7& 76.8& 73.0 & 46.2 & 78.3& 41.4 &79.7 &48.6 &13.7 & 69.9\\
       \rowcolor{gray!10}\textbf{LI-Boost-PGN }& 98.7* & \textbf{86.4} & \textbf{87.8} & \textbf{92.1} &\textbf{83.9} & \textbf{62.1} & \textbf{74.3} & \textbf{80.2} & \textbf{81.3}& \textbf{82.6}& \textbf{81.5}& \textbf{78.9}&\textbf{50.4} &\textbf{82.3} & \textbf{50.8}&\textbf{84.8}&\textbf{65.4} & \textbf{17.2}& \textbf{74.5}\\
    \hline
    MUMODIG & 97.1* & 72.8 & 78.4 & 84.7 & 72.1 & 42.9 & 58.6 & 67.7 & 66.4 & 62.4& 60.8& 55.1 & 37.6 &68.2 &26.4 &52.2 & 26.7 & 8.9& 57.7\\
     \rowcolor{gray!10}\textbf{LI-Boost-MUMODIG} & \textbf{98.6*} & \textbf{83.4} & \textbf{85.6} & \textbf{90.9} & \textbf{81.8} & \textbf{60.0} & \textbf{72.8} & \textbf{80.0} & \textbf{77.7}  & \textbf{74.4}& \textbf{72.7}& \textbf{69.5}& \textbf{41.3} & \textbf{78.7} & \textbf{33.4} &\textbf{90.2} & \textbf{45.8} & \textbf{12.3}& \textbf{69.4}\\
    \hline
\end{tabular}
}

\label{tab:gradient-based-attack}
\end{table*}

\begin{table*}[tb]
\centering
\captionsetup{labelsep=newline, justification=centering}
\caption{
 Attack success rates (\%) on nine normally trained models and nine defense mechanisms of various \textbf{input transformation-based attacks} w/wo LI-Boost. The adversarial examples are crafted on RN-50. * indicates the white-box model. The best results are \textbf{bold}, while methods incorporating LI-Boost are highlighted in \colorbox{gray!10}{gray}.
}
  \setlength{\tabcolsep}{1.5pt} 
  \setlength{\extrarowheight}{1.2pt}
\resizebox{\linewidth}{!}{
\begin{tabular}{c|ccccc|cccc|ccccccccc|c}
      \hline
      \multirow{2}{*}{\parbox{2.85cm}{\centering \textbf{Input Transforma-\\tion-based Attacks}}}& \multicolumn{5}{c|}{\textbf{CNNs}} & \multicolumn{4}{c|}{\textbf{ViTs}} & \multicolumn{9}{c|}{\textbf{Defenses}} & \multirow{2}{*}{\textbf{Avg.}}  \\
        & \textbf{RN-50} &\textbf{Inc-v3}&\textbf{MN-v3}&\textbf{DN-121}&\textbf{FSNet}&\textbf{ViT} &\textbf{PiT} &\textbf{Visformer} &\textbf{Swin} & \textbf{Inc-v3$_{ens3}$}& \textbf{Inc-v3$_{ens4}$} & \textbf{IncRes-v2$_{ens}$} & \textbf{AT} & \textbf{HGD} & \textbf{RS} & \textbf{NRP} & \textbf{DiffPure} & \textbf{MIMIR}\\
       \hline
    DIM &92.7*& 52.4 &56.7 & 64.4 & 46.9 & 23.9 & 35.0 & 42.1 & 43.6 & 39.9& 39.5& 34.6&34.9 &40.3 & 23.6&33.6&19.2& 7.2& 40.6\\
      \rowcolor{gray!10}\textbf{LI-Boost-DIM} &\textbf{98.1*} &\textbf{61.0} &\textbf{68.3} & \textbf{75.8} & \textbf{61.1} & \textbf{36.0}& \textbf{47.7}& \textbf{57.2}& \textbf{56.4} & \textbf{50.8} & \textbf{50.5}& \textbf{45.7} &\textbf{36.3} &\textbf{54.9} &\textbf{26.9} &\textbf{42.2}&\textbf{30.4} & \textbf{8.1} & \textbf{50.4}\\
      \hline
    \textit{Admix} &99.3* &59.4 &67.4 &77.6 & 54.6 &27.7 &41.8 &52.5 &53.3 & 43.2& 43.0& 36.8 &35.7 &47.9 &24.6 &44.4 & 20.8 & 7.9 & 46.6\\
      \rowcolor{gray!10}\textbf{LI-Boost-\textit{Admix}} &\textbf{99.5*} &\textbf{71.7} &\textbf{80.5} &\textbf{86.5} &\textbf{73.9} &\textbf{44.8} &\textbf{58.5} &\textbf{70.5} &\textbf{69.2}  & \textbf{61.7}& \textbf{61.0}& \textbf{56.3}&\textbf{38.4} &\textbf{66.9} &\textbf{30.9} &\textbf{57.7}& \textbf{38.4}  & \textbf{9.9}& \textbf{59.8}\\
    \hline
    SIA &99.3* &76.2 &89.1 &92.9 &81.3 &43.5 &66.8 &78.4 &76.6 & 61.6 & 58.7& 52.0 &38.0 &71.0 &27.2 &57.0& 25.4& 8.6& 61.3\\
      \rowcolor{gray!10}\textbf{LI-Boost-SIA} &\textbf{99.8*} &\textbf{87.0} &\textbf{95.1} &\textbf{96.8} &\textbf{91.8} &\textbf{64.0} &\textbf{81.2} &\textbf{90.3} &\textbf{88.1} & \textbf{77.1} & \textbf{75.8}& \textbf{71.1} &\textbf{42.2} &\textbf{86.8} &\textbf{36.0} &\textbf{71.2} &\textbf{45.4} & \textbf{11.3}& \textbf{72.8}\\
    \hline
    BSR &98.6* &84.6 &92.8 &95.7 &87.5 &53.1 &75.5 &84.7 &81.4 & 72.4& 69.7 & 64.5 &39.2 &81.4 &28.7& 58.2 & 31.3 & 9.4& 67.2\\
      \rowcolor{gray!10}\textbf{LI-Boost-BSR} &\textbf{99.2*} &\textbf{91.3} &\textbf{96.4} &\textbf{97.8} &\textbf{94.5} &\textbf{70.6} &\textbf{85.1} &\textbf{93.6} &\textbf{90.6} &\textbf{83.5} & \textbf{82.4}& \textbf{78.0} &\textbf{43.2} &\textbf{91.6} &\textbf{38.1} &\textbf{71.7} &\textbf{51.0} & \textbf{12.3}& \textbf{76.2}\\
    \hline

    SID & 98.9*& 90.8& 93.3& 95.7& 90.1& 69.3& 82.4& 89.4& 87.8& 83.9& 82.9& 79.1& 45.3& 88.5&39.9 &72.5 &48.2& 12.6& 75.0\\
    \rowcolor{gray!10}\textbf{LI-Boost-SID} & \textbf{99.7*}& \textbf{94.8}& \textbf{97.2}& \textbf{98.1}& \textbf{96.0}& \textbf{82.5}& \textbf{90.0}& \textbf{95.3}& \textbf{94.2}& \textbf{91.2}& \textbf{91.3}& \textbf{88.6} & \textbf{52.0}& \textbf{95.2}& \textbf{56.6}& \textbf{84.6}&\textbf{72.6} & \textbf{17.0}& \textbf{83.2}\\
    \hline
    
\end{tabular}
}
\label{tab:input-transformation-based-attack}
\vspace{-.3cm}
\end{table*}

\section{Experiments}

\subsection{Experimental Setup}
\label{sec:setup}

\textbf{Dataset.} We evaluate the proposed LI-Boost using 5,000 images from the validation set of the ImageNet dataset~\cite{russakovsky2015imagenet} for the image classification task, covering 1,000 categories. For the image captioning task, considering the higher computational costs of VLMs, we further uniformly randomly select 1,000 images from the aforementioned 5,000 images.

\textbf{Models.} To validate its effectiveness, we adopt various architectures as the victim models, including five CNNs, \textit{i.e.}, ResNet-50 (RN-50)~\cite{he2016deep}, Inception-v3 (Inc-v3)~\cite{2016inceptionv3}, MobileNet-v3-large (MN-v3)~\cite{howard2019mobilev3}, DenseNet-121 (DN-121)~\cite{huang2017densenet}, 
FasterNet-M (FSNet)~\cite{FSNet} and four ViTs, \ie, ViT-B16 (ViT)~\cite{2021vit}, PiT-B (PiT)~\cite{heo2021pit}, Visformer-Small (Visformer)~\cite{chen2021visformer}, Swin-Tiny-Patch4-Window7 (Swin)~\cite{liu2021swin}. To further substantiate the effectiveness of LI-Boost, we also consider several defense mechanisms, including three adversarially trained defense models, namely Inc-v3$_{ens3}$~\cite{tramer2018ensemble}, Inc-v3$_{ens4}$~\cite{tramer2018ensemble}, and IncRes-v2$_{ens}$~\cite{tramer2018ensemble}, and six state-of-the-art defense methods, namely AT~\cite{Wong2020ATdefense}, HGD~\cite{liao2018defense}, RS~\cite{cohen2019RS}, NRP~\cite{naseer2020NRP}, DiffPure~\cite{nie2022diffpure}, and MIMIR~\cite{xu2026mimir}. 

To evaluate the real-world impact of the proposed method, we test its effectiveness on three widely deployed commercial vision API systems, including Baidu, Alibaba, and Tencent. We also conduct image classification experiments across a diverse set of twelve representative vision-language models, including six open-source VLMs, \ie, Qwen3-VL~\cite{Qwen3VL}, Qwen2.5-VL~\cite{Qwen2.5-VL}, LLaVA-OneVision (LLaVA)~\cite{LLaVAOneVision}, Phi3.5-vision (Phi3.5)~\cite{phi3}, Intern-VL-3.5 (Intern-VL)~\cite{2025internvl35}, GLM-4.6V (GLM-V)~\cite{2025glm4} and six widely used commercial closed-source VLMs, \ie, GPT-4o~\cite{2024gpt}, GPT-5.2~\cite{2025gpt5}, Gemini-2.0-flash (Gemini)~\cite{2025gemini}, Claude-sonnet-4-5 (Claude-s)~\cite{2025claude45}, Claude-opus-4-5 (Claude-o)~\cite{2025claude45} and Grok-4-fast (Grok)~\cite{2025grok}. Furthermore, the evaluation is extended to the image captioning task, specifically targeting the six closed-source models to scrutinize their vulnerability.

\textbf{Baselines.} To assess the generality of LI-Boost, we establish several baselines encompassing multiple categories of transfer-based attacks, including \textbf{gradient-based attacks} (MI-FGSM~\cite{dong2018mifgsm}, VMI-FGSM~\cite{wang2021vmi}, PGN~\cite{ge2023boosting}, MUMODIG~\cite{ren2025mumodig}), \textbf{input transformation-based attacks} (DIM~\cite{xie2019dim},  \textit{Admix}~\cite{wang2021admix}, SIA~\cite{Wang2023SIA}, BSR~\cite{wang2024bsr}, SID~\cite{zhou2025sid}), \textbf{model-related attacks} (SGM~\cite{wu2020sgm}, Linbp~\cite{guo2020linbp},  BPA~\cite{wang2023bpa}, VDC~\cite{zhang2024VirtualDense}, FPR~\cite{ren2025fpr}), \textbf{advanced objective functions} (ILA~\cite{huang2019ila}, FIA~\cite{wang2021fia}, ILPD~\cite{Li2023ILPD}, BFA~\cite{WANG2024bfa}) and \textbf{ensemble attack}~\cite{dong2018mifgsm}. For consistency and fairness, we adopt MI-FGSM as the default backbone baseline across all experiments.

\textbf{Evaluation.}
\label{eval}We employ the attack success rates to assess the efficacy of attacks. To ensure a fair and consistent comparison across different attacks, we adopt a common attack setting with the perturbation budget $\epsilon = 16/255$, number of iterations $T = 10$, step size $\alpha = \epsilon/T$, and the decay factor $\mu=1.0$. We adopt $k=6$, $N=30$, and a logarithmic distribution to sample the translated perturbations for LI-Boost. All the baselines adopt the default parameters as in their original papers, which are detailed in the supplementary material, and all experiments are conducted on a server equipped with eight H20 GPUs. Input images are preprocessed to a resolution of 224$\times$224 for most models, while for Inc-v3, the resolution is set to 299$\times$299 to meet its architectural requirements.

\begin{table*}[tbp]
\centering
\captionsetup{labelsep=newline, justification=centering}
\caption{
 Attack success rates (\%) on nine normally trained models and nine defense mechanisms of various \textbf{model-related attacks} w/wo LI-Boost. The adversarial examples are crafted on RN-50, except for VDC and FPR, which are based on ViT. * indicates the white-box model. The best results are \textbf{bold}, while methods incorporating LI-Boost are highlighted in \colorbox{gray!10}{gray}.}
  \setlength{\tabcolsep}{1.5pt} 
  \setlength{\extrarowheight}{1.2pt}
 
\resizebox{\linewidth}{!}{
\begin{tabular}{c|ccccc|cccc|ccccccccc|c}
      \hline
      \multirow{2}{*}{\parbox{2.85cm}{\centering \textbf{Model-related \\ Attacks}}}& \multicolumn{5}{c|}{\textbf{CNNs}} & \multicolumn{4}{c|}{\textbf{ViTs}}  & \multicolumn{9}{c|}{\textbf{Defenses}} & \multirow{2}{*}{\textbf{Avg.}} \\
        & \textbf{RN-50} &\textbf{Inc-v3}&\textbf{MN-v3}&\textbf{DN-121} &\textbf{FSNet} &\textbf{ViT} &\textbf{PiT} &\textbf{Visformer} &\textbf{Swin} & \textbf{Inc-v3$_{ens3}$}& \textbf{Inc-v3$_{ens4}$} & \textbf{IncRes-v2$_{ens}$} & \textbf{AT} & \textbf{HGD} & \textbf{RS} & \textbf{NRP} & \textbf{DiffPure} &\textbf{MIMIR}\\
\hline
    SGM &99.5* &44.8 &57.2 &61.3 & 31.8 &15.0 &27.7 &33.4 &38.8 &30.5 &29.3 &24.1  &35.0 &22.8 &23.3 &29.3 &14.2 & 6.9& 34.7\\
       \rowcolor{gray!10}\textbf{LI-Boost-SGM}&\textbf{100.0*} &\textbf{61.5} &\textbf{78.4} &\textbf{82.0} &\textbf{46.7} &\textbf{29.1} &\textbf{46.4} &\textbf{57.6} &\textbf{61.0} &\textbf{40.6} &\textbf{39.3} &\textbf{32.5} &\textbf{36.8} &\textbf{48.0} &\textbf{27.1} &\textbf{41.3} &\textbf{25.4} & \textbf{8.0}& \textbf{47.9}\\
    \hline
    Linbp &89.2* &44.4 &55.8 &62.3 & 29.0 &9.6 &17.2 &28.6 &31.8 &31.5 &30.4 &24.7  &34.5 &24.2 &22.7 &27.7 &22.0 & 7.4& 32.9\\
       \rowcolor{gray!10}\textbf{LI-Boost-Linbp} &\textbf{99.2*} &\textbf{60.1} &\textbf{76.9} &\textbf{85.4} & \textbf{52.4} &\textbf{15.2} &\textbf{25.5} &\textbf{49.9} &\textbf{44.9} &\textbf{45.8} &\textbf{43.0} & \textbf{36.0}  &\textbf{34.6} &\textbf{43.7} &\textbf{24.1} &\textbf{32.8}&\textbf{23.0}& \textbf{7.5}&\textbf{44.4}\\
    \hline
    BPA &89.9* &79.6 &88.1 &96.4 & 66.9 &30.4 &46.0 &64.4 &65.7 & 65.2& 64.1 & 53.5&37.5 &69.2 & 27.9&47.9&28.1 & 7.2& 57.1\\
       \rowcolor{gray!10}\textbf{LI-Boost-BPA} &\textbf{93.0*} &\textbf{86.1} &\textbf{92.1} &\textbf{98.4} &\textbf{77.3} &\textbf{39.4} &\textbf{53.7} &\textbf{73.8} &\textbf{75.0} & \textbf{75.9}& \textbf{75.6}& \textbf{66.6}  &\textbf{40.6} &\textbf{81.4} & \textbf{34.7}&\textbf{57.1}&\textbf{43.3} & \textbf{8.6} &\textbf{65.1}\\
    \hline
    VDC &51.7 &58.6 &67.0 &65.6 &52.1 &\textbf{97.5*} &55.2 &59.3 &71.5 &45.7 &45.8 &40.0 &38.2 &41.8 & 28.8&35.8 &28.4  & 9.5& 49.6\\
       \rowcolor{gray!10}\textbf{LI-Boost-VDC} &\textbf{61.3} &\textbf{65.8} &\textbf{73.4} &\textbf{72.9} & \textbf{62.0} &96.7* &\textbf{66.7} &\textbf{68.9} &\textbf{76.8} &\textbf{52.3} &\textbf{53.4} &\textbf{47.7}  &\textbf{39.4} &\textbf{52.0} &\textbf{33.9} &\textbf{41.9} &\textbf{38.9} & \textbf{11.2} & \textbf{56.4}\\
    \hline
    FPR & 43.2 & 51.8 & 57.0 & 57.4 & 43.5 & \textbf{98.2*} & 45.8 & 49.7 & 61.3 & 38.9& 39.2& 33.4& 35.4& 33.7 & 24.8 & 30.4 & 22.0 & 8.0 &43.0\\
       \rowcolor{gray!10}\textbf{LI-Boost-FPR} &\textbf{53.5} &\textbf{57.8} &\textbf{63.7} &\textbf{63.5} &\textbf{54.6} & 96.8* &\textbf{58.1} &\textbf{60.4} &\textbf{68.3} & \textbf{44.7}& \textbf{45.7}& \textbf{40.1}&\textbf{36.7} &\textbf{43.3} &\textbf{27.9} &\textbf{34.8} &\textbf{29.9} & \textbf{8.9} & \textbf{49.4}\\      
    \hline
\end{tabular}
}
\label{tab:modelrelatedeval}
\end{table*}

\begin{table*}[tb]
\centering
\captionsetup{labelsep=newline, justification=centering}
\caption{
 Attack success rates (\%) on nine normally trained models and nine defense mechanisms of various \textbf{advanced objective functions} w/wo LI-Boost. The adversarial examples are crafted on RN-50. * indicates the white-box model. The best results are \textbf{bold}, while methods incorporating LI-Boost are highlighted in \colorbox{gray!10}{gray}.
}
  \setlength{\tabcolsep}{1.5pt} 
  \setlength{\extrarowheight}{1.2pt}
\resizebox{\linewidth}{!}{
\begin{tabular}{c|ccccc|cccc|ccccccccc|c}
      \hline
      \multirow{2}{*}{\parbox{2.85cm}{\centering \textbf{Advanced Objec-\\tive Functions}}}& \multicolumn{5}{c|}{\textbf{CNNs}} & \multicolumn{4}{c|}{\textbf{ViTs}}   & \multicolumn{9}{c|}{\textbf{Defenses}} & \multirow{2}{*}{\textbf{Avg.}} \\
        & \textbf{RN-50} &\textbf{Inc-v3}&\textbf{MN-v3}&\textbf{DN-121} &\textbf{FSNet} &\textbf{ViT} &\textbf{PiT} &\textbf{Visformer} &\textbf{Swin} & \textbf{Inc-v3$_{ens3}$}& \textbf{Inc-v3$_{ens4}$} & \textbf{IncRes-v2$_{ens}$} & \textbf{AT} & \textbf{HGD} & \textbf{RS} & \textbf{NRP} & \textbf{DiffPure} & \textbf{MIMIR}\\
       \hline
    ILA &90.0* &29.0 &37.9 &42.4 &22.4 &8.2 &15.4 &20.8 &27.1 &21.9 & 21.4& 16.6 &33.4 &13.9 &21.6 &20.0&11.4 &5.8 &25.5\\
       \rowcolor{gray!10}\textbf{LI-Boost-ILA} &\textbf{93.2*} &\textbf{41.3} &\textbf{56.7} &\textbf{64.6} &\textbf{36.2} &\textbf{12.2} &\textbf{23.2} &\textbf{33.4} &\textbf{39.1}& \textbf{30.6}& \textbf{30.4}& \textbf{24.1} &\textbf{33.8} &\textbf{24.3} &\textbf{22.6} &\textbf{24.6}&\textbf{14.1} & \textbf{6.4} &\textbf{33.9}\\
    \hline
    FIA &77.8* &37.5 &45.1 &53.4 &23.3 &8.1 &15.8 &20.9 &29.1 &27.8 &27.0 &20.1 &35.3 &16.6 &23.4 &24.7&12.2 & 6.0 &28.0\\
       \rowcolor{gray!10}\textbf{LI-Boost-FIA} &\textbf{89.6*} &\textbf{53.6} &\textbf{65.1} &\textbf{76.2} &\textbf{42.7} &\textbf{13.9}&\textbf{25.5} &\textbf{37.7} &\textbf{45.4} &\textbf{41.4} &\textbf{41.9} &\textbf{31.5}   &\textbf{36.7} &\textbf{32.8} &\textbf{25.2} &\textbf{30.8}&\textbf{15.4} & \textbf{6.5} &\textbf{39.6}\\
    \hline
    ILPD &\textbf{95.0*} &65.6 &74.1 &80.6 &65.0 &62.0 &52.7 &61.4 &61.9& 55.1& 54.9& 49.5 &46.8 &57.0 &27.5 &55.3 &28.5 & 9.6 &55.7\\
       \rowcolor{gray!10}\textbf{LI-Boost-ILPD}&94.3* &\textbf{69.5} &\textbf{79.7} &\textbf{84.7} &\textbf{69.6} &\textbf{66.9} &\textbf{56.3} &\textbf{67.7} &\textbf{65.9} & \textbf{60.7}& \textbf{60.4}& \textbf{55.9}&\textbf{51.5} &\textbf{62.5} &\textbf{31.0} &\textbf{58.3} &\textbf{35.1} & \textbf{10.7}&\textbf{60.0}\\
    \hline
    BFA &\textbf{98.8*} &82.9 &90.5 &94.5 &84.4 &46.0 &67.5 &79.8 &79.7& 72.8& 70.6& 63.1 & 39.5& 77.0&29.0 &68.5 & 27.1 & 8.5 &65.6\\
       \rowcolor{gray!10}\textbf{LI-Boost-BFA} &98.7* &\textbf{86.8} &\textbf{92.6} &\textbf{96.0} &\textbf{87.9} &\textbf{53.8} &\textbf{72.1} &\textbf{84.8} &\textbf{83.8} & \textbf{79.7}& \textbf{78.6}& \textbf{72.8} &\textbf{42.3} &\textbf{83.7} &\textbf{36.3} &\textbf{74.5} &\textbf{44.4} & \textbf{9.6} &\textbf{71.0}\\
    \hline
\end{tabular}
}
\vspace{-.3cm}
\label{tab:advancedeval}
\end{table*}

\subsection{Evaluation on Gradient-based Attacks}
 
To validate the effectiveness of our proposed LI-Boost, we first integrate it into various gradient-based attacks, \textit{i.e.}, MI-FGSM, VMI-FGSM, PGN, and MUMODIG. We generate adversarial examples on RN-50 and evaluate transferability on the other CNNs, ViTs, and defense mechanisms. The results are summarized in Tab.~\ref{tab:gradient-based-attack}, and results using other surrogate models are in supplementary material.

As we can observe, LI-Boost significantly improves the white-box attack performance on RN-50, underscoring the advantage of increasing local invariance to strengthen adversarial perturbations. Regarding black-box performance, MI-FGSM exhibits the lowest transferability among the baseline methods, whereas VMI-FGSM, PGN, and MUMODIG demonstrate considerably stronger attack capabilities. Notably, LI-Boost consistently boosts the performance across both CNN and ViT architectures. On average, the attack success rates show significant improvement, with the increases of 9.7$\%$, 11.9$\%$, 4.7$\%$  and 12.2$\%$ for MI-FGSM, VMI-FGSM, PGN, and MUMODIG, respectively. Furthermore, even when facing robust defense mechanisms, LI-Boost significantly enhances the attack performance. For instance, LI-Boost-PGN consistently maintains its dominance under all six evaluated defense approaches, reaching an attack success rate of 84.8\% against NRP and 65.4\% against DiffPure; even when faced with the SOTA method MIMIR, it still outperforms the baseline by 3.5\%. These consistent gains highlight the effectiveness and generalizability of LI-Boost across diverse model architectures and defense strategies, reveal the limitations of existing defenses, and raise new critical concerns about model robustness under transfer-based adversarial attacks.

\subsection{Evaluation on Input Transformation-based Attacks}
 
To assess the generality of LI-Boost, we integrate it with five prominent input transformation-based attacks, \textit{i.e.}, DIM, \textit{Admix}, SIA, BSR, and SID. As shown in Tab.~\ref{tab:input-transformation-based-attack}, LI-Boost significantly enhances the performance of white-box attacks, achieving near-perfect success rates of approximately $100.0\%$. This further corroborates the hypothesis that increasing local invariance strengthens adversarial attacks. Under black-box settings, LI-Boost consistently boosts the performance of various input transformation-based attacks. Overall, the integration of LI-Boost results in substantial performance gains over the baseline methods under normally trained models: an improvement of  8.6\%$\sim$15.1\% for DIM, 8.9\%$\sim$19.3\% for \textit{Admix}, 3.9\%$\sim$20.5\% for SIA, 2.1\%$\sim$17.5\% for BSR, and 2.4\%$\sim$13.2\% for SID. Furthermore, attacks augmented with LI-Boost demonstrate superior robustness under various defense mechanisms. For instance, the performance of \textit{Admix} when coordinated with LI-Boost increases by 19.5\% when attacking the IncRes-v2$_{ens}$ model, and LI-Boost-SID achieves a formidable success rate of 91.3\% on Inc-v3$_{ens4}$, representing a compelling advancement over the vanilla baseline. Besides, results of utilizing other surrogate models are presented in supplementary material. These significant improvements not only underscore the remarkable effectiveness of LI-Boost in enhancing transferability across diverse attack scenarios, but also reinforce the strategic value of local invariance in enhancing the cross-model generalization of adversarial examples that remain potent even under stringent black-box constraints.


\subsection{Evaluation on Model-related Attacks}
 
To evaluate the efficacy of LI-Boost in model-related attacks, we integrate it with five prominent methods, \textit{i.e.}, SGM, Linbp, BPA for CNNs and VDC, FPR for ViTs. The experimental results, presented in \mbox{Tab.}~\ref{tab:modelrelatedeval}, demonstrate that attacks augmented with LI-Boost not only maintain high success rates in white-box settings but also achieve substantial improvements over the baseline methods in black-box scenarios: 14.1\%$\sim $24.2\% for SGM, 5.6\%$\sim$23.4\% for Linbp, 2.0\%$\sim$10.4\% for BPA, 5.3\%$\sim$11.5\% for VDC and 6.0\%$\sim$12.3\% for FPR. These results highlight that LI-Boost significantly outperforms the baselines by considerable margins. Moreover, LI-Boost achieves higher success rates across evaluated defense strategies. These findings underscore the effectiveness of LI-Boost in augmenting adversarial attacks, highlighting local invariance as a pivotal mechanism for overcoming model-specific architectural boundaries.
\begin{table*}[tb]
\centering
\captionsetup{labelsep=newline, justification=centering}
\caption{
 Attack success rates (\%) on nine normally trained models and nine defense mechanisms of various \textbf{ensemble attacks} w/wo LI-Boost. The adversarial examples are crafted on RN-50, Inc-v3, MN-v3, and DN-121.  * indicates the white-box models. The best results are \textbf{bold}, while methods incorporating LI-Boost are highlighted in \colorbox{gray!10}{gray}.
}
  \setlength{\tabcolsep}{1.5pt} 
  \setlength{\extrarowheight}{1.2pt}
\resizebox{\linewidth}{!}{
\begin{tabular}{c|ccccc|cccc|ccccccccc|c}
      \hline
      \multirow{2}{*}{\parbox{2.85cm}{\centering \textbf{Ensemble\\ Attacks}}}& \multicolumn{5}{c|}{\textbf{CNNs}} & \multicolumn{4}{c|}{\textbf{ViTs}} & \multicolumn{9}{c|}{\textbf{Defenses}} &  \multirow{2}{*}{\textbf{Avg.}} \\
        & \textbf{RN-50} &\textbf{Inc-v3}&\textbf{MN-v3}&\textbf{DN-121} &\textbf{FSNet} &\textbf{ViT} &\textbf{PiT} &\textbf{Visformer} &\textbf{Swin}& \textbf{Inc-v3$_{ens3}$}& \textbf{Inc-v3$_{ens4}$} & \textbf{IncRes-v2$_{ens}$} & \textbf{AT} & \textbf{HGD} & \textbf{RS} & \textbf{NRP} & \textbf{DiffPure} & \textbf{MIMIR}\\
       \hline
      MI-FGSM$_{ens}$ & 95.4* & 99.8* & 99.3* & \textbf{100.0*} &67.8 & 39.3 & 53.7 & 66.6 & 68.5 &61.0 &60.6 & 51.3& 37.2 & 66.6& 27.1 &44.7 & 24.6& 8.5& 59.6\\
       \rowcolor{gray!10}\textbf{LI-Boost-MI$_{ens}$} & \textbf{97.9*} & \textbf{100.0*} & \textbf{99.6*} & \textbf{100.0*} & \textbf{85.6} & \textbf{58.4} & \textbf{71.0} & \textbf{83.2} & \textbf{83.9}  &\textbf{78.4} &\textbf{77.7} &\textbf{70.6} &\textbf{39.9} & \textbf{85.3}& \textbf{34.5}&\textbf{58.3}&\textbf{40.9} & \textbf{10.0}& \textbf{70.8}\\
    \hline
    VMI-FGSM$_{ens}$ & 97.3* & 99.9* & 99.4* &  \textbf{100.0*} & 84.7& 60.1 & 73.0 & 82.4 & 83.3& 80.6& 79.6& 73.1 & 40.5& 84.5& 33.4& 66.0 & 39.7 & 13.9& 71.7\\
       \rowcolor{gray!10}\textbf{LI-Boost-VMI$_{ens}$} & \textbf{99.3*} & \textbf{100.0*} & \textbf{99.7*} & \textbf{100.0*} & \textbf{93.1} & \textbf{73.7} & \textbf{84.5} & \textbf{91.8} & \textbf{92.4}& \textbf{91.3}& \textbf{90.0}& \textbf{85.4} & \textbf{45.0} & \textbf{94.0}&\textbf{42.6} &\textbf{83.6}& \textbf{58.2} & \textbf{15.1} & \textbf{80.0}\\
    \hline
    PGN$_{ens}$ & \textbf{98.8*} & \textbf{100.0*} & \textbf{99.6*}  & \textbf{100.0*} & 94.6 & 81.2 & 88.7 &94.1& 94.1 & 95.1& 94.8& 91.7& 54.9 & 95.9& 58.4 &90.0 &71.0& 19.6& 84.6\\
       \rowcolor{gray!10}\textbf{LI-Boost-PGN$_{ens}$} & 98.7* & 99.7* & 99.5* & \textbf{100.0*} & \textbf{95.4} & \textbf{83.6} & \textbf{90.0} & \textbf{94.5} & \textbf{94.6} & \textbf{96.2}& \textbf{95.9}& \textbf{93.5}& \textbf{60.4} &\textbf{96.7} & \textbf{68.8}&\textbf{93.3}&\textbf{83.6} & \textbf{22.7}& \textbf{87.1} \\
    \hline
    MUMODIG$_{ens}$ & \textbf{99.6*} & \textbf{99.8*} & \textbf{99.8*}  & \textbf{100.0*} & 97.2 & 84.2 & 92.3 &97.1& 96.3 & 95.9& 95.8& 93.2&46.4 & 98.0 &  40.4 &82.4  & 52.8& 14.0& 82.5\\
       \rowcolor{gray!10}\textbf{LI-Boost-MUMODIG$_{ens}$} & \textbf{99.6*} & \textbf{99.8*} & \textbf{99.8*} & \textbf{100.0*} & \textbf{98.1} & \textbf{89.1} & \textbf{95.0} & \textbf{98.2} &  \textbf{97.5}& \textbf{97.5}&\textbf{97.0}& \textbf{95.6} & \textbf{51.9}&\textbf{98.8}&\textbf{51.1} & \textbf{90.6}& \textbf{72.8} &\textbf{19.2} & \textbf{86.2}\\
    \hline
    
    DIM$_{ens}$ & 97.8* & \textbf{99.9*} & 99.6* & \textbf{100.0*} & 86.1 & 61.5 & 74.5 &84.4& 84.7 & 82.7& 81.2& 75.3& 39.9 & 87.6& 31.9 &60.4 &38.2& 10.1& 72.0 \\
       \rowcolor{gray!10}\textbf{LI-Boost-DIM$_{ens}$} & \textbf{99.0*} & \textbf{99.9*} & \textbf{99.8*} & \textbf{100.0*} & \textbf{93.2} & \textbf{76.1} & \textbf{84.4} & \textbf{92.4} & \textbf{92.1} & \textbf{90.8}& \textbf{90.0}& \textbf{85.5}& \textbf{44.4} &\textbf{94.4} & \textbf{42.7}&\textbf{71.8}&\textbf{58.1}&\textbf{13.1} & \textbf{79.3} \\
    \hline
    \textit{Admix}$_{ens}$ & \textbf{99.5*} & \textbf{100.0*} & 99.8* &\textbf{100.0*} & 92.7 & 69.5 & 82.7 &91.8& 92.0& 90.3&89.0 &83.5& 44.5 & 93.5 & 37.7 &75.9 &43.2 &11.8 & 77.6\\
       \rowcolor{gray!10}\textbf{LI-Boost-\textit{Admix}$_{ens}$} & 99.4* & \textbf{100.0*} & \textbf{100.0*} & \textbf{100.0*} & \textbf{96.1} & \textbf{83.1} & \textbf{89.6} & \textbf{95.6} & \textbf{95.4} & \textbf{95.0}& \textbf{95.0} & \textbf{92.0}&\textbf{51.0} &\textbf{97.5} & \textbf{53.3}&\textbf{85.5}&\textbf{70.2} & \textbf{17.1}& \textbf{84.2} \\
    \hline
    SIA$_{ens}$ & 99.8* & \textbf{100.0*} & \textbf{100.0*}  & \textbf{100.0*} & 98.0 & 82.3 & 93.6 &97.9& 97.3 & 95.2& 94.3& 90.2 & 44.6 & 98.6& 37.5 &78.7 &46.1 & 12.8& 81.5\\
       \rowcolor{gray!10}\textbf{LI-Boost-SIA$_{ens}$} & \textbf{99.9*} & 99.9* & \textbf{100.0*} & \textbf{100.0*} & \textbf{99.5} & \textbf{91.9} & \textbf{96.8} & \textbf{99.3} & \textbf{99.0} & \textbf{98.2}& \textbf{98.0}& \textbf{96.4} &  \textbf{51.8} &\textbf{99.5} & \textbf{53.2}&\textbf{89.5}&\textbf{71.0} &\textbf{17.4} & \textbf{86.7} \\
    \hline
    BSR$_{ens}$ & 99.8* & 99.6* & \textbf{100.0*}  & \textbf{100.0*} & 89.5 & 80.4& 92.5 & 97.6 & 96.0  & 94.7& 94.1& 90.3&  45.7 & 98.2 &  37.8 & 74.3 & 48.4 & 12.1& 80.6\\
       \rowcolor{gray!10}\textbf{LI-Boost-BSR$_{ens}$}& \textbf{99.9*} &  \textbf{99.9*} &  \textbf{100.0*} & \textbf{100.0*} & \textbf{99.3} & \textbf{90.6} & \textbf{95.8} & \textbf{99.2} & \textbf{98.7} & \textbf{97.6}& \textbf{97.8}& \textbf{96.2} & \textbf{52.4} &\textbf{99.5} & \textbf{53.4}&\textbf{86.2}&\textbf{74.4} & \textbf{15.0}& \textbf{86.4} \\
    \hline
    SID$_{ens}$ & 99.8* & \textbf{99.9*} & \textbf{99.9*} & \textbf{100.0*}& 99.0 & 92.5 & 96.9 & 99.1 & 98.8 & 98.6& 98.7& 97.7& 55.2 & 99.4 & 55.9 & 90.4 & 74.5 & 17.9& 87.5\\
       \rowcolor{gray!10}\textbf{LI-Boost-SID$_{ens}$}& \textbf{99.9*} &  \textbf{99.9*} &  \textbf{99.9*} & \textbf{100.0*} & \textbf{99.3} & \textbf{93.6} & \textbf{96.8} & \textbf{99.3} & \textbf{99.0} & \textbf{98.9}& \textbf{99.0}& \textbf{98.2} & \textbf{61.6} &\textbf{99.6} & \textbf{72.1}&\textbf{93.7}&\textbf{89.1} & \textbf{22.4}&  \textbf{90.1}\\
    \hline
\end{tabular}
}
\vspace{-.3cm}
\label{tab:ensemble-attack}
\end{table*}
\subsection{Evaluation on Advanced Objective Functions} 
To validate the effectiveness of LI-Boost in advanced objective functions, we integrate our LI-Boost with  ILA, FIA, ILPD, and BFA. The results are presented in Tab.~\ref{tab:advancedeval}. As we can see from the table, under white-box settings,  LI-Boost significantly improves the success rates of ILA and FIA by 3.2\% and 11.8\%, respectively, while maintaining the performance of ILPD and BFA. For black-box settings, ILA exhibits the weakest performance among the three baseline methods, whereas FIA, ILPD, and BFA demonstrate superior efficacy. Notably, LI-Boost substantially enhances the attack performance across both CNNs and ViTs. In particular, the magnitudes of improvement for ILA, FIA, ILPD, and BFA are 3.2\%$\sim$22.2\%, 5.8\%$\sim$22.8\%, 3.6\%$\sim$6.3\% and 1.5\%$\sim$7.8\%, respectively. Additionally, we evaluate the attack performance against different defenses, where LI-Boost can still boost the baselines' performance. For instance, ILPD achieves an average success rate of 42.7\% while LI-Boost-ILPD attains 47.3\%. These performance improvements convincingly illustrate that LI-Boost can significantly boost the adversarial transferability and further validate that local invariance is fundamental to ensuring the potency of attacks.

\subsection{Evaluation on Ensemble Attack}

To further validate the efficacy of our method, we adopt the ensemble attack as in MI-FGSM~\cite{dong2018mifgsm}, by fusing the logit outputs of diverse models. The adversarial examples are generated on RN-50, Inc-v3, MN-v3, and DN-121 using nine baselines w/wo LI-Boost, and all ensemble models are assigned equal weights. As shown in Tab.~\ref{tab:ensemble-attack}, empirical results reveal that baseline methods consistently achieve enhanced adversarial transferability when integrated with LI-Boost. The augmented methods not only exhibit improved attack success rates in white-box scenarios but also demonstrate remarkable performance gains in black-box settings. Furthermore, comprehensive evaluations across nine representative defense mechanisms highlight the effectiveness of our approach. Such consistent improvements across various baselines demonstrate that LI-Boost can further unlock the potential of ensemble attacks, ensuring higher transferability performance against various black-box models and attack scenarios.

\subsection{Practical Threats to Commercial Vision API Systems}
\begin{table}[tbp]
\captionsetup{labelsep=newline, justification=centering}
\centering
\caption{Attack success rates (\%) of various adversarial attacks against three commercial vision API systems. The last two columns denote the average success rates across all platforms. The best results are \textbf{bold} while methods incorporating LI-Boost are highlighted in \colorbox{gray!10}{gray}. The surrogate model is RN-50.}  
\label{tab:cloud_apis}
\renewcommand{\arraystretch}{1} 
\setlength{\tabcolsep}{1.5pt}
\resizebox{\columnwidth}{!}{ 
\begin{tabular}{c|cc|cc|cc|cc}
\hline
\multirow{2}{*}{\textbf{Method}} & \multicolumn{2}{c|}{\textbf{Baidu}} & \multicolumn{2}{c|}{\textbf{Alibaba}} & \multicolumn{2}{c|}{\textbf{Tencent}} & \multicolumn{2}{c}{\textbf{Avg.}} \\ 
& Top-1 & Top-5 & Top-1 & Top-5 & Top-1 & Top-5 & Top-1 & Top-5 \\ 
\hline
MI-FGSM       & 65.9 & 41.8 & 45.5 & 18.2 & 54.9 & 20.1 & 55.4 & 26.7 \\
\rowcolor{gray!10}
\textbf{LI-Boost-MI}  & \textbf{70.8} & \textbf{48.9} & \textbf{56.2} & \textbf{28.4} & \textbf{64.2} & \textbf{30.3} & \textbf{63.7} & \textbf{35.9} \\ 
\hline
BSR          & 85.2 & 71.1 & 87.5 & 71.2 & 72.9 & 42.6 & 81.9 & 61.6 \\
\rowcolor{gray!10}
\textbf{LI-Boost-BSR} & \textbf{89.3} & \textbf{77.7} & \textbf{92.5} & \textbf{81.2} & \textbf{81.3} & \textbf{55.2} & \textbf{87.7}& \textbf{71.4}\\ 
\hline
ILA & 60.7 & 35.3 & 36.3 & 11.6 & 50.9 & 17.3 & 49.3 & 21.4 \\
\rowcolor{gray!10}
\textbf{LI-Boost-ILA} & \textbf{70.1} & \textbf{47.9}& \textbf{58.7} & \textbf{31.1} & \textbf{66.3} & \textbf{32.1} & \textbf{65.0} & \textbf{37.0}\\ 
\hline
BPA          & 85.8 & 72.7 & 86.8 & 68.4 & 78.6 & 53.3 & 83.7 & 64.8 \\
\rowcolor{gray!10}
\textbf{LI-Boost-BPA} & \textbf{89.5} & \textbf{78.1} & \textbf{90.3} & \textbf{76.5} & \textbf{83.1} & \textbf{61.2} & \textbf{87.6} & \textbf{71.9} \\ 
\hline
\end{tabular}
}
\vspace{-.3cm}
\end{table}

  To further validate the efficacy of LI-Boost and examine its practical security implications in commercial vision API systems, we evaluate the transferability of our approach against three representative systems. As reported in Tab.~\ref{tab:cloud_apis}, the integration of LI-Boost consistently improves the Top-1 and Top-5 attack success rates across all methods and platforms. For instance, when integrated with MI-FGSM, LI-Boost increases the average Top-1 attack success rate from 55.4\% to 63.7\%. Notably, LI-Boost-BPA achieves the strongest performance, with an average Top-1 success rate of 87.6\% and an average Top-5 success rate of 71.9\%. Specifically, LI-Boost-BSR achieves a peak Top-1 attack success rate of 92.5\% on the Alibaba platform, underscoring its significant threat to commercial vision API systems.

  Furthermore, the efficacy of our approach is even more pronounced under the stricter Top-5 criterion. A notable case is LI-Boost-ILA, which nearly triples the attack success rate from a baseline of 11.6\% to 31.1\% when attacking the Alibaba vision API system. This substantial gain demonstrates that LI-Boost can effectively resurrect relatively weak attackers, enabling them to bypass the robust defense mechanisms of commercial vision API service systems. Besides, we also present visualizations of examples against the Alibaba Vision API in the supplementary material. These consistent improvements underscore the effectiveness of LI-Boost in deceiving the predictions of black-box commercial image recognition models and substantial security risks uncovered by LI-Boost to the integrity of top-tier commercial vision API systems.

\begin{table*}[htbp] 
\centering
\captionsetup{labelsep=newline, justification=centering}
\caption{Attack success rates (\%) on six open-source and six closed-source VLMs of four attacks w/wo LI-Boost on image classification task. The best results are \textbf{bold}, while methods incorporating LI-Boost are highlighted in \colorbox{gray!10}{gray}. The adversarial examples are crafted on RN-50.}
  \setlength{\tabcolsep}{1.5pt} 
  \setlength{\extrarowheight}{1.2pt}
\resizebox{\linewidth}{!}
{
\begin{tabular}{c|cccccc|cccccc|c} 
\hline
      \multirow{2}{*}{\textbf{Attacks}} & \multicolumn{6}{c|}{\textbf{Open-source}} & \multicolumn{6}{c|}{\textbf{Closed-source}} & \multirow{2}{*}{\textbf{Avg.}} \\
      & \textbf{Qwen3-VL} & \textbf{Qwen2.5-VL} & \textbf{LLaVA} & \textbf{Phi3.5} &\textbf{Intern-VL} &\textbf{GLM-V} & \textbf{GPT-4o} & \textbf{GPT-5.2} & \textbf{Gemini} & \textbf{Claude-s} & \textbf{Claude-o} & \textbf{Grok}\\
    \hline
      MIFGSM & 26.2 & 31.6 & 24.2 & 41.9 & 37.4 & 21.0& 20.6&31.0 & 24.4& 38.7& 31.9 & 34.3 & 30.3\\
      \rowcolor{gray!10}\textbf{LI-Boost-MI} &\textbf{31.2} & \textbf{38.2} & \textbf{30.3}& \textbf{47.2} & \textbf{42.8}&\textbf{25.9} &\textbf{25.5} &\textbf{36.7} & \textbf{29.2} & \textbf{41.9} & \textbf{38.8} & \textbf{40.7} & \textbf{35.7}\\
      \hline
      BSR & 54.9 & 61.3 & 51.1 & 64.4 & 66.8 & 45.3 & 54.4 & 62.2 & 54.6 & 57.6& 68.2 & 62.4 & 58.6\\
      \rowcolor{gray!10}\textbf{LI-Boost-BSR} &\textbf{64.6} & \textbf{70.3} & \textbf{60.9}& \textbf{70.3} &\textbf{72.3} & \textbf{58.0} & \textbf{59.7}& \textbf{70.0}&\textbf{59.8} &  \textbf{66.5} & \textbf{76.0} & \textbf{70.0} & \textbf{66.5}\\
      \hline
      ILA & 24.8 & 30.1 & 22.9 & 40.7 & 34.7 &19.8 & 20.5 & 28.4 & 22.3 & 33.1 & 28.4 & 30.9 & 28.1\\
      \rowcolor{gray!10}\textbf{LI-Boost-ILA} &\textbf{31.4} & \textbf{41.1} & \textbf{29.8}& \textbf{49.2} & \textbf{45.4}& \textbf{24.8}& \textbf{26.3} & \textbf{39.4} & \textbf{28.6}& \textbf{42.7}& \textbf{41.8} & \textbf{40.2} & \textbf{36.7}\\
      \hline
      BPA & 46.8 & 54.9 & 45.0 & 59.8 & 58.3 &39.2 &41.4 &56.8 & 42.6 & 53.4& 59.5 & 56.5 & 51.2\\
      \rowcolor{gray!10}\textbf{LI-Boost-BPA} & \textbf{52.7}& \textbf{61.3} & \textbf{51.7}& \textbf{64.6} & \textbf{63.5}& \textbf{46.9}& \textbf{45.6}& \textbf{61.5} & \textbf{45.3}& \textbf{60.3}& \textbf{63.9} & \textbf{62.5} & \textbf{56.7}\\
    \hline
\end{tabular}
}
\label{tab:attackvlm}
\vspace{-.2cm}
\end{table*}

\subsection{Practical Threats to Vision-Language Models} 

To further validate the generality of LI-Boost, we evaluate its attack effectiveness on vision–language models (VLMs). We first conduct experiments on an image classification task involving twelve VLMs. Adversarial examples are generated on the RN-50 backbone using MI-FGSM, BSR, ILA, BPA, and their corresponding LI-Boost-enhanced variants. These adversarial samples are then transferred to various VLMs, which are prompted to identify the primary object in each image. The prompt for classification is illustrated in Fig.~\ref{prompt:imageclassify}, and the quantitative attack results are reported in Tab.~\ref{tab:attackvlm}, which demonstrate the superior transferability of attacks enhanced with LI-Boost in the image classification task. Specifically, LI-Boost consistently improves the attack success rates of all baseline methods across the twelve evaluated VLMs. For example, LI-Boost increases the average attack success rate of BSR from 58.6\% to 66.5\%, corresponding to a substantial 7.9\% absolute improvement. These consistent enhancements across both open-source and closed-source models suggest that LI-Boost effectively exploits model-agnostic adversarial features, thereby improving transferability.

\begin{figure}[tbp]
    \centering
    \begin{promptbox}{}
        Output the kind in this image using one word. \\
        Use the specific, standard singular noun for the animal or object (e.g., \\ {\color{Score10}{\textbf{``dog''} is correct}}; {\color{red}{do not use plural forms like ``puppies''}}; \\ 
        {\color{Score10}{\textbf{``koi''} is correct}}; {\color{red}{do not use broader terms like ``fish''}}). \\
        Do not use adjectives, descriptive terms, or general categories.\\ 
        Only output a single singular noun.
    \end{promptbox}
\vspace{-.2cm}
    \caption{Prompt for image classification task.}
    \label{prompt:imageclassify} 
    \vspace{-.3cm}
\end{figure}

\begin{figure}[tbp]
    \centering
    \footnotesize 
    \renewcommand{\arraystretch}{0.2} 
    \setlength{\tabcolsep}{2pt} 

    \begin{tabular}{ccc:ccc}
        \textbf{\scriptsize Clean} & \textbf{\scriptsize MI-FGSM} & \textbf{\scriptsize LI-Boost-MI} & \textbf{\scriptsize Clean} & \textbf{\scriptsize MI-FGSM} & \textbf{\scriptsize LI-Boost-MI}\\ [1.5ex]
        
        \includegraphics[width=0.15\linewidth]{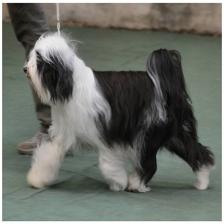} &
        \includegraphics[width=0.15\linewidth]{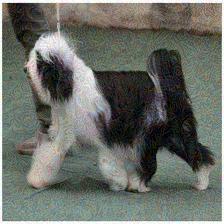} &
        \includegraphics[width=0.15\linewidth]{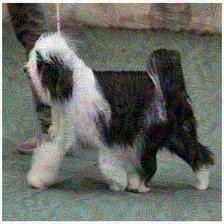} &

        \includegraphics[width=0.15\linewidth]{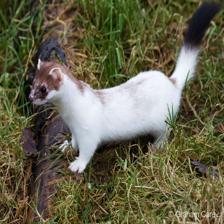} &
        \includegraphics[width=0.15\linewidth]{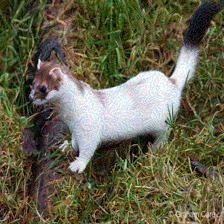} &
        \includegraphics[width=0.15\linewidth]{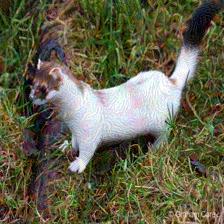} \\[1ex]
        \textbf{\textcolor{green!50!black}{\scriptsize Dog}} & \textbf{\textcolor{gray}{\scriptsize Dog}} & \textbf{\textcolor{red}{\scriptsize Cat}} & \textbf{\textcolor{green!50!black}{\scriptsize Ferret}} & \textbf{\textcolor{gray}{\scriptsize Ferret}} & \textbf{\textcolor{red}{\scriptsize Cat}} \\ [2ex]

        \includegraphics[width=0.15\linewidth]{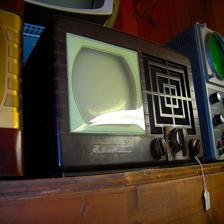} &
        \includegraphics[width=0.15\linewidth]{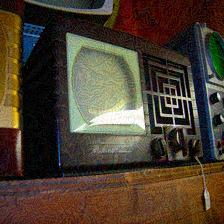} &
        \includegraphics[width=0.15\linewidth]{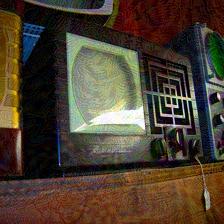} &
        \includegraphics[width=0.15\linewidth]{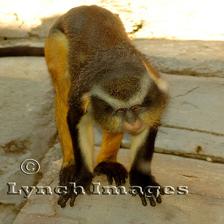} &
        \includegraphics[width=0.15\linewidth]{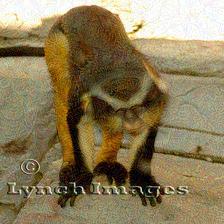} &
        \includegraphics[width=0.15\linewidth]{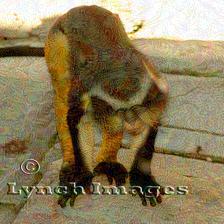} \\[1ex]
        \textbf{\textcolor{green!50!black}{\scriptsize Television}} & \textbf{\textcolor{gray}{\scriptsize Television}} & \textbf{\textcolor{red}{\scriptsize Labyrinth}} & \textbf{\textcolor{green!50!black}{\scriptsize Monkey}} & \textbf{\textcolor{gray}{\scriptsize Monkey}} & \textbf{\textcolor{red}{\scriptsize Koala}} \\ [2ex]


        \includegraphics[width=0.15\linewidth]{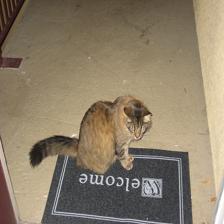} &
        \includegraphics[width=0.15\linewidth]{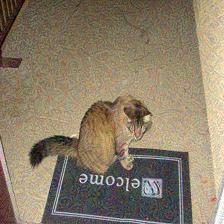} &
        \includegraphics[width=0.15\linewidth]{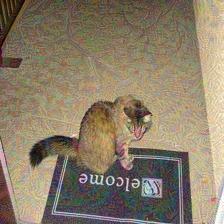} &
        \includegraphics[width=0.15\linewidth]{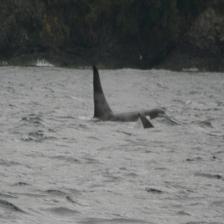} &
        \includegraphics[width=0.15\linewidth]{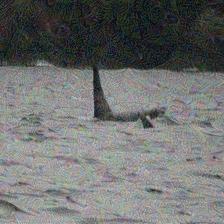} &
        \includegraphics[width=0.15\linewidth]{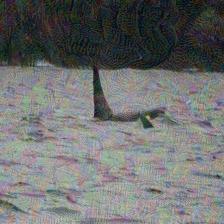} \\[1ex]
        
        \textbf{\textcolor{green!50!black}{\scriptsize Cat}} & \textbf{\textcolor{gray}{\scriptsize Cat}} & \textbf{\textcolor{red}{\scriptsize Ferret}} &        \textbf{\textcolor{green!50!black}{\scriptsize Orca}} & \textbf{\textcolor{gray}{\scriptsize Orca}} & \textbf{\textcolor{red}{\scriptsize Goose}} \\ [1ex]
        
    \end{tabular}

    \caption{Visualizations and comparisons of clean images, adversarial examples generated through MI-FGSM and LI-Boost-MI against the RN-50 on Qwen3-VL classification task.  \textbf{\textcolor{green!50!black}{Green text}} represents the ground-truth labels. \textbf{\textcolor{gray}{Gray text}} denotes the results of failed MI-FGSM examples. \textcolor{red}{\textbf{Red text}} indicates the misclassification of LI-Boost-MI.}
    \label{fig:qwen3vlclassify}
    \vspace{-.6cm}
\end{figure}

Furthermore, to qualitatively illustrate the impact of LI-Boost, we present visualizations and results on the state-of-the-art open-source VLM, Qwen3-VL, as shown in Fig.~\ref{fig:qwen3vlclassify}. While the vanilla MI-FGSM fails to deceive the model and produces predictions consistent with the ground-truth labels, LI-Boost-MI consistently induces significant misclassifications. These examples highlight the strong attack capability of the LI-Boost-enhanced methods.

In addition, to evaluate the cross-task generalizability of our method, we extend the evaluation to image captioning. We leverage an LLM-as-a-judge framework~\cite{2023llmjudge}, using GPT-4o to quantify the semantic shift between captions generated from clean and adversarial images. The prompt for evaluation is illustrated in Fig.~\ref{fig:promptimagecaption}, and an attack is deemed successful if the score falls below 8, indicating a significant disruption in multimodal semantic alignment.
Tab.~\ref{table_asr_results} shows that LI-Boost consistently outperforms all baseline attacks. For instance, LI-Boost-BPA improves the average attack success rates by 11.4\%, while LI-Boost-BSR reaches a peak attack success rate of 84.3\% on Grok. As illustrated in Fig.~\ref{fig:grokcaption}, qualitative results further highlight this efficacy: while vanilla BSR only causes minor attribute errors, LI-Boost-BSR triggers a complete semantic collapse, forcing the model to generate entirely irrelevant descriptions. Similar results on Claude-o presented in the supplementary material confirm that LI-Boost consistently induces profound semantic deviations where baselines fail. These findings expose security vulnerabilities in current VLMs, underscoring the need for more robust alignment.

\begin{figure}[tbp]
\begin{promptbox}{}
    \textbf{Task} \\
    Please score the semantic deviation between the Ground Truth (GT) captions and Adversarial (Adv) captions to evaluate the Vision-Language Model (VLM) robustness.
    \vspace{\baselineskip}
    
    \textbf{Data} \\
    - Ground Truth: \enquote{\{clean\_caption\}} \\
    - Adversarial: \enquote{\{adv\_caption\}}
     \vspace{\baselineskip}
     
    \textbf{Rubric (Priority: Objects $>$ Actions $>$ Attributes $>$ Background)}
    \begin{itemize}
        \item {\color{Score10}\textbf{10:} Equivalent; no information loss.}
        \item {\color{Score89}\textbf{8-9:} Minor deviation; Primary Objects and Actions are correct, but with Attribute or Background errors.}
        \item {\color{Score57}\textbf{5-7:} Semantic shift; Related context but contains incorrect Objects or Action errors.}
        \item {\color{Score24}\textbf{2-4:} Hallucination; The scene is largely unrelated or fabricated.}
        \item {\color{Score1}\textbf{1:} Failure; Nonsensical or gibberish output.}
    \end{itemize}
    \vspace{\baselineskip}
\textbf{Calibration}
\\
        -\textit{Example 1:}  
        \\
        \textbf{Ground Truth:} ``Black cat on wooden table'' $\rightarrow$ \textbf{Adversarial:} ``Black cat on table'' $\rightarrow$ \textbf{Score: 9}, Reason: ``attribute missing: wooden'' \\
        -\textit{Example 2:}\\
        \textbf{Ground Truth:} ``Man playing guitar'' $\rightarrow$ \textbf{Adversarial:} ``Woman playing violin'' $\rightarrow$ \textbf{Score: 6}, Reason: ``object error: man/guitar''
    \vspace{\baselineskip}
    
    \textbf{Output Requirement} \\
    Return ONLY a JSON object. No other text.
    \begin{quote}
    \texttt{\{ \\
        \hspace*{1em} "score": <int>, \\
        \hspace*{1em} "reason": "<string>" \\
        \}}
    \end{quote}
\end{promptbox}
\vspace{-.2cm}
\caption{Prompt for image captioning task.}
\label{fig:promptimagecaption}
\vspace{-.6cm}
\end{figure}

\begin{table}[ht]
\centering
\footnotesize 
\setlength{\tabcolsep}{1.5pt} 
\captionsetup{labelsep=newline, justification=centering}
\caption{\textsc{Attack Success Rates (\%) of Different Adversarial Methods against Various Commercial VLMs on Image Captioning Task. The best results are  \textbf{bold} while methods incorporating LI-Boost are highlighted in \colorbox{gray!10}{gray}. The surrogate model is RN-50.}}
\label{table_asr_results}
\begin{tabular}{c|cccccc|c} 
\hline
\textbf{Method} & \textbf{GPT-4o} & \textbf{GPT-5.2} & \textbf{Gemini} & \textbf{Claude-s} & \textbf{Claude-o} & \textbf{Grok} & \textbf{Avg.} \\
\hline
MI-FGSM & 22.0 & 36.3 & 19.0 & 42.7 & 29.0 & 36.0 & 30.8 \\
\rowcolor{gray!10}\textbf{LI-Boost-MI}  & \textbf{27.8} & \textbf{42.1} & \textbf{23.9} & \textbf{52.0} & \textbf{39.3} & \textbf{47.2} & \textbf{38.7} \\
\hline
BSR & 60.1 & 72.2 & 54.5 & 67.8 & 73.1 & 70.8 & 66.4 \\
\rowcolor{gray!10}\textbf{LI-Boost-BSR} & \textbf{68.2}& \textbf{80.4} & \textbf{62.3} & \textbf{78.7} & \textbf{81.8} & \textbf{84.3} & \textbf{76.0} \\
\hline
ILA & 19.0 & 33.6& 16.9 & 35.7 & 23.6 & 29.5 & 26.4 \\
\rowcolor{gray!10}\textbf{LI-Boost-ILA}& \textbf{30.6}& \textbf{47.4} & \textbf{26.5} & \textbf{47.3} & \textbf{41.3} & \textbf{44.6} & \textbf{39.6} \\
\hline
BPA  & 51.6 & 66.4 & 45.3 & 63.4 & 65.9 & 68.9 & 60.3 \\
\rowcolor{gray!10}\textbf{LI-Boost-BPA} & \textbf{57.9} & \textbf{72.4} & \textbf{53.2} & \textbf{72.6}  & \textbf{74.1} & \textbf{80.3} & \textbf{68.4} \\
\hline
\end{tabular}
\vspace{-.2cm}
\end{table}

\begin{figure}[htbp]
    \centering
    \footnotesize 
    \setlength{\tabcolsep}{2pt} 
    \renewcommand{\arraystretch}{1.5}

    \begin{tabular}{ccc}
        \textbf{Clean} & \textbf{BSR} & \textbf{LI-Boost-BSR} \\
        
        \includegraphics[width=0.31\linewidth]{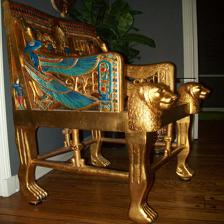} &
        \includegraphics[width=0.31\linewidth]{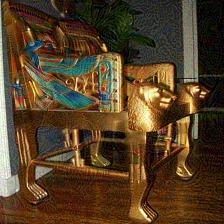} &
        \includegraphics[width=0.31\linewidth]{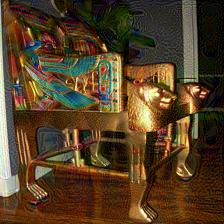} \\
        
        \begin{minipage}[t]{0.31\linewidth}
            \centering \textbf{\textcolor{green!50!black}{Golden throne with lion legs, peacock backrest, and intricate Egyptian-style inlays.}}
        \end{minipage} &
        \begin{minipage}[t]{0.31\linewidth}
            \centering \textbf{\textcolor{gray}{Ornate gold chair with lion-paw feet and vibrant peacock-motif backrest.}}
        \end{minipage} &
        \begin{minipage}[t]{0.31\linewidth}
            \centering \textbf{\textcolor{red}{Colorful, iridescent glass-like box topped with a shiny golden panther sculpture on a glass surface.}}
        \end{minipage} \\
    \end{tabular}

    \caption{Visualizations of the clean image and its adversarial counterparts with captions generated by Grok. \textbf{\textcolor{green!50!black}{Green text}} denotes the ground truth or benign description. \textbf{\textcolor{gray}{Gray text}} represents the result where BSR fails to induce significant semantic change. \textbf{\textcolor{red}{Red text}} highlights the significant semantic deviation achieved by LI-Boost-BSR.}
    \label{fig:grokcaption}
    \vspace{-.5cm}
\end{figure}

\subsection{Ablation Studies}

To gain deeper insights into LI-Boost, we conduct a series of ablation experiments to study the impact of hyperparameters, \textit{i.e.}, the random sampling distribution, the number of sampled perturbations $N$, and the upper bound of translated pixels $k$. All the adversarial examples are generated on RN-50. The default setting is $N=30$, $k=6$, and a Logarithmic distribution for sampling. 

\textbf{On the sampling distribution.} 
Intuitively, local invariance within smaller neighborhoods holds greater significance than that in larger neighborhoods. Consequently, the choice of sampling distribution plays a critical role. To explore the impact of sampling distribution, we employ three distinct distributions as illustrated in Fig.~\ref{fig:probability_density}, and the details are presented in the supplementary material. As shown in Fig.~\ref{fig:ablation_sampling_res}, Uniform distribution yields the weakest performance, as it fails to differentiate among translated pixels. Nevertheless, it substantially surpasses MI-FGSM, highlighting the superiority of LI-Boost. Both Normal and Logarithmic distributions achieve better attack performance since they assign various levels of importance to different translated pixels. However, there is still a decrease in white-box efficacy when utilizing the Normal distribution. The Logarithmic distribution achieves the best empirical performance among the three compared distributions, as it strikes a balance between emphasizing smaller neighborhoods and maintaining coverage of larger translated pixels compared to the Normal distribution.


\textbf{On the number of sampled perturbations $N$.} 
We test LI-Boost-MI with various $N$ to analyze its impact on attack performance. As shown in Fig.~\ref{fig:ablation_a_N}, the attack performance is significantly boosted with larger $N$ but exhibits diminishing returns beyond $N=30$. Considering the growth of computational cost from gradient computations as shown in Eq.~\eqref{gradient_calculate}, we empirically select $N=30$ in our experiments.

\textbf{On the upper bound of translated pixels $k$.} We conduct LI-Boost-MI using various $k$, \ie, upper bounds of translated pixels, to explore its impact on adversarial attack performance. Fig.~\ref{fig:ablation_b_k} shows that even with $k$ = 1, our method already outperforms MI-FGSM, highlighting its superior transferability. The attack performance reaches its peak at $k$ = 6, suggesting that local invariance significantly enhances adversarial transferability. However, performance diminishes with excessively large $k$, as it becomes increasingly difficult to craft effective perturbations within a broader search space. Consequently, we set $k$ = 6 to strike an optimal balance between white-box attack strength and black-box transferability.

\begin{figure}[t]
    \centering
    \setcounter{subfigure}{0}
    \small
    \captionsetup[subfigure]{
        font=footnotesize,
        labelfont=normalfont,
        textfont=normalfont,
        justification=justified,
        singlelinecheck=false
    }
    \begin{subfigure}[t]{0.48\linewidth}
        \centering
        \begin{tikzpicture}[trim axis left, trim axis right]
            \begin{axis}[
                width=1.2\linewidth,
                height=1.56\linewidth,
                grid=both,
                minor tick num=1,
                minor grid style={gray!25, dashed},
                major grid style={gray!25, dashed},
                xlabel={Translated pixels},
                ylabel={Probability},
                ylabel style={font=\scriptsize, yshift=-16pt},
                xlabel style={font=\scriptsize, yshift=5pt},
                tick label style={font=\scriptsize},
                xtick distance=1,
                yticklabel style={/pgf/number format/fixed},
                ymin=0, ymax=0.48,
                xmin=0.8, xmax=6.2,
                xtick={1,2,3,4,5,6},
                enlargelimits=false,
                clip=true,
                legend style={
                    at={(1, 1)},
                    anchor=north east,
                    font=\scriptsize,
                    fill=none,
                    inner sep=2pt,
                    column sep=3pt,
                    cells={anchor=west}
                },
                legend image post style={scale=0.8}
            ]
                \addplot+[
                    mark=*,
                    mark options={fill=Blue, solid},
                    mark size=1.8pt,
                    color=Blue,
                    line width=0.8pt,
                    each nth point=1
                ] coordinates {
                    (1, 0.1667) (2, 0.1667) (3, 0.1667) (4, 0.1667) (5, 0.1667) (6, 0.1667)
                };
                \addlegendentry{Uniform}

                \addplot+[
                    mark=triangle*,
                    mark options={fill=Orange},
                    mark size=1.8pt,
                    color=Orange,
                    line width=0.8pt
                ] coordinates {
                    (1,0.44) (2,0.30) (3,0.15) (4,0.06) (5,0.02) (6,0.01)
                };
                \addlegendentry{Normal}

                \addplot+[
                    mark=square*,
                    mark options={fill=Purple},
                    mark size=1.8pt,
                    color=Purple,
                    line width=0.8pt
                ] coordinates {
                    (1,0.39) (2,0.27) (3,0.16) (4,0.11) (5,0.05) (6,0.02)
                };
                \addlegendentry{Logarithmic}
            \end{axis}
        \end{tikzpicture}
        \caption{The probability distribution of adopted Uniform, Normal and Logarithmic distributions.}
        \label{fig:probability_density}
    \end{subfigure}
    \hfill
    \begin{subfigure}[t]{0.48\linewidth}
        \centering
        \begin{tikzpicture}[trim axis left, trim axis right]
            \begin{axis}[
                ybar,
                bar width=2.3pt,
                bar shift=-0.06cm,
                bar shift=-0.03cm,
                bar shift=+0.00cm,
                bar shift=+0.03cm,
                ylabel={Attack success rates (\%)},
                ylabel style={font=\scriptsize, yshift=-8pt},
                width=1.29\linewidth,
                height=1.5\linewidth,
                xtick={0,1,2,3,4,5,6,7,8},
                xtick style={draw=none},   
                xticklabels={RN-50,Inc-v3,MN-v3,DN-121,FSNet,ViT,PiT,Visformer,Swin},
                ymajorgrids=true,
                grid style=dashed,
                xticklabel style={rotate=45, anchor=east, xshift=4pt, font=\scriptsize},
                yticklabel style={font=\scriptsize},
                xmin=0.1, xmax=8,
                ymin=0, ymax=100,
                enlarge x limits=0.1, 
                legend style={
                    at={(0.205, 0.911)},
                    anchor=west,
                    legend columns=2,
                    font=\scriptsize,
                    /tikz/every even column/.append style={column sep=0.1cm},
                    cells={anchor=center},
                    fill=none,
                    inner sep=2pt
                },
                legend image code/.code={
                    \draw[#1, draw=none] (0cm,0cm) rectangle (0.15cm,0.06cm);
                },
                ylabel style={font=\scriptsize, yshift=-12pt},
                tick label style={font=\scriptsize},
                clip mode=individual, 
                trim axis left, trim axis right
            ]
                
            \addplot[fill=Pink, draw=none, bar shift=-0.16cm] 
                coordinates {(0,94.9) (1,34.5) (2,40.6) (3,45.9) (4,25.2) (5,10.5) (6,18.0) (7,23.1) (8,27.8)};
            \addlegendentry{MI-FGSM}
            
            \addplot[fill=Blue, draw=none, bar shift=-0.08cm] 
                coordinates {(0,84.2) (1,39.4) (2,50.2) (3,53.5) (4,34.3) (5,14.8) (6,23.4) (7,32.3) (8,36.2)};
            \addlegendentry{Uniform}
            
            \addplot[fill=Orange, draw=none, bar shift=0cm] 
                coordinates {(0,92.2) (1,41.6) (2,52.7) (3,58.0) (4,38.1) (5,16.8) (6,26.4) (7,35.4) (8,38.3)};
            \addlegendentry{Normal}
            
            \addplot[fill=Purple, draw=none, bar shift=0.08cm] 
                coordinates {(0,97.1) (1,45.3) (2,55.9) (3,61.9) (4,41.8) (5,19.1) (6,29.6) (7,38.9) (8,41.4)};
            \addlegendentry{Logarithmic}
            \end{axis}
        \end{tikzpicture}
        \caption{Attack success rates (\%) of MI-FGSM and LI-Boost-MI with three distributions on nine models.}
        \label{fig:ablation_sampling_res}
    \end{subfigure}
    \caption{Ablation studies of various sampling distributions.}
    \vspace{-.3cm}
\end{figure}

\begin{figure}[tb]
    \centering
    \captionsetup[subfigure]{width=0.95\textwidth}
    \captionsetup[subfigure]{
        font=footnotesize,
        labelfont=normalfont,
        textfont=normalfont,
        justification=centering,
        singlelinecheck=false,
        margin={.4cm, 0cm}
    }
    \hspace{-1cm}
    \begin{minipage}[t]{0.42\linewidth}
        \subfloat[The hyper-parameter $N$]{
            \begin{tikzpicture}[clip]
                \begin{axis}[
                    width=1.322\linewidth,
                    height=1.8\linewidth,
                    grid=both,
                    minor grid style={gray!25, dashed},
                    major grid style={gray!25, dashed},
                    xlabel={Number of sampled perturbations $N$}, 
                    ylabel={Attack success rates (\%)},
                    ylabel style={font=\scriptsize, xshift=0pt, yshift=-0.5cm},
                    xlabel style={font=\scriptsize, yshift=5pt},
                    tick label style={font=\scriptsize},
                    xtick distance=8,
                    ymin=13, ymax=101,
                    xmin=3, xmax=42,
                    xtick={5,10,15,20,25,30,35,40},
                   legend style={at={(0.5, 0.65)}, anchor=south, legend columns=3,inner xsep=0.7pt, inner ysep=0.8pt,  align=center, font=\tiny, minimum width=0.01cm, column sep=0.01cm, row sep=0.005cm, fill opacity=0.5, /tikz/every even column/.append style={column sep=0.01cm}},
                    legend image post style={scale=0.3},
                ]
                \addplot[line width=0.8pt, solid, mark=*, color=resnet50, mark options={mark size=1.5pt}] 
                    table[x=num, y=ResNet-50, col sep=comma]{figs/data/ablation_samplenumber.csv};
                \addlegendentry{RN-50}
                \addplot[line width=0.8pt, solid, mark=square*, color=incv3, mark options={mark size=1.5pt}] 
                    table[x=num, y=Inc-v3, col sep=comma]{figs/data/ablation_samplenumber.csv};
                \addlegendentry{Inc-v3}
                \addplot[line width=0.8pt, solid, mark=triangle*, color=mobi, mark options={mark size=1.5pt}] 
                    table[x=num, y=MobileNet, col sep=comma]{figs/data/ablation_samplenumber.csv};
                \addlegendentry{MN-v3}
                \addplot[line width=0.8pt, solid, mark=diamond*, color=dense, mark options={mark size=1.5pt}] 
                    table[x=num, y=DenseNet, col sep=comma]{figs/data/ablation_samplenumber.csv};
                \addlegendentry{DN-121}
                 \addplot[line width=0.8pt, solid, mark=pentagon*, color=conv, mark options={mark size=1.5pt}] 
                    table[x=num, y=FSNet, col sep=comma]{figs/data/ablation_samplenumber.csv};
                \addlegendentry{FSNet}
                \addplot[line width=0.8pt, solid, mark=10-pointed star, color=vit, mark options={mark size=1.5pt}] 
                    table[x=num, y=ViT, col sep=comma]{figs/data/ablation_samplenumber.csv};
                \addlegendentry{ViT}
                \addplot[line width=0.8pt, solid, mark=Mercedes star flipped, color=pit, mark options={mark size=2.2pt, line width=1.0pt}] 
                    table[x=num, y=PiT, col sep=comma]{figs/data/ablation_samplenumber.csv};
                \addlegendentry{PiT}
                \addplot[line width=0.8pt, solid, mark=star, color=visformer, mark options={mark size=1.9pt, line width=1.0pt}] 
                    table[x=num, y=Visformer, col sep=comma]{figs/data/ablation_samplenumber.csv};
                \addlegendentry{Visformer}
                \addplot[line width=0.8pt, solid, mark=Mercedes star, color=swin, mark options={mark size=2.2pt, line width=1.0pt}] 
                    table[x=num, y=Swin, col sep=comma]{figs/data/ablation_samplenumber.csv};
                \addlegendentry{Swin}
            \end{axis}
            \end{tikzpicture}
            \label{fig:ablation_a_N}
        }
    \end{minipage}
    \hspace{.6cm}
    \begin{minipage}[t]{0.42\linewidth}
        \subfloat[The hyper-parameter $k$]{
            \begin{tikzpicture}[clip]
                \begin{axis}[
                    width=1.322\linewidth,
                    height=1.8\linewidth,
                    grid=both,
                    minor grid style={gray!25, dashed},
                    major grid style={gray!25, dashed},
                    xlabel={Upper bound of translated pixels $k$},
                    ylabel={Attack success rates (\%)},
                    ylabel style={font=\scriptsize, xshift=0cm, yshift=-0.6cm},
                    xlabel style={font=\scriptsize, yshift=5pt},
                    tick label style={font=\scriptsize},
                    xtick distance=8,
                    ymin=12, ymax=101,
                    xmin=0, xmax=11,
                    xtick={1,2,3,4,5,6,7,8,9,10},
                    legend style={at={(0.5, 0.65)}, anchor=south, legend columns=3,inner xsep=0.65pt, inner ysep=0.8pt,  align=center, font=\tiny, minimum width=0.01cm, column sep=0.01cm, row sep=0.005cm, fill opacity=0.5, /tikz/every even column/.append style={column sep=0.01cm}},
                    legend image post style={scale=0.3},
                ]
                \addplot[line width=0.8pt, solid, mark=*, color=resnet50, mark options={mark size=1.5pt}] 
                    table[x=k, y=ResNet-50, col sep=comma]{figs/data/ablation_move_pixels.csv};
                \addlegendentry{RN-50}
                \addplot[line width=0.8pt, solid, mark=square*, color=incv3, mark options={mark size=1.5pt}] 
                    table[x=k, y=Inc-v3, col sep=comma]{figs/data/ablation_move_pixels.csv};
                \addlegendentry{Inc-v3}
                \addplot[line width=0.8pt, solid, mark=triangle*, color=mobi, mark options={mark size=1.5pt}] 
                    table[x=k, y=MobileNet, col sep=comma]{figs/data/ablation_move_pixels.csv};
                \addlegendentry{MN-v3}
                \addplot[line width=0.8pt, solid, mark=diamond*, color=dense, mark options={mark size=1.5pt}] 
                    table[x=k, y=DenseNet, col sep=comma]{figs/data/ablation_move_pixels.csv};
                \addlegendentry{DN-121}
                \addplot[line width=0.8pt, solid, mark=pentagon*, color=conv, mark options={mark size=1.5pt}] 
                    table[x=k, y=FSNet, col sep=comma]{figs/data/ablation_move_pixels.csv};
                \addlegendentry{FSNet}
                \addplot[line width=0.8pt, solid, mark=10-pointed star, color=vit, mark options={mark size=1.5pt}] 
                    table[x=k, y=ViT, col sep=comma]{figs/data/ablation_move_pixels.csv};
                \addlegendentry{ViT}
                \addplot[line width=0.8pt, solid, mark=Mercedes star flipped, color=pit, mark options={mark size=2.2pt, line width=1.0pt}] 
                    table[x=k, y=PiT, col sep=comma]{figs/data/ablation_move_pixels.csv};
                \addlegendentry{PiT}
                \addplot[line width=0.8pt, solid, mark=star, color=visformer, mark options={mark size=1.9pt, line width=1.0pt}] 
                    table[x=k, y=Visformer, col sep=comma]{figs/data/ablation_move_pixels.csv};
                \addlegendentry{Visformer}
                \addplot[line width=0.8pt, solid, mark=Mercedes star, color=swin, mark options={mark size=2.2pt, line width=1.0pt}] 
                    table[x=k, y=Swin, col sep=comma]{figs/data/ablation_move_pixels.csv};
                \addlegendentry{Swin}
            \end{axis}
            \end{tikzpicture}
            \label{fig:ablation_b_k}
        }
    \end{minipage}
    \caption{Attack success rates (\%) on nine models with various hyper-parameters $N$ and $k$. The adversarial examples are generated by LI-Boost-MI on RN-50.}
    \label{fig:ablation_K}
    \vspace{-.6cm}
\end{figure}
\subsection{Comparisons and Differences with TIM}
Translation-Invariant Method (TIM)~\cite{dong2019TIM} is a closely related work that boosts adversarial transferability. There are many differences between our method and TIM.

\textbf{TIM} attributes the transferability failure to the fact that adversarial examples are tightly coupled to the discriminative regions of surrogate models. Since different models attend to different regions, the adversarial examples transfer poorly. To mitigate this problem, TIM optimizes the following objective:
\begin{equation}
    \delta^* = \arg\max_{||\delta||_p \le \epsilon} \sum_{i,j} w_{ij} (J\left(\Gamma(x+\delta,i,j), y; \theta)\right),
    \label{eq:tim_optimize}
\end{equation}where $\Gamma(x+\delta,i,j)$ translates $x+\delta$ by $i$ and $j$ pixels along the two-dimensions and $w_{ij}$ is the weight for loss. To solve this objective, TIM exploits the translation invariance~\cite{lecun1998convolutional} prior of convolutional neural networks and \textit{\textbf{smooths the gradients using a pre-defined kernel}}.

\textbf{LI-Boost} instead attributes the failure to the perturbation itself. The perturbation $\delta$ lacks the local invariance that natural images possess, collapsing under small spatial shifts, whereas natural images retain their semantics under the same shift. To inject natural-image-like spatial stability into the perturbation, LI-Boost therefore \textit{\textbf{fixes the clean image $x$ and translates only $\delta$, then averaging gradients over its local neighborhood}}:
\begin{equation}
    \delta^* = \mathop{\arg\max}\limits_{||\delta||_{p}\leq \epsilon} \mathbb{E}_{\delta' \in \mathcal{N}_k(\delta)} \left[J(x + \delta', y; \theta)\right],
\end{equation}
where $\mathcal{N}_k(\delta)$ denotes the spatial neighborhood of $\delta$ under log-distributed shifts, and $x$ remains unchanged. Consequently, TIM and LI-Boost differ substantially in both motivation, optimization objective and methodology.

To further validate the differences between TIM and LI-Boost empirically, we also report results in Tab.~\ref{tab:tim_comparison}. It can be seen that LI-Boost-TIM preserves the white-box attack effectiveness and leverages the transferability with considerable margins, confirming the superiority of LI-Boost, the differences with TIM and the two priors are complementary.

\begin{table*}[htbp]
\centering
\caption{Attack success rates (\%) of \textbf{TIM} w/o LI-Boost on nine models. For the single-model settings, the adversarial examples are crafted on RN-50, Inc-v3, MN-v3, DN-121, FSNet, ViT, PiT, Visformer, and Swin, respectively. For the ensemble setting, RN-50, Inc-v3, MN-v3, and DN-121 are adopted as surrogate models. * indicates the white-box model.}
\resizebox{\linewidth}{!}{
\begin{tabular}{ccccccccccc}
\hline
\textbf{Surrogate Model} & \textbf{Attack} &~~\textbf{RN-50}& \textbf{Inc-v3} & \textbf{MN-v3} & \textbf{DN-121} & \textbf{FSNet} & \textbf{ViT} & \textbf{PiT} & \textbf{Visformer} & \textbf{Swin} \\
\hline

\multirow{2}{*}{\textbf{RN-50}} 
& TIM & ~~82.8* & 35.0 & 40.4 & 50.3 & 27.4 & 12.7 & 18.2 & 23.8 & 29.0 \\
& \cellcolor{gray!15}\textbf{LI-Boost-TIM} & \cellcolor{gray!15}~~\textbf{88.4*} & \cellcolor{gray!15}\textbf{36.7} & \cellcolor{gray!15}\textbf{44.0} & \cellcolor{gray!15}\textbf{55.1} & \cellcolor{gray!15}\textbf{35.4} & \cellcolor{gray!15}\textbf{16.0} & \cellcolor{gray!15}\textbf{21.5} & \cellcolor{gray!15}\textbf{29.8} & \cellcolor{gray!15}\textbf{34.1} \\
\hline
\multirow{2}{*}{\textbf{Inc-v3}} 
& TIM & ~~37.9 & 99.5* & 48.0 & 57.1 & 31.8 & 15.8 & 21.1 & 26.6 & 33.7 \\
& \cellcolor{gray!15}\textbf{LI-Boost-TIM} & \cellcolor{gray!15}~~\textbf{44.2} & \cellcolor{gray!15}\textbf{99.5*} & \cellcolor{gray!15}\textbf{53.0} & \cellcolor{gray!15}\textbf{63.7} & \cellcolor{gray!15}\textbf{37.5} & \cellcolor{gray!15}\textbf{17.8} & \cellcolor{gray!15}\textbf{23.3} & \cellcolor{gray!15}\textbf{31.6} & \cellcolor{gray!15}\textbf{38.0} \\
\hline
\multirow{2}{*}{\textbf{MN-v3}} 
& TIM & ~~45.5 & 47.3 & 99.6* & 66.8 & 35.5 & 22.3 & 28.0 & 40.4 & 48.0 \\
& \cellcolor{gray!15}\textbf{LI-Boost-TIM} & ~~\cellcolor{gray!15}\textbf{54.6} & \cellcolor{gray!15}\textbf{50.4} & \cellcolor{gray!15}\textbf{99.7*} & \cellcolor{gray!15}\textbf{74.1} & \cellcolor{gray!15}\textbf{43.0} & \cellcolor{gray!15}\textbf{26.1} & \cellcolor{gray!15}\textbf{32.7} & \cellcolor{gray!15}\textbf{47.6} & \cellcolor{gray!15}\textbf{54.7} \\
\hline
\multirow{2}{*}{\textbf{DN-121}} 
& TIM & ~~57.8 & 52.5 & 62.9 & \textbf{100.0*} & 44.1 & 19.9 & 25.8 & 39.6 & 44.4 \\
& \cellcolor{gray!15}\textbf{LI-Boost-TIM} & \cellcolor{gray!15}~~\textbf{65.0} & \cellcolor{gray!15}\textbf{55.4} & \cellcolor{gray!15}\textbf{68.7} & \cellcolor{gray!15}\textbf{100.0*} & \cellcolor{gray!15}\textbf{51.9} & \cellcolor{gray!15}\textbf{21.7} & \cellcolor{gray!15}\textbf{29.1} & \cellcolor{gray!15}\textbf{45.5} & \cellcolor{gray!15}\textbf{49.2} \\
\hline
\multirow{2}{*}{\textbf{FSNet}} 
& TIM & ~~51.6 & 45.8 & 52.8 & 63.2 & 96.0* & 26.0 & 34.3 & 46.1 & 51.3 \\
& \cellcolor{gray!15}\textbf{LI-Boost-TIM} & \cellcolor{gray!15}~~\textbf{61.7} & \cellcolor{gray!15}\textbf{50.0} & \cellcolor{gray!15}\textbf{58.5} & \cellcolor{gray!15}\textbf{71.7} & \cellcolor{gray!15}\textbf{96.5*} & \cellcolor{gray!15}\textbf{31.6} & \cellcolor{gray!15}\textbf{41.6} & \cellcolor{gray!15}\textbf{56.9} & \cellcolor{gray!15}\textbf{59.1} \\
\hline
\multirow{2}{*}{\textbf{ViT}} 
& TIM & ~~35.0 & 36.5 & 45.7 & 49.7 & 32.6 & 86.8* & 33.1 & 36.8 & 44.3  \\
& \cellcolor{gray!15}\textbf{LI-Boost-TIM} & \cellcolor{gray!15}~~\textbf{37.4} & \cellcolor{gray!15}\textbf{37.0} & \cellcolor{gray!15}\textbf{46.4} & \cellcolor{gray!15}\textbf{50.8} & \cellcolor{gray!15}\textbf{35.9} & \cellcolor{gray!15}\textbf{88.0*} & \cellcolor{gray!15}\textbf{37.2} & \cellcolor{gray!15}\textbf{41.0} & \cellcolor{gray!15}\textbf{48.6} \\
\hline
\multirow{2}{*}{\textbf{PiT}} 
& TIM & ~~36.0 & 37.9 & 45.9 & 49.2 & 32.5 & 24.9 & 88.9* & 38.8 & 45.4 \\
& \cellcolor{gray!15}\textbf{LI-Boost-TIM} & \cellcolor{gray!15}~~\textbf{40.0} & \cellcolor{gray!15}\textbf{38.5} & \cellcolor{gray!15}\textbf{48.3} & \cellcolor{gray!15}\textbf{52.1} & \cellcolor{gray!15}\textbf{37.8} & \cellcolor{gray!15}\textbf{30.3} & \cellcolor{gray!15}\textbf{90.5*} & \cellcolor{gray!15}\textbf{45.1} & \cellcolor{gray!15}\textbf{50.8} \\
\hline
\multirow{2}{*}{\textbf{Visformer}} 
& TIM & ~~47.5 & 45.0 & 56.0 & 61.3 & 47.5 & 32.4 & 44.8 & \textbf{95.8*} & 58.3 \\
& \cellcolor{gray!15}\textbf{LI-Boost-TIM} & \cellcolor{gray!15}~~\textbf{51.8} & \cellcolor{gray!15}\textbf{45.5} & \cellcolor{gray!15}\textbf{59.3} & \cellcolor{gray!15}\textbf{65.0} & \cellcolor{gray!15}\textbf{52.8} & \cellcolor{gray!15}\textbf{36.9} & \cellcolor{gray!15}\textbf{50.6} & \cellcolor{gray!15}95.2* & \cellcolor{gray!15}\textbf{62.9} \\
\hline
\multirow{2}{*}{\textbf{Swin}} 
& TIM & ~~44.6 & 43.2 & 57.2 & 60.2 & 44.5 & 31.6 & 40.8 & 50.9 & 97.0* \\
& \cellcolor{gray!15}\textbf{LI-Boost-TIM} & \cellcolor{gray!15}~~\textbf{51.9} & \cellcolor{gray!15}\textbf{46.6} & \cellcolor{gray!15}\textbf{62.4} & \cellcolor{gray!15}\textbf{66.3} & \cellcolor{gray!15}\textbf{52.0} & \cellcolor{gray!15}\textbf{38.9} & \cellcolor{gray!15}\textbf{50.7} & \cellcolor{gray!15}\textbf{60.4} & \cellcolor{gray!15}\textbf{97.4*} \\
\hline

\multirow{2}{*}{\textbf{Ensemble}} 
& ~~TIM$_{ens}$ & ~~90.2* & \textbf{97.2*} & 97.0* & \textbf{100.0*} & 68.6 & 38.8 & 49.2 & 65.4 & 67.8 \\
& \cellcolor{gray!15}\textbf{LI-Boost-TIM$_{ens}$} & \cellcolor{gray!15}~~\textbf{91.5*} & \cellcolor{gray!15}96.8* & \cellcolor{gray!15}\textbf{97.2*} & \cellcolor{gray!15}\textbf{100.0*} & \cellcolor{gray!15}\textbf{74.3} & \cellcolor{gray!15}\textbf{43.0} & \cellcolor{gray!15}\textbf{53.5} & \cellcolor{gray!15}\textbf{70.6} & \cellcolor{gray!15}\textbf{73.0} \\
\hline
\end{tabular}
}
\label{tab:tim_comparison}
\vspace{-0.5em}
\end{table*}

\subsection{Equal-Compute Comparison with MI-FGSM}

\begin{figure}[t]
  \centering
  \hspace{.8cm}
  \begin{tikzpicture}[trim axis left, trim axis right]
      \begin{axis}[
          ybar,
          bar width=5.0pt,
          ylabel={Attack Success Rates (\%)},
          ylabel style={font=\scriptsize, yshift=-12pt},
          width=\linewidth,
          height=0.55\linewidth,
          xtick={0,1,2,3,4,5,6,7,8},
          xtick style={draw=none},
          xticklabels={RN-50,Inc-v3,MN-v3,DN-121,FSNet,ViT,PiT,Visformer,Swin},
          ytick={0,20,40,60,80,100},
          ymajorgrids=true,
          grid style=dashed,
          xticklabel style={rotate=45, anchor=east, xshift=4pt, font=\scriptsize},
          yticklabel style={font=\scriptsize},
          xmin=0.0, xmax=8.0,
          ymin=0, ymax=100,
          enlarge x limits=0.08,
          legend style={
              at={(0.27, 1.0)},
              anchor=north west,
              legend columns=3,
              font=\scriptsize,
              /tikz/every even column/.append style={column sep=0.1cm},
              cells={anchor=center},
              fill=none,
              inner sep=2pt
          },
          legend image code/.code={
              \draw[#1, draw=none] (0cm,0cm) rectangle (0.15cm,0.06cm);
          },
          clip mode=individual,
          trim axis left, trim axis right
      ]

      \addplot[fill=mibase, draw=none, bar shift=-0.18cm]
          coordinates {(0,94.9) (1,34.5) (2,40.6) (3,45.9) (4,25.2) (5,10.7) (6,18.0) (7,23.1) (8,27.8)};
      \addlegendentry{MI (baseline)}

      \addplot[fill=mi300, draw=none, bar shift=0cm]
          coordinates {(0,95.4) (1,30.4) (2,41.2) (3,45.9) (4,25.6) (5,10.6) (6,17.3) (7,23.1) (8,27.9)};
      \addlegendentry{MI (300 iters)}

      \addplot[fill=liboost, draw=none, bar shift=0.18cm]
          coordinates {(0,97.0) (1,45.3) (2,55.2) (3,62.0) (4,41.8) (5,19.1) (6,29.1) (7,38.7) (8,41.9)};
      \addlegendentry{LI-Boost-MI}

      \end{axis}
  \end{tikzpicture}
  \caption{Attack success rates (\%) under equal-compute comparisons.
  MI (300 iters) matches the total gradient evaluations of LI-Boost-MI
  ($N\!=\!30$, 10 iterations). The adversarial examples are generated on RN-50.}
  \vspace{-.5cm}
  \label{fig:equal_compute}
  \end{figure}
To verify that the gains of LI-Boost stem from the local invariance objective rather than the additional gradient evaluations, we provide an equal-compute comparison. We choose MI-FGSM as the baseline for this comparison because it introduces no auxiliary mechanism beyond momentum accumulation, making it the cleanest setting to isolate the effect of additional gradient evaluations from the local invariance objective. MI-FGSM (300 iters) runs 300 gradient evaluations per example, matching the total computation of LI-Boost-MI ($N=30$, 10 iterations). All experiments use RN-50 as the surrogate model. 

As shown in Fig.~\ref{fig:equal_compute}, increasing MI-FGSM from 10 to 300 iterations yields negligible improvement or even slight degradation on black-box targets (\eg, Inc-v3 drops from 34.5\% to 30.4\%), consistent with the well-known overfitting phenomenon where excessive iterations cause the perturbation to overfit the surrogate model and lose transferability. In contrast, LI-Boost-MI achieves substantially higher attack success rates across all nine target models under the same compute budget, confirming that the performance gains are attributable to the local invariance objective itself rather than additional gradient computational costs.

\subsection{Evaluation Across Varying Perturbation Budgets}
\begin{figure}[t]
  \centering
  \begin{tikzpicture}
  \begin{axis}[
      width=\linewidth,
      height=0.55\linewidth,
      xlabel={Perturbation Budget $\epsilon$ ($\times 1/255$)},
      ylabel={Attack Success Rates (\%)},
      xlabel style={font=\scriptsize},
      ylabel style={font=\scriptsize, yshift=-5pt},
      tick label style={font=\scriptsize},
      xtick={4,6,8,10,12,14},
      xticklabel style={anchor=north},
      enlarge x limits=0.12,
      ymin=10, ymax=50,
      ytick={10,20,30,40,50},
      ymajorgrids=true,
      grid style=dashed,
      legend style={
          at={(0, 1.0)},
          anchor=north west,
          font=\scriptsize,
          fill=none,
          inner sep=1.5pt
      },
  ]

  \addplot[
      color=blueperturb,
      mark=o,
      mark size=2pt,
      line width=1pt,
  ] coordinates {
      (4, 13.90) (6, 18.01) (8, 22.12) (10, 26.17) (12, 29.44) (14, 32.62)
  };
  \addlegendentry{MI-FGSM}

  \addplot[
      color=greenperturb,
      mark=square*,
      mark size=2pt,
      line width=1pt,
  ] coordinates {
      (4, 16.62) (6, 22.71) (8, 28.57) (10, 34.39) (12, 39.16) (14, 43.91)
  };
  \addlegendentry{LI-Boost-MI}

  \end{axis}
  \end{tikzpicture}
  \caption{Attack success rates (\%) of MI-FGSM and LI-Boost-MI under varying
  perturbation budgets, averaged over all nine normally trained models with RN-50 as
  the surrogate.}
  \vspace{-.1cm}
  \label{fig:epsilon_ablation}
\end{figure}
To verify that the gains of LI-Boost hold under smaller perturbation budgets, we evaluate MI-FGSM and LI-Boost-MI with $\epsilon$ ranging from $4/255$ to $14/255$, using RN-50 as the surrogate model. As illustrated in Fig.~\ref{fig:epsilon_ablation}, LI-Boost-MI consistently surpasses MI-FGSM at every budget, and the margin widens as $\epsilon$ grows. Notably, at $\epsilon = 8/255$, $10/255$, and $12/255$, LI-Boost-MI outperforms MI-FGSM at the next larger budget, \ie, LI-Boost effectively grants an additional $2/255$ of perturbation budget, suggesting that LI-Boost makes more effective use of the given budget rather than benefiting from larger perturbations.

\subsection{Orthogonality with Input Transformation-Based Methods}
Input transformation-based attacks all apply transformations to the composite input $x+\delta$, diversifying the \textbf{whole visual context} seen by the model. LI-Boost operates on the perturbation side, optimizing $\delta$ to possess locally invariant structure consistent with the spatial coherence of natural images. Therefore, LI-Boost is orthogonal to input transformation-based methods, and combining them consistently improves adversarial transferability.

\section{Further Discussions}

To further substantiate that LI-Boost improves adversarial transferability by enhancing local invariance, we conduct two complementary analyses: a direct measurement of local invariance and an evaluation of loss landscape flatness.

First, we quantify the local invariance ($k=6$) of five representative attacks w/wo LI-Boost, namely MI-FGSM, BSR,  BPA, ILA, and FPR. As shown in Tab.~\ref{tab:discussion}, LI-Boost simultaneously increases both local invariance and attack performance across all the baselines. The consistent co-improvement supports the interpretation that enhanced local invariance is associated with and contributes to the observed transferability gains.

Second, adversarial examples residing in flatter regions are known to transfer more effectively, as the perturbation remains effective under the distribution shift introduced by unseen target models~\cite{ge2023boosting}. We measure the zeroth-order average-case flatness~\cite{qiu2026boosting} as Eq.\ref{eq:r0}:
\begin{equation}
      {R}^{(0)}_\xi(x+\delta) = \mathbb{E}_{{x}' \sim \mathrm{Unif}({B_\xi})} [\left\| J({x}') - J({x+\delta}) \right\|_2],
    \label{eq:r0}
  \end{equation}
where ${B}_\xi$ denotes the $\ell_\infty$ ball of radius $\xi$ centered at the adversarial example $x+\delta$, approximated via Monte Carlo sampling~\cite{1949monte} with 20 probes at $\xi{=}32/255$, and $x'$ is a uniformly sampled neighbor of $x+\delta$ within $B_\xi$.

 \begin{table}[ht]
  \centering
  \caption{Average attack success rates on nine normally trained models(\%), local invariance ($k=6$),
  average-case flatness $R^{(0)}_\xi$, and time consumption (s/img) of various attacks w/wo LI-Boost on RN-50, except for FPR, which generates examples on ViT.}
  \label{tab:discussion}
  \resizebox{\linewidth}{!}{
  \begin{tabular}{c|c|ccccc}
  \hline
  \textbf{Metric} & \textbf{LI-Boost} & \textbf{MI-FGSM} & \textbf{BSR} & \textbf{BPA} & \textbf{ILA} & \textbf{FPR} \\
  \hline
  \multirow{2}{*}{ASR $\uparrow$}
  & \textcolor{purple}{\xmark}          & \textcolor{purple}{35.6} & \textcolor{purple}{83.7} & \textcolor{purple}{69.7} &
  \textcolor{purple}{32.5} & \textcolor{purple}{56.4} \\
  & \cellcolor{gray!10}\textcolor{green!50!black}{\cmark} & \cellcolor{gray!10}\textbf{\textcolor{green!50!black}{47.8}} &
  \cellcolor{gray!10}\textbf{\textcolor{green!50!black}{91.0}} & \cellcolor{gray!10}\textbf{\textcolor{green!50!black}{76.5}} &
  \cellcolor{gray!10}\textbf{\textcolor{green!50!black}{44.4}} & \cellcolor{gray!10}\textbf{\textcolor{green!50!black}{64.1}} \\
  \hline
  \multirow{2}{*}{LI $\uparrow$}
  & \textcolor{purple}{\xmark}          & \textcolor{purple}{0.24} & \textcolor{purple}{0.55} & \textcolor{purple}{0.36} &
  \textcolor{purple}{0.15} & \textcolor{purple}{0.30} \\
  & \cellcolor{gray!10}\textcolor{green!50!black}{\cmark} & \cellcolor{gray!10}\textbf{\textcolor{green!50!black}{0.41}} &
  \cellcolor{gray!10}\textbf{\textcolor{green!50!black}{0.92}} & \cellcolor{gray!10}\textbf{\textcolor{green!50!black}{0.56}} &
  \cellcolor{gray!10}\textbf{\textcolor{green!50!black}{0.34}} & \cellcolor{gray!10}\textbf{\textcolor{green!50!black}{0.42}} \\
  \hline
  \multirow{2}{*}{$R^{(0)}_\xi \downarrow$}
  & \textcolor{purple}{\xmark}          & \textcolor{purple}{0.51} & \textcolor{purple}{2.75} & \textcolor{purple}{2.08} &
  \textcolor{purple}{2.94} & \textcolor{purple}{2.67} \\
  & \cellcolor{gray!10}\textcolor{green!50!black}{\cmark} & \cellcolor{gray!10}\textbf{\textcolor{green!50!black}{0.49}} &
  \cellcolor{gray!10}\textbf{\textcolor{green!50!black}{2.53}} & \cellcolor{gray!10}\textbf{\textcolor{green!50!black}{1.84}} &
  \cellcolor{gray!10}\textbf{\textcolor{green!50!black}{2.88}} & \cellcolor{gray!10}\textbf{\textcolor{green!50!black}{2.28}} \\
  \hline
  \multirow{2}{*}{Time $\downarrow$}
  & \textcolor{purple}{\xmark}          & \textcolor{purple}{\textbf{0.37}} & \textcolor{purple}{\textbf{0.79}} & \textcolor{purple}{\textbf{0.38}} &
  \textcolor{purple}{\textbf{0.47}} & \textcolor{purple}{\textbf{0.36}} \\
  & \cellcolor{gray!10}\textcolor{green!50!black}{\cmark} & \cellcolor{gray!10}\textcolor{green!50!black}{2.97} &
  \cellcolor{gray!10}\textcolor{green!50!black}{15.21} & \cellcolor{gray!10}\textcolor{green!50!black}{4.10} &
  \cellcolor{gray!10}\textcolor{green!50!black}{2.14} & \cellcolor{gray!10}\textcolor{green!50!black}{2.97} \\
  \hline
  \end{tabular}
  }
  \end{table}

As reported in Tab.~\ref{tab:discussion}, LI-Boost consistently reduces $R^{(0)}_\xi$ across all five baselines, confirming that enhanced local invariance steers optimization toward flatter regions and thereby improves transferability. Note that $R^{(0)}_\xi$ measures absolute loss variation and is not normalized across attacks: methods such as BSR generate perturbations that induce larger absolute loss values, inflating their baseline $R^{(0)}_\xi$ relative to simpler attacks like MI-FGSM. \textbf{The meaningful flatness comparison is therefore within each column}, where LI-Boost yields a consistent reduction of $R^{(0)}_\xi$ against its own baseline. 

Besides, we also report the per-image runtime (s/img) of five representative methods. Timing measurements in this section are conducted on a single RTX 4090 GPU. From Tab.~\ref{tab:discussion}, we can observe that applying LI-Boost introduces additional computational overhead per iteration. Input transformation-based attacks (\eg, BSR) exhibit higher overhead, since their internal augmentation mechanisms naturally rely on repeated sampling. Despite this, the overhead remains within an acceptable range, making it a practical method for boosting adversarial transferability.

\section{Conclusions}
In this study, we introduce \textit{local invariance} of adversarial perturbations and empirically demonstrate a positive correlation between the local invariance of adversarial perturbations on \textit{a surrogate model} and their transferability \textit{across diverse victim models}. Building on this insight, we propose LI-Boost, a novel method designed to enhance the local invariance of adversarial perturbations on a single model for better adversarial transferability. Through extensive experiments conducted on the ImageNet dataset, we validate the effectiveness of LI-Boost across a variety of transfer-based attacks, including CNNs, ViTs, defense mechanisms, commercial vision API systems, and vision-language models. Our findings not only underscore the efficacy of the proposed approach but also provide valuable insights into potential avenues for advancing adversarial attacks. One limitation of LI-Boost is the computational cost introduced by sampling $N$ perturbations. Future work will explore adaptive mechanisms for the hyperparameters and extend the local invariance principle in AI security.
\label{sec:conclusion}

\bibliography{ref}
\bibliographystyle{IEEEtran}
\clearpage
\clearpage
\appendices

\setcounter{page}{1}
\setcounter{equation}{0}
\renewcommand{\theequation}{A-\arabic{equation}}
\setcounter{figure}{0}
\renewcommand{\thefigure}{A-\arabic{figure}}
\setcounter{table}{0}
\renewcommand{\thetable}{A-\arabic{table}}

\begin{center}
    {\textsc{Supplementary Material}\par}
\end{center}

\vspace{-.5cm}
\subsection{Parameter Settings}
\label{parameters}

\begin{center}
\scriptsize
\renewcommand{\arraystretch}{.98} 

\tablefirsthead{
    \hline 
    \textbf{Category} & \textbf{Method} & \textbf{Parameters} \\ 
    \hline \hline
}

\tablehead{
    \hline 
    \textbf{Category} & \textbf{Method} \textit{(Cont.)} & \textbf{Parameters} \\ 
    \hline \hline
}

\tabletail{
    \hline 
    \multicolumn{3}{r}{\textit{Continued on next column...}} \\
}

\tablelasttail{\hline}
\vspace{-.3cm}
\captionof{table}{Hyperparameters for various transfer-based attack baselines.}
\label{tab:parameters}

\begin{supertabular}{>{\centering\arraybackslash}m{1.5cm} c >{\centering\arraybackslash}p{4.3cm}}

\multirow{16}{*}{\parbox{1.5cm}{\centering \textbf{Gradient-based Attacks}}} 
    & \multirow{4}{*}{MI-FGSM~\cite{dong2018mifgsm}} & perturbation budget $\epsilon = 16/255$,\\
    & & number of iterations $T=10$, \\
    & & step size $\alpha = \epsilon/T=1.6/255$, \\
    & & decay factor $\mu=1.0$\\
    \cmidrule{2-3}
    & \multirow{2}{*}{VMI-FGSM~\cite{wang2021vmi}} & number of sampled examples $N_s=20$, \\
    & & upper bound of neighborhood $\zeta=1.5$\\
    \cmidrule{2-3}
    & \multirow{4}{*}{PGN~\cite{ge2023boosting}} & number of sampled examples $N_s=20$, \\
    & & balanced coefficient $c_b=0.5$,\\
    & & upper bound of neighborhood $\zeta=3.0\times\epsilon$\\
    \cmidrule{2-3}
    & \multirow{6}{*}{MUMODIG~\cite{ren2025mumodig}} & position factor $\lambda_{p} = 0.65$, \\ 
    & & region number $N_R=2$, \\ 
    & & interpolation point number $N_T=1$, \\
    & & number of sampled baselines $N_B=1$,\\
    & & number of sampled transformations $N_T=6$\\
    \midrule \midrule

\multirow{21}{*}{\parbox{1.5cm}{\centering \textbf{Input \\ Transformation-based \\ Attacks}}} 
    & \multirow{2}{*}{DIM~\cite{xie2019dim}} & resize rate $r=1.1$, \\
    & & diversity probability $p_{di}=0.5$\\ 
    \cmidrule{2-3}
    & \multirow{2}{*}{TIM~\cite{dong2019TIM}} & kernel type: Gaussian,\\
    & & kernel size $k_s=15$ \\
    \cmidrule{2-3}
    & \multirow{3}{*}{\textit{Admix}~\cite{wang2021admix}} & number of scaled copies $m_1 = 5$, \\
    & & number of admixed images $m_2 = 3$, \\
    & & admix strength $\eta=0.2$ \\
    \cmidrule{2-3}
    & \multirow{2}{*}{SIA~\cite{Wang2023SIA}} & number of blocks $s=3$,\\
    & & number of transformed images $N_t=20$ \\
    \cmidrule{2-3}
    & \multirow{3}{*}{BSR~\cite{wang2024bsr}} & number of blocks $s=3$,\\
    & & number of shuffled images $N_u=20$, \\
    & & range of rotation angles $\tau = 24^{\circ}$ \\
    \cmidrule{2-3}
    & \multirow{5}{*}{SID~\cite{zhou2025sid}} & number of sampled $N_s=20$,\\
    & & downsampling factors $\beta=0.1$, \\
    & & number of blocks $s=2$, \\
    & & probability of block fusion $p_{fusion}=0.5$, \\
    & & linear fusion weight $w_{linear}=0.5$ \\
    \midrule \midrule

\multirow{16}{*}{\parbox{1.5cm}{\centering \textbf{Model-related Attacks}}} 
    & SGM~\cite{wu2020sgm} & residual gradient decay $\gamma=0.5$\\
    \cmidrule{2-3}
    & \multirow{2}{*}{Linbp~\cite{guo2020linbp}} & number of iterations $T=300$,\\
    & & the first layer to be modified is the first residual unit in the third meta block.\\
    \cmidrule{2-3}
    & \multirow{2}{*}{BPA~\cite{wang2023bpa}} & temperature coefficient $c_t = 10$,\\
    & & the first layer to be modified is the first residual unit in the third meta block.\\
    \cmidrule{2-3}
    & \multirow{3}{*}{VDC~\cite{zhang2024VirtualDense}} & patch size $P_s=16$,\\
    & & scale factor $s_f = 0.5$, \\
    & & residual gradient decay $\gamma = 0.5$ \\
    \cmidrule{2-3}
    & \multirow{5}{*}{FPR~\cite{ren2025fpr}} & diversity factor $d_f = 25$, \\
    & & scale factor $s_f = 0.8$, \\
    & & attenuation factor $a_f=0.3$, \\
    & & index set of diversified blocks $I=[0,1,4,9,11]$ \\
    \midrule \midrule

\multirow{17}{*}{\parbox{1.5cm}{\centering \textbf{Advanced Objective Functions}}} 
    & ILA~\cite{huang2019ila} & coefficient $c=1.0$\\
    \cmidrule{2-3}
    & \multirow{5}{*}{FIA~\cite{wang2021fia}} & drop probability $p_{dr}=0.3$,\\
    & & number of aggregated gradients $N_a=30$,\\
    & & the target layer to attack is the last layer of the second block.\\
    \cmidrule{2-3}
    & \multirow{6}{*}{ILPD~\cite{Li2023ILPD}} & number of iterations $T=100$, \\
    & & noise size $\sigma=0.05$, \\
    & & coefficient $c=0.1$, \\
    & & step size $\alpha=1/255$, \\
    & & the target layer to attack is the third building block of the second ResNet meta layer.\\
    \cmidrule{2-3}
    & \multirow{4}{*}{BFA~\cite{WANG2024bfa}} & perturbation mask size $s_{mask}=28 $, \\
    & & number of fitting iteration steps $T=30$, \\ 
    & & the target layer to attack is the last layer of the second block.\\
\midrule
\end{supertabular}
\end{center}

In this section, we provide the detailed parameter settings for the baseline attacks employed in our work. These settings are consistent with the corresponding papers to ensure fair and comprehensive evaluations. We delineate the hyperparameters
for each category of baseline methods in Tab.~\ref{tab:parameters}. 

All the defense mechanisms are pre-trained on the ImageNet dataset and evaluated on a single model. AT~\cite{Wong2020ATdefense}and HGD~\cite{liao2018defense} adopt the official models provided in the corresponding papers. RS~\cite{cohen2019RS} utilizes the defense model ResNet-50 with a noise level of 0.5. For NRP~\cite{naseer2020NRP} and DiffPure~\cite{nie2022diffpure}, we choose ResNet-101 as the target classifier. For MIMIR~\cite{xu2026mimir}, we choose the most robust ViT-B trained with 2-step APGD~\cite{2021apgd} adversarial training for 300 epochs.

\subsection{Extended Evaluations Using Other Surrogate Models}
\label{appendix:othermodels}
To further validate the effectiveness of LI-Boost, we conduct experiments
w/wo LI-Boost on various surrogate models. The results of gradient-based attacks are summarized in Tab.\ref{tab:gradient_othermodels}, while the results of input transformation-based attacks are presented in Tab.~\ref{tab:input_othermodels}. It is important to note that model-related attacks and advanced objective functions depend on the specific architectures of deep learning models. Since their implementation was not provided for a wide range of surrogate models, we also do not report the results for these cases. 

These empirical results consistently demonstrate that incorporating LI-Boost yields substantial performance gains over all baseline attacks, achieving significantly higher attack success rates across diverse architectures. For instance, when utilizing Inc-v3 as the surrogate model, the classical PGN attack achieves an attack success rate of 57.3\% against the Swin model. However, with the integration of LI-Boost, the performance is significantly elevated to 70.4\%, representing a substantial absolute gain of 13.1\%. Similar noteworthy improvements are also observed in input transformation-based attacks. For example, when employing DN-121 as the surrogate model to attack ViT, the baseline \textit{Admix} yields an attack success rate of 45.2\%. Upon integration with LI-Boost, the performance is remarkably bolstered to 67.0\%, achieving a significant performance leap of 21.8\%. These empirical findings further validate that LI-Boost consistently serves as a potent enhancer for a wide spectrum of various categories of attacks, effectively narrowing the gap between the surrogate model and target model distributions.

\begin{table*}[t]
  \centering
  \captionsetup{labelsep=newline, justification=centering}
  \caption{Transfer attack success rates (\%) of gradient-based attacks using various surrogate models. The best results are \textbf{bold}, while methods incorporating LI-Boost are highlighted in \colorbox{gray!10}{gray}.}
  \label{tab:gradient_othermodels}
  
  \resizebox{.95\linewidth}{!}{
    \setlength{\tabcolsep}{2pt} 
    \setlength{\extrarowheight}{1.5pt}
    \begin{tabular}{c|cccccccc|cccccccc}
      \hline
      \multirow{2}{*}{\textbf{Attacks}} & \multicolumn{8}{c|}{\textbf{Inc-v3} $\Longrightarrow$} & \multicolumn{8}{c}{\textbf{MN-v3} $\Longrightarrow$} \\
      & \textbf{RN-50} & \textbf{MN-v3} & \textbf{DN-121} & \textbf{FSNet} & \textbf{ViT} & \textbf{PiT} & \textbf{Visformer} & \textbf{Swin} & \textbf{RN-50} & \textbf{Inc-v3} & \textbf{DN-121} & \textbf{FSNet} & \textbf{ViT} & \textbf{PiT} & \textbf{Visformer} & \textbf{Swin} \\
      \hline
      MI-FGSM & 34.1 & 46.7 & 50.6 & 25.7 & 14.2 & 20.5 & 26.6 & 31.8 & 41.7 & 50.6& 60.0 & 31.4 & 17.9 & 26.6 & 36.9 & 42.7 \\
      \rowcolor{gray!10}\textbf{LI-Boost-MI} & \textbf{43.8} & \textbf{56.9} & \textbf{62.6} & \textbf{37.3} & \textbf{20.8} & \textbf{26.4} & \textbf{35.0} & \textbf{40.9} & \textbf{65.6}& \textbf{67.6}& \textbf{80.7} & \textbf{53.3} & \textbf{33.0} & \textbf{46.1} & \textbf{61.1} & \textbf{65.7} \\
      \hline
      VMI-FGSM & 50.0 & 60.3 & 66.3 & 41.6 & 25.5 & 32.7 & 39.7 & 44.8 & 67.5 & 73.0 & 80.7 & 57.6 & 37.2 & 51.6 & 64.2 & 69.7 \\
      \rowcolor{gray!10}\textbf{LI-Boost-VMI} & \textbf{52.5} & \textbf{62.0} & \textbf{69.1} & \textbf{44.2} & \textbf{27.0} & \textbf{35.1} & \textbf{42.6} & \textbf{48.0} & \textbf{75.5} & \textbf{79.0} & \textbf{87.2} & \textbf{66.0} & \textbf{44.5} & \textbf{58.6} & \textbf{71.7} & \textbf{77.2} \\
      \hline
      PGN & 63.1 & 75.7 & 81.2 & 55.7 & 55.7 & 43.7 & 51.7 & 57.3 & 80.3 & 86.5 & 92.1 & 70.9 & 49.6 & 64.7 & 76.4 & 82.5 \\
      \rowcolor{gray!10}\textbf{LI-Boost-PGN} &\textbf{77.3} & \textbf{83.8} & \textbf{89.7} & \textbf{70.8} & \textbf{47.4} & \textbf{57.4} & \textbf{66.6} & \textbf{70.4} & \textbf{84.9} & \textbf{89.8} & \textbf{94.5} & \textbf{78.9} & \textbf{59.1} & \textbf{71.9} & \textbf{82.0} & \textbf{86.2} \\
      \hline
      MUMODIG & 65.1 & 76.6 & 81.7 & 57.0 & 34.0 & 42.7 & 53.9 & 58.2 & 83.5 & 88.7 & 93.3 & 75.0 & 51.8 & 68.7 & 81.0 & 82.7 \\
      \rowcolor{gray!10}\textbf{LI-Boost-MUMODIG} & \textbf{89.4} & \textbf{92.1} & \textbf{96.2} & \textbf{85.0} & \textbf{62.4} & \textbf{73.5} & \textbf{83.4} & \textbf{81.8} & \textbf{90.1} & \textbf{92.0} & \textbf{96.0} & \textbf{84.6} & \textbf{68.9} & \textbf{81.0} & \textbf{88.8} & \textbf{90.2} \\
      \hline
      \multirow{2}{*}{\textbf{Attacks}} & \multicolumn{8}{c|}{\textbf{DN-121} $\Longrightarrow$} & \multicolumn{8}{c}{\textbf{FSNet} $\Longrightarrow$} \\
      & \textbf{RN-50} & \textbf{Inc-v3} & \textbf{MN-v3} & \textbf{FSNet} & \textbf{ViT} & \textbf{PiT} & \textbf{Visformer} & \textbf{Swin} & \textbf{RN-50} & \textbf{Inc-v3} & \textbf{MN-v3} & \textbf{DN-121} & \textbf{ViT} & \textbf{PiT} & \textbf{Visformer} & \textbf{Swin} \\
      \hline
      MI-FGSM & 67.1 & 61.5 & 71.5 & 49.6 & 24.3 & 33.6 & 47.9 & 50.2 & 44.8& 42.5& 51.6 & 53.4 & 20.0 & 31.8 & 39.9 & 47.5 \\
      \rowcolor{gray!10}\textbf{LI-Boost-MI} & \textbf{82.2} & \textbf{74.9} & \textbf{84.8} & \textbf{70.1} & \textbf{35.8} & \textbf{47.1} & \textbf{66.1} & \textbf{66.3} & \textbf{65.4}& \textbf{71.5}& \textbf{71.2} & \textbf{75.4} & \textbf{37.8} & \textbf{54.8} & \textbf{67.2} & \textbf{69.6} \\
      \hline
      VMI-FGSM & 84.7 & 79.7 & 86.0 & 72.3 & 42.7 & 54.6 & 69.6 & 70.5 & 69.7 & 62.8 & 69.3 & 73.0 & 44.0 & 57.6 & 66.1 & 70.5\\
      \rowcolor{gray!10}\textbf{LI-Boost-VMI} & \textbf{90.9} & \textbf{86.5} & \textbf{91.1} & \textbf{80.7} & \textbf{50.1} & \textbf{63.6} & \textbf{78.1} & \textbf{78.6} & \textbf{88.4} & \textbf{80.5} & \textbf{85.8} & \textbf{89.6} & \textbf{65.6} & \textbf{79.1} & \textbf{86.1} & \textbf{88.2} \\
      \hline
      PGN & 94.1 & 93.3 & 95.2 & 86.9 & 60.0 & 72.4 & 84.8 & 85.1 & 93.7 & 89.9 & 92.0 & 94.5 & 78.6 & 88.4 & 92.1 & 93.1 \\
      \rowcolor{gray!10}\textbf{LI-Boost-PGN} & \textbf{95.0} & \textbf{94.1} & \textbf{95.2} & \textbf{88.6} & \textbf{64.4} & \textbf{75.5} & \textbf{86.4} & \textbf{86.9} & \textbf{94.6} & \textbf{92.2} & \textbf{93.4} & \textbf{95.6} &\textbf{83.1} & \textbf{90.7} & \textbf{93.4} & \textbf{93.7} \\
      \hline
      MUMODIG & 95.2 & 93.6 & 95.0 & 86.7 & 55.7 & 69.0 & 84.5 & 82.2 & 88.3 & 81.5 & 86.0 & 89.9 & 63.2 & 79.2 & 85.7 & 85.8 \\
      \rowcolor{gray!10}\textbf{LI-Boost-MUMODIG} & \textbf{97.2} & \textbf{96.5} & \textbf{97.1} & \textbf{92.8} & \textbf{70.7} & \textbf{80.7} & \textbf{92.0} & \textbf{89.4} & \textbf{92.4} & \textbf{85.9} & \textbf{89.8} & \textbf{93.3} & \textbf{74.4} & \textbf{85.7} & \textbf{90.7} & \textbf{90.8} \\
      \hline
      \multirow{2}{*}{\textbf{Attacks}} & \multicolumn{8}{c|}{\textbf{ViT} $\Longrightarrow$} & \multicolumn{8}{c}{\textbf{PiT} $\Longrightarrow$} \\
      & \textbf{RN-50} & \textbf{Inc-v3} & \textbf{MN-v3} & \textbf{DN-121} & \textbf{FSNet}  & \textbf{PiT} & \textbf{Visformer} & \textbf{Swin} & \textbf{RN-50} & \textbf{Inc-v3} & \textbf{MN-v3} & \textbf{DN-121} & \textbf{FSNet} & \textbf{ViT} & \textbf{Visformer} & \textbf{Swin} \\
      \hline
      MI-FGSM & 43.7 & 51.3 & 57.8 & 57.2 & 43.4 & 45.6 & 49.3 & 61.5 & 44.3 & 48.5 & 57.0 & 54.4 & 41.3 & 30.6 & 50.0 & 53.5 \\
      \rowcolor{gray!10}\textbf{LI-Boost-MI} & \textbf{53.2} & \textbf{57.8} & \textbf{63.2} & \textbf{64.7} & \textbf{53.4} & \textbf{58.1} & \textbf{60.1} & \textbf{68.4} & \textbf{56.8} & \textbf{56.1} & \textbf{67.0} & \textbf{64.5} & \textbf{54.4} & \textbf{45.0} & \textbf{64.6} & \textbf{67.6} \\
      \hline
      VMI-FGSM & 55.7 & 61.8 & 66.9 & 66.1 & 58.3 & 62.1 & 64.0 & 73.3 & 61.6 & 62.1 & 69.9 & 67.8 & 61.7 & 52.3 & 69.5 & 72.0 \\
      \rowcolor{gray!10}\textbf{LI-Boost-VMI} & \textbf{61.6} & \textbf{67.8} & \textbf{72.1} & \textbf{71.7} & \textbf{65.6} & \textbf{68.7} & \textbf{69.5} & \textbf{77.5} & \textbf{69.7} & \textbf{70.4} & \textbf{76.6} & \textbf{75.7} & \textbf{70.9} & \textbf{60.4} & \textbf{77.1} & \textbf{78.5}  \\
      \hline
      PGN & 76.3 & 78.9 & 83.7 & 83.1 & 78.5 & 83.4 & 83.2 & 87.6 & 78.9 & 79.4 & 83.5 & 82.7 & 80.1 & 76.3 & 84.5 & \textbf{85.3} \\
      \rowcolor{gray!10}\textbf{LI-Boost-PGN} & \textbf{78.2} & \textbf{80.8} & \textbf{84.6} & \textbf{84.9} & \textbf{80.7} & \textbf{84.7} & \textbf{84.7} & \textbf{88.5} & \textbf{79.5} & \textbf{80.5} & \textbf{83.7} & \textbf{82.9} & \textbf{80.5} & \textbf{77.5} & \textbf{84.6} & 85.2 \\
      \hline
      MUMODIG & 70.9 & 74.9 & 78.2 & 77.1 & 73.1 & 77.8 & 78.1 & 80.9 & 76.2 & 75.7 & 80.9 & 79.7 & 77.8 & 69.8 & 82.8 & 83.7 \\
      \rowcolor{gray!10}\textbf{LI-Boost-MUMODIG} & \textbf{77.7} & \textbf{78.1} & \textbf{82.6} & \textbf{83.0} & \textbf{79.5} & \textbf{83.7} & \textbf{84.0} & \textbf{86.3} & \textbf{81.7} & \textbf{79.9} & \textbf{85.3} & \textbf{84.7} & \textbf{83.0} & \textbf{78.6} & \textbf{88.2} & \textbf{88.2} \\
      \hline
      \multirow{2}{*}{\textbf{Attacks}} & \multicolumn{8}{c|}{\textbf{Visformer} $\Longrightarrow$} & \multicolumn{8}{c}{\textbf{Swin} $\Longrightarrow$} \\
      & \textbf{RN-50} & \textbf{Inc-v3} & \textbf{MN-v3}& \textbf{DN-121} & \textbf{FSNet} & \textbf{ViT} & \textbf{PiT}  & \textbf{Swin} & \textbf{RN-50} & \textbf{Inc-v3} & \textbf{MN-v3} & \textbf{DN-121} & \textbf{FSNet} & \textbf{ViT} & \textbf{PiT} & \textbf{Visformer}\\
      \hline
      MI-FGSM & 52.4 & 52.5 & 65.6 & 63.5 & 53.6 & 32.8 & 52.0 & 64.0 & 32.8 & 36.9 & 50.0 & 44.2 & 33.0 & 22.2 & 30.5 & 38.9 \\
      \rowcolor{gray!10}\textbf{LI-Boost-MI} & \textbf{68.4} & \textbf{64.0} & \textbf{77.7} & \textbf{77.4} & \textbf{71.3} & \textbf{51.1} & \textbf{69.8} & \textbf{78.4} & \textbf{59.0} & \textbf{55.0} & \textbf{73.6} & \textbf{68.7} & \textbf{59.8} & \textbf{44.5} & \textbf{59.2} & \textbf{69.3} \\
      \hline
      VMI-FGSM & 73.7 & 71.0 & 80.6 & 80.3 & 76.7 & 59.8 & 76.7 & 82.8 & 57.4 & 58.5 & 71.6 & 66.6 & 69.1 & 51.0 & 61.9 & 68.9  \\
      \rowcolor{gray!10}\textbf{LI-Boost-VMI} & \textbf{77.8} & \textbf{75.3} & \textbf{82.9} & \textbf{82.7} & \textbf{80.1} & \textbf{66.2} & \textbf{80.4} & \textbf{85.5} & \textbf{76.3} & \textbf{76.7} & \textbf{87.8} & \textbf{84.4} & \textbf{81.4} & \textbf{71.3} & \textbf{81.4} & \textbf{87.3} \\
      \hline
      PGN & 88.6 & 87.5 & \textbf{91.5} & 92.4 & \textbf{90.0} & 83.3 & 90.9 & \textbf{92.7} & 85.5 & 86.9 & 93.5 & 91.3 & 89.0 & 85.7 & 90.0 & 92.7 \\
      \rowcolor{gray!10}\textbf{LI-Boost-PGN} & \textbf{89.1} & \textbf{88.3} & 91.3 & \textbf{92.7} & 89.5 & \textbf{84.3} & \textbf{90.9} & 92.5 & \textbf{87.4} & \textbf{88.4} & \textbf{93.5} & \textbf{92.6} & \textbf{90.1} & \textbf{87.0} & \textbf{91.3} & \textbf{93.3} \\
      \hline
      MUMODIG & 88.8 & 85.8 & 91.8 & 91.8 & 89.9 & 76.2 & 90.5 & 92.4 & 80.8 & 80.3 & 88.9 & 86.7 & 84.0 & 69.5 & 84.5 & 87.9 \\
      \rowcolor{gray!10}\textbf{LI-Boost-MUMODIG} & \textbf{90.9} & \textbf{88.7} & \textbf{92.3} & \textbf{93.3} & \textbf{92.0} & \textbf{82.6} & \textbf{92.8} & \textbf{93.9} & \textbf{87.4} & \textbf{86.0} & \textbf{92.9} & \textbf{91.7} & \textbf{89.9} & \textbf{81.9} & \textbf{90.6} & \textbf{93.2} \\
      \hline
    \end{tabular}
  }
\end{table*}
\begin{table*}[t]
  \centering
  \captionsetup{labelsep=newline, justification=centering}
  \caption{Transfer attack success rates (\%) of input transformation-based attacks using various surrogate models. The best results are \textbf{bold}, while methods incorporating LI-Boost are highlighted in \colorbox{gray!10}{gray}.}
  \label{tab:input_othermodels}
  
  \resizebox{.95\linewidth}{!}{
    \setlength{\tabcolsep}{2pt} 
    \setlength{\extrarowheight}{1.5pt}
    \begin{tabular}{c|cccccccc|cccccccc}
      \hline
      \multirow{2}{*}{\textbf{Attacks}} & \multicolumn{8}{c|}{\textbf{Inc-v3} $\Longrightarrow$} & \multicolumn{8}{c}{\textbf{MN-v3} $\Longrightarrow$} \\
      & \textbf{RN50} & \textbf{MN-v3} & \textbf{DN-121} & \textbf{FSNet} & \textbf{ViT} & \textbf{PiT} & \textbf{Visformer} & \textbf{Swin} & \textbf{RN50} & \textbf{Inc-v3} & \textbf{DN-121} & \textbf{FSNet} & \textbf{ViT} & \textbf{PiT} & \textbf{Visformer} & \textbf{Swin} \\
      \hline
      DIM & 46.0 & 58.6 & 65.0 & 39.2 & 22.0 & 29.3 & 36.3 & 41.6 & 64.7 & 74.1 & 81.9 & 54.4 & 36.4 & 50.0 & 62.0 & 66.8 \\
      \rowcolor{gray!10}\textbf{LI-Boost-DIM} & \textbf{55.6} & \textbf{67.1} & \textbf{74.1} & \textbf{50.0} & \textbf{29.5} & \textbf{35.6} & \textbf{46.3} & \textbf{51.8} & \textbf{80.2} & \textbf{83.0} & \textbf{91.0} & \textbf{71.4} & \textbf{51.7} & \textbf{63.3} & \textbf{77.2} & \textbf{80.4} \\
      \hline
      \textit{Admix} & 56.3 & 68.3 & 75.4 & 45.5 & 25.3 & 33.8 & 43.6 & 48.7 & 70.9 & 75.8 & 85.3 & 58.1 & 36.5 & 53.0 & 66.8 & 71.9 \\
      \rowcolor{gray!10}\textbf{LI-Boost-\textit{Admix}} & \textbf{79.8} & \textbf{84.9} & \textbf{91.7} & \textbf{62.7} & \textbf{49.0} & \textbf{52.6} & \textbf{67.0} & \textbf{70.0} & \textbf{85.9} & \textbf{88.6} & \textbf{94.3} & \textbf{76.5} & \textbf{59.1} & \textbf{70.5} & \textbf{83.4} & \textbf{85.2} \\
      \hline
      SIA & 77.5 & 88.0 & 90.6 & 66.9 & 39.3 & 52.9 & 66.0 & 68.7 & 82.2 & 82.6 & 92.6 & 71.3 & 46.0 & 65.5 & 79.4 & 83.1 \\
      \rowcolor{gray!10}\textbf{LI-Boost-SIA} & \textbf{91.3} & \textbf{96.4} & \textbf{97.9} & \textbf{85.8} & \textbf{61.7} & \textbf{72.3} & \textbf{84.5} & \textbf{85.9} & \textbf{92.7} & \textbf{90.4} & \textbf{97.4} & \textbf{84.8} & \textbf{63.6} & \textbf{79.4} & \textbf{90.5} & \textbf{91.9} \\
      \hline
      BSR & 78.7 & 88.8 & 92.3 & 69.5 & 43.2 & 54.9 & 68.5 & 70.6 & 88.7 & 89.2 & 96.0 & 79.5 & 57.2 & 76.5 & 85.7 & 87.3 \\
      \rowcolor{gray!10}\textbf{LI-Boost-BSR} & \textbf{90.7} & \textbf{96.0} & \textbf{98.3} & \textbf{86.2} & \textbf{61.8} & \textbf{70.1} & \textbf{84.7} & \textbf{85.5} & \textbf{93.8} & \textbf{92.8} & \textbf{98.3} & \textbf{87.1} & \textbf{69.2} & \textbf{81.7} & \textbf{91.6} & \textbf{92.0} \\
      \hline
      SID & 85.2 & 90.4 & 96.2 & 79.6 & 52.5 & 63.7 & 76.8 & 78.0 & 93.8 & 94.8 & 98.6 & 88.7 & 71.4 & 85.3 & 93.0 & 94.0 \\
      \rowcolor{gray!10}\textbf{LI-Boost-SID} & \textbf{92.2} & \textbf{95.5} & \textbf{98.5} & \textbf{89.4} & \textbf{63.0} & \textbf{70.5} & \textbf{86.7} & \textbf{85.5} & \textbf{96.6} & \textbf{95.7} & \textbf{99.3} & \textbf{92.9} & \textbf{79.1} & \textbf{88.0} & \textbf{95.5} & \textbf{96.4} \\
      \hline
      \multirow{2}{*}{\textbf{Attacks}} & \multicolumn{8}{c|}{\textbf{DN-121} $\Longrightarrow$} & \multicolumn{8}{c}{\textbf{FSNet} $\Longrightarrow$} \\
      & \textbf{RN50} & \textbf{Inc-v3} & \textbf{MN-v3} & \textbf{FSNet} & \textbf{ViT} & \textbf{PiT} & \textbf{Visformer} & \textbf{Swin} & \textbf{RN50} & \textbf{Inc-v3} & \textbf{MN-v3} & \textbf{DN-121} & \textbf{ViT} & \textbf{PiT} & \textbf{Visformer} & \textbf{Swin} \\
      \hline
      DIM & 82.5 & 80.7 & 84.6 & 69.0 & 38.6 & 49.7 & 65.5 & 65.2 & 44.1 & 38.5 & 52.0 & 53.4 & 19.5 & 31.4 & 39.8 & 47.1 \\
      \rowcolor{gray!10}\textbf{LI-Boost-DIM} & \textbf{90.9} & \textbf{87.6} & \textbf{92.0} & \textbf{81.4} & \textbf{50.8} & \textbf{60.3} & \textbf{79.0} & \textbf{77.0} & \textbf{71.3} & \textbf{56.9} & \textbf{71.3} & \textbf{75.3} & \textbf{37.8} & \textbf{54.8} & \textbf{67.6} & \textbf{70.1} \\
      \hline
      \textit{Admix} & 91.3 & 87.5 & 91.3 & 77.8 & 45.2 & 58.7 & 75.9 & 74.5 & 82.0 & 72.1 & 79.6 & 84.5 & 52.7 & 69.0 & 79.5 & 81.9 \\
      \rowcolor{gray!10}\textbf{LI-Boost-\textit{Admix}} & \textbf{95.8} & \textbf{94.7} & \textbf{96.7} & \textbf{88.8} & \textbf{67.0} & \textbf{71.0} & \textbf{87.8} & \textbf{86.3} & \textbf{90.8} & \textbf{82.0} & \textbf{88.9} & \textbf{92.0} & \textbf{69.3} & \textbf{81.0} & \textbf{88.6} & \textbf{89.5} \\
      \hline
      SIA & 98.2 & 93.9 & 98.5 & 90.7 & 58.0 & 74.8 & 90.3 & 87.6 & 93.9 & 82.5 & 92.3 & 93.5 & 61.6 & 83.7 & 90.5 & 92.2 \\
      \rowcolor{gray!10}\textbf{LI-Boost-SIA} & \textbf{99.2} & \textbf{97.0} & \textbf{99.3} & \textbf{97.0} & \textbf{71.9} & \textbf{83.5} & \textbf{95.6} & \textbf{93.5} & \textbf{97.9} & \textbf{88.1} & \textbf{97.1} & \textbf{97.6} & \textbf{78.8} & \textbf{92.3} & \textbf{96.4} & \textbf{96.7} \\
      \hline
      BSR & 97.4 & 94.6 & 97.7 & 90.1 & 60.9 & 75.7 & 89.5 & 86.3 & 95.9 & 87.1 & 95.0 & 96.2 & 69.1 & 88.7 & 93.5 & 92.7 \\
      \rowcolor{gray!10}\textbf{LI-Boost-BSR} & \textbf{98.5} & \textbf{96.5} & \textbf{98.7} & \textbf{94.2} & \textbf{68.1} & \textbf{78.1} & \textbf{93.5} & \textbf{90.8} & \textbf{98.1} & \textbf{92.1} & \textbf{97.7} & \textbf{98.4} & \textbf{99.6} & \textbf{92.8} & \textbf{96.6} & \textbf{96.4} \\
      \hline
      SID & 97.9 & 97.6 & 98.4 & 93.4 & 70.2 & 80.2 & 93.0 & 91.0 & 97.0 & 95.1 & 96.7 & 97.5 & 85.8 & 94.2 & 96.5 & 96.8 \\
      \rowcolor{gray!10}\textbf{LI-Boost-SID} & \textbf{98.5} & \textbf{97.9} & \textbf{99.1} & \textbf{96.2} & \textbf{74.4} & \textbf{80.9} & \textbf{94.7} & \textbf{93.0} & \textbf{98.7} & \textbf{97.0} & \textbf{98.5} & \textbf{98.8} & \textbf{91.3} & \textbf{96.6} & \textbf{98.4} & \textbf{98.2} \\
      \hline
      \multirow{2}{*}{\textbf{Attacks}} & \multicolumn{8}{c|}{\textbf{ViT} $\Longrightarrow$} & \multicolumn{8}{c}{\textbf{PiT} $\Longrightarrow$} \\
      & \textbf{RN50} & \textbf{Inc-v3} & \textbf{MN-v3} & \textbf{DN-121} & \textbf{FSNet} & \textbf{PiT} & \textbf{Visformer} & \textbf{Swin} & \textbf{RN50} & \textbf{Inc-v3} & \textbf{MN-v3} & \textbf{DN-121} & \textbf{FSNet} & \textbf{ViT} & \textbf{Visformer} & \textbf{Swin} \\
      \hline
      DIM & 55.4 & 61.9 & 64.7 & 65.4 & 58.6 & 62.3 & 62.7 & 68.3 & 60.1 & 63.0 & 69.5 & 68.1 & 62.2 & 52.6 & 69.8 & 71.9 \\
      \rowcolor{gray!10}\textbf{LI-Boost-DIM} & \textbf{64.0} & \textbf{66.7} & \textbf{71.8} & \textbf{72.0} & \textbf{70.8} & \textbf{72.1} & \textbf{72.4} & \textbf{76.2} & \textbf{69.1} & \textbf{68.4} & \textbf{76.2} & \textbf{74.8} & \textbf{73.8} & \textbf{65.2} & \textbf{78.4} & \textbf{79.2} \\
      \hline
      \textit{Admix} & 61.2 & 66.9 & 72.2 & 71.9 & 63.2 & 67.5 & 69.7 & 80.7 & 60.4 & 57.7 & 69.1 & 66.8 & 61.3 & 45.6 & 67.9 & 71.2 \\
      \rowcolor{gray!10}\textbf{LI-Boost-\textit{Admix}} & \textbf{72.4} & \textbf{73.3} & \textbf{81.4} & \textbf{81.2} & \textbf{73.7} & \textbf{77.5} & \textbf{80.3} & \textbf{85.5} & \textbf{71.6} & \textbf{65.0} & \textbf{78.2} & \textbf{76.0} & \textbf{73.8} & \textbf{61.3} & \textbf{79.1} & \textbf{81.0} \\
      \hline
      SIA & 82.3 & 80.2 & 88.4 & 86.7 & 82.9 & 88.3 & 87.7 & 90.8 & 87.7 & 79.2 & 91.1 & 89.0 & 87.6 & 78.7 & 93.1 & 93.7 \\
      \rowcolor{gray!10}\textbf{LI-Boost-SIA} & \textbf{88.8} & \textbf{84.8} & \textbf{92.7} & \textbf{91.7} & \textbf{89.6} & \textbf{93.1} & \textbf{93.3} & \textbf{94.7} & \textbf{92.8} & \textbf{85.6} & \textbf{95.1} & \textbf{93.6} & \textbf{93.5} & \textbf{89.6} & \textbf{97.0} & \textbf{97.0} \\
      \hline
      BSR & 85.9 & 85.1 & 89.8 & 89.1 & 86.6 & 90.6 & 89.9 & 90.4 & 88.4 & 84.5 & 92.4 & 90.8 & 89.9 & 81.2 & 93.8 & 94.1 \\
      \rowcolor{gray!10}\textbf{LI-Boost-BSR} & \textbf{89.7} & \textbf{87.4} & \textbf{93.0} & \textbf{92.2} & \textbf{90.5} & \textbf{93.4} & \textbf{93.3} & \textbf{93.1} & \textbf{91.0} & \textbf{86.9} & \textbf{94.1} & \textbf{93.3} & \textbf{92.3} & \textbf{88.1} & \textbf{95.5} & \textbf{95.9} \\
      \hline
      SID & 84.8 & 84.4 & 88.5 & 89.0 & 86.5 & 89.9 & 90.0 & 90.4 & 92.0 & 89.9 & 94.0 & 93.8 & 93.4 & 90.0 & 96.1 & 96.2 \\
      \rowcolor{gray!10}\textbf{LI-Boost-SID} & \textbf{90.8} & \textbf{88.6} & \textbf{93.8} & \textbf{93.9} & \textbf{92.3} & \textbf{94.8} & \textbf{94.8} & \textbf{95.0} & \textbf{94.7} & \textbf{92.2} & \textbf{96.3} & \textbf{96.4} & \textbf{96.2} & \textbf{94.7} & \textbf{98.2} & \textbf{98.0} \\
      \hline
      \multirow{2}{*}{\textbf{Attacks}} & \multicolumn{8}{c|}{\textbf{Visformer} $\Longrightarrow$} & \multicolumn{8}{c}{\textbf{Swin} $\Longrightarrow$} \\
      & \textbf{RN50} & \textbf{Inc-v3} & \textbf{MN-v3} & \textbf{DN-121} & \textbf{FSNet} & \textbf{ViT} & \textbf{PiT} & \textbf{Swin} & \textbf{RN50} & \textbf{Inc-v3} & \textbf{MN-v3} & \textbf{DN-121} & \textbf{FSNet} & \textbf{ViT} & \textbf{PiT} & \textbf{Visformer} \\
      \hline
      DIM & 73.3 & 71.8 & 80.9 & 81.1 & 75.4 & 58.9 & 76.4 & 80.9 & 67.2 & 69.2 & 79.3 & 76.1 & 71.4 & 56.7 & 72.7 & 77.1 \\
      \rowcolor{gray!10}\textbf{LI-Boost-DIM} & \textbf{79.5} & \textbf{76.5} & \textbf{85.1} & \textbf{86.2} & \textbf{84.5} & \textbf{68.6} & \textbf{83.2} & \textbf{86.7} & \textbf{79.4} & \textbf{77.0} & \textbf{87.5} & \textbf{85.2} & \textbf{84.5} & \textbf{70.8} & \textbf{83.8} & \textbf{87.9} \\
      \hline
      \textit{Admix} & 77.4 & 73.5 & 84.3 & 83.2 & 78.9 & 58.6 & 80.3 & 86.3 & 47.5 & 42.3 & 63.7 & 57.3 & 47.7 & 33.6 & 45.6 & 56.5 \\
      \rowcolor{gray!10}\textbf{LI-Boost-\textit{Admix}} & \textbf{84.4} & \textbf{79.7} & \textbf{89.2} & \textbf{89.1} & \textbf{86.2} & \textbf{72.2} & \textbf{86.9} & \textbf{90.4} & \textbf{72.8} & \textbf{61.8} & \textbf{84.1} & \textbf{80.2} & \textbf{74.3} & \textbf{57.6} & \textbf{74.4} & \textbf{81.9} \\
      \hline
      SIA & 92.6 & 83.5 & 94.8 & 94.2 & 92.9 & 75.9 & 93.4 & 96.0 & 83.2 & 73.9 & 92.7 & 87.3 & 84.8 & 66.5 & 85.0 & 90.8 \\
      \rowcolor{gray!10}\textbf{LI-Boost-SIA} & \textbf{95.3} & \textbf{88.2} & \textbf{96.9} & \textbf{96.9} & \textbf{96.8} & \textbf{85.5} & \textbf{96.1} & \textbf{97.9} & \textbf{92.8} & \textbf{84.1} & \textbf{97.6} & \textbf{95.2} & \textbf{95.5} & \textbf{82.6} & \textbf{94.6} & \textbf{96.9} \\
      \hline
      BSR & 95.1 & 90.7 & 96.5 & 97.0 & 95.4 & 81.1 & 95.5 & 96.8 & 92.4 & 87.5 & 96.8 & 95.2 & 93.7 & 77.7 & 94.9 & 95.9 \\
      \rowcolor{gray!10}\textbf{LI-Boost-BSR} & \textbf{96.4} & \textbf{92.3} & \textbf{97.5} & \textbf{98.0} & \textbf{96.9} & \textbf{87.4} & \textbf{96.9} & \textbf{97.9} & \textbf{96.1} & \textbf{91.5} & \textbf{98.3} & \textbf{97.6} & \textbf{97.0} & \textbf{87.4} & \textbf{97.5} & \textbf{98.0} \\
      \hline
      SID & 95.9 & 94.1 & 96.9 & 97.3 & 96.3 & 90.8 & 97.0 & 97.5 & 95.5 & 93.9 & 98.5 & 97.9 & 96.4 & 89.2 & 97.5 & 98.4 \\
      \rowcolor{gray!10}\textbf{LI-Boost-SID} & \textbf{96.6} & \textbf{94.4} & \textbf{97.7} & \textbf{97.9} & \textbf{97.1} & \textbf{92.9} & \textbf{97.5} & \textbf{98.3} & \textbf{97.8} & \textbf{95.6} & \textbf{99.1} & \textbf{99.0} & \textbf{98.4} & \textbf{94.4} & \textbf{98.6} & \textbf{99.3} \\
      \hline
    \end{tabular}
  }
\end{table*}

\subsection{Sampling Distributions}
\label{appendix:sampling}
In this section, we detail the three sampling distributions for pixel translation employed in our study: uniform, normal, and logarithmic. For simplicity, we define the random variable for these distributions as the number of translated pixels. Given the actual number of translated pixels $x_p$, upperbound $k$, the probability mass function are as follows:

\medskip

\textbf{Uniform:}
\begin{multline}
    P_{\text{uniform}}(X=x_p;k) = \begin{cases}
        \frac{1}{k}, & \text{for} \,\, x_p \in \{ 1, 2,\dots, k\}\\
        0, & \text{otherwise}
    \end{cases}
    \label{uniform}
\end{multline}

\textbf{Normal:}
\begin{equation}
   P_{\text{normal}}(X=x_p; k) = \begin{cases}
\frac{\exp(-\frac{x_p^2}{2\sigma^2})}{Z_{\text{normal}}}, & \text{for } x_p \in \{1, 2, \dots, k\} \\
0, & \text{otherwise}
\end{cases}
    \label{normal}
\end{equation}
where $\mu=0$, $\sigma=2$ and $Z_{\text{normal}}$ is the normalization constant, given by:
\begin{equation}
    Z_{\text{normal}} = \sum_{i=1}^{k} \exp(-\frac{i^2}{2\sigma^2}) = \sum_{i=1}^{k} \exp(-\frac{i^2}{8})
\end{equation}

\textbf{Logarithmic:}
\begin{equation}
P_{\text{logarithmic}}(X=x_p; k) = \begin{cases}
\frac{\ln(\frac{k+1}{x_p})}{Z_{\text{logarithmic}}}, & \text{for } x_p \in \{1, 2, \dots, k\} \\
0, & \text{otherwise}
\end{cases}
\end{equation}
where $Z_{\text{logarithmic}}$ is the normalization constant, defined as:
\begin{equation}
Z_{\text{logarithmic}} = \sum_{i=1}^{k} \ln(\frac{k+1}{i})    
\end{equation}

\subsection{Additional Visualizations on Alibaba Commercial Vision API System and Claude-o}
\label{appdenx:api2vlm}
In this section, we present visual examples of adversarial attacks against the Alibaba Vision API for image classification and Claude-4o for image captioning task.

\begin{figure*}[tbp]
    \centering
    \footnotesize 
    \renewcommand{\arraystretch}{0.1} 
    \setlength{\tabcolsep}{3pt} 

    \begin{tabular}{ccc:ccc}
        \textbf{\scriptsize Clean} & \textbf{\scriptsize MI-FGSM} & \textbf{\scriptsize LI-Boost-MI} & \textbf{\scriptsize Clean} & \textbf{\scriptsize MI-FGSM} & \textbf{\scriptsize LI-Boost-MI}\\ [.3ex]
        
        \includegraphics[width=0.12\linewidth]{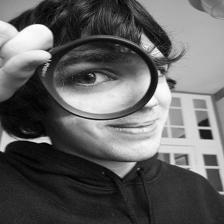} &
        \includegraphics[width=0.12\linewidth]{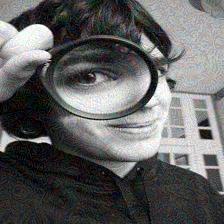} &
        \includegraphics[width=0.12\linewidth]{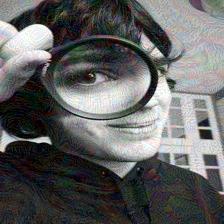} &

        \includegraphics[width=0.12\linewidth]{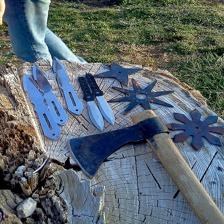} &
        \includegraphics[width=0.12\linewidth]{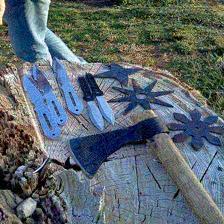} &
        \includegraphics[width=0.12\linewidth]{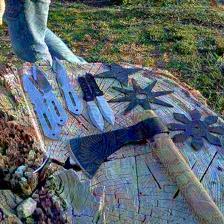} \\[1ex]
        \textbf{\textcolor{green!50!black}{\scriptsize Magnifying Glass}} & \textbf{\textcolor{gray}{\scriptsize Magnifying Glass}} & \textbf{\textcolor{red}{\scriptsize Mask}} & \textbf{\textcolor{green!50!black}{\scriptsize Axe }} & \textbf{\textcolor{gray}{\scriptsize Axe }} & \textbf{\textcolor{red}{\scriptsize Painting}} \\ [2ex]

        \includegraphics[width=0.12\linewidth]{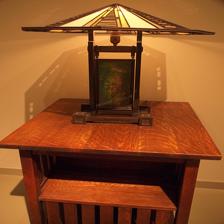} &
        \includegraphics[width=0.12\linewidth]{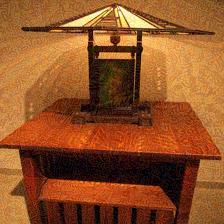} &
        \includegraphics[width=0.12\linewidth]{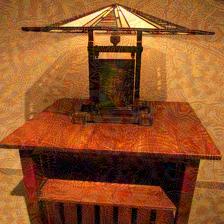} &
        \includegraphics[width=0.12\linewidth]{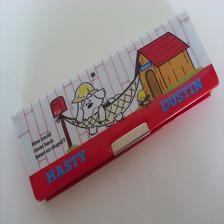} &
        \includegraphics[width=0.12\linewidth]{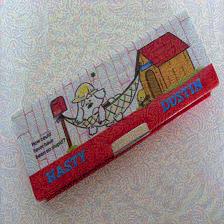} &
        \includegraphics[width=0.12\linewidth]{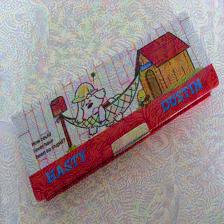} \\[1ex]
        \textbf{\textcolor{green!50!black}{\scriptsize Desk Lamp}} & \textbf{\textcolor{gray}{\scriptsize Desk Lamp}} & \textbf{\textcolor{red}{\scriptsize Birdcage}} & \textbf{\textcolor{green!50!black}{\scriptsize Pencil Case}} & \textbf{\textcolor{gray}{\scriptsize Pencil Case}} & \textbf{\textcolor{red}{\scriptsize Broom}} \\ [2ex]
        
\midrule
    
\textbf{\scriptsize Clean} & \textbf{\scriptsize BSR} & \textbf{\scriptsize LI-Boost-BSR} & \textbf{\scriptsize Clean} & \textbf{\scriptsize BSR} & \textbf{\scriptsize LI-Boost-BSR}\\ [.3ex]
        \includegraphics[width=0.12\linewidth]{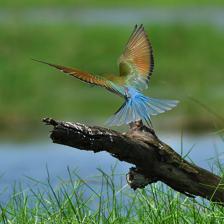} &
        \includegraphics[width=0.12\linewidth]{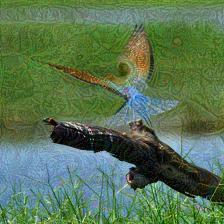} &
        \includegraphics[width=0.12\linewidth]{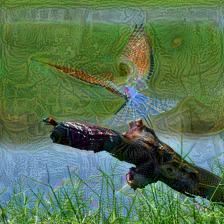} &
        \includegraphics[width=0.12\linewidth]{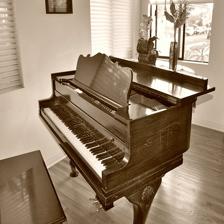} &
        \includegraphics[width=0.12\linewidth]{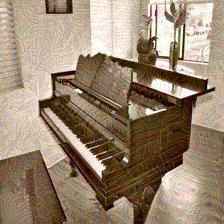} &
        \includegraphics[width=0.12\linewidth]{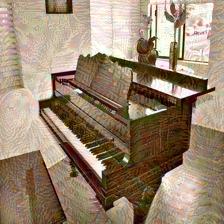} \\[1ex]
        
        \textbf{\textcolor{green!50!black}{\scriptsize Dragonfly}} & \textbf{\textcolor{gray}{\scriptsize Dragonfly}} & \textbf{\textcolor{red}{\scriptsize Turtle}} &        \textbf{\textcolor{green!50!black}{\scriptsize Grand piano}} & \textbf{\textcolor{gray}{\scriptsize Grand piano}} & \textbf{\textcolor{red}{\scriptsize Sofa}} \\ [1ex]

        \includegraphics[width=0.12\linewidth]{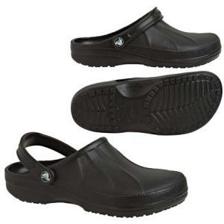} &
        \includegraphics[width=0.12\linewidth]{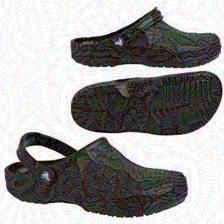} &
        \includegraphics[width=0.12\linewidth]{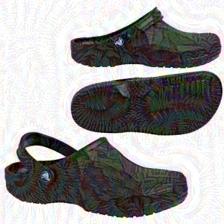} &

        \includegraphics[width=0.12\linewidth]{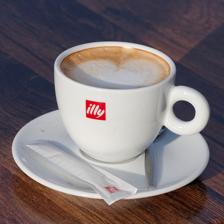} &
        \includegraphics[width=0.12\linewidth]{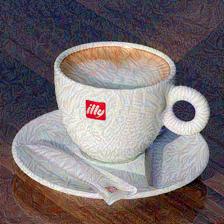} &
        \includegraphics[width=0.12\linewidth]{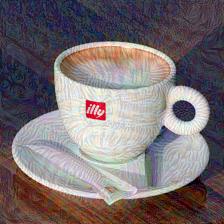} \\[1ex]
        \textbf{\textcolor{green!50!black}{\scriptsize Shoes}} & \textbf{\textcolor{gray}{\scriptsize Shoes}} & \textbf{\textcolor{red}{\scriptsize Snowboard}} & \textbf{\textcolor{green!50!black}{\scriptsize Cup}} & \textbf{\textcolor{gray}{\scriptsize Cup}} & \textbf{\textcolor{red}{\scriptsize Basket}} \\ [2ex]
\midrule
\textbf{\scriptsize Clean} & \textbf{\scriptsize ILA} & \textbf{\scriptsize LI-Boost-ILA} & \textbf{\scriptsize Clean} & \textbf{\scriptsize ILA} & \textbf{\scriptsize LI-Boost-ILA}\\ [.3ex]
        \includegraphics[width=0.12\linewidth]{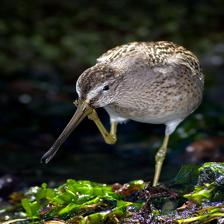} &
        \includegraphics[width=0.12\linewidth]{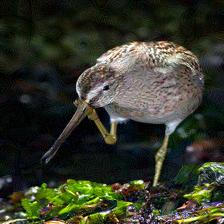} &
        \includegraphics[width=0.12\linewidth]{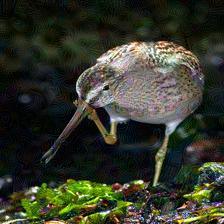} &
        \includegraphics[width=0.12\linewidth]{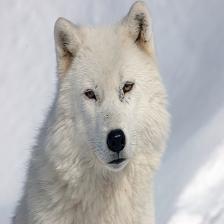} &
        \includegraphics[width=0.12\linewidth]{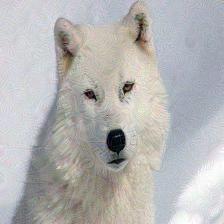} &
        \includegraphics[width=0.12\linewidth]{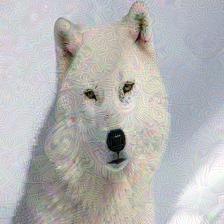} \\[1ex]
        \textbf{\textcolor{green!50!black}{\scriptsize Sandpiper}} & \textbf{\textcolor{gray}{\scriptsize Sandpiper}} & \textbf{\textcolor{red}{\scriptsize Chameleon}} & \textbf{\textcolor{green!50!black}{\scriptsize Wolf}} & \textbf{\textcolor{gray}{\scriptsize Wolf}} & \textbf{\textcolor{red}{\scriptsize Polar Bear}} \\ [2ex]

        \includegraphics[width=0.12\linewidth]{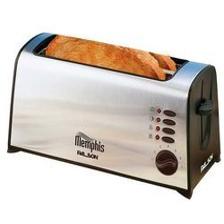} &
        \includegraphics[width=0.12\linewidth]{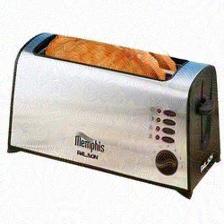} &
        \includegraphics[width=0.12\linewidth]{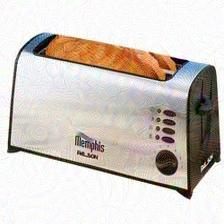} &
        \includegraphics[width=0.12\linewidth]{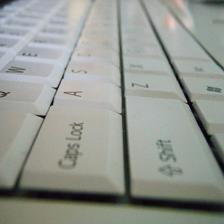} &
        \includegraphics[width=0.12\linewidth]{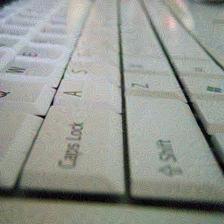} &
        \includegraphics[width=0.12\linewidth]{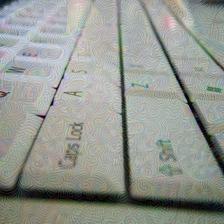} \\[1ex]
        
        \textbf{\textcolor{green!50!black}{\scriptsize Toaster}} & \textbf{\textcolor{gray}{\scriptsize Toaster}} & \textbf{\textcolor{red}{\scriptsize  Sharpener}} &        \textbf{\textcolor{green!50!black}{\scriptsize Keyboard}} & \textbf{\textcolor{gray}{\scriptsize Keyboard}} & \textbf{\textcolor{red}{\scriptsize Bill}} \\ [1ex]
\midrule
\textbf{\scriptsize Clean} & \textbf{\scriptsize BPA} & \textbf{\scriptsize LI-Boost-BPA} & \textbf{\scriptsize Clean} & \textbf{\scriptsize BPA} & \textbf{\scriptsize LI-Boost-BPA}\\ [.3ex]
        \includegraphics[width=0.12\linewidth]{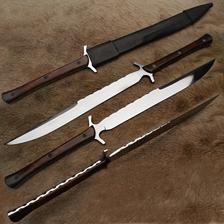} &
        \includegraphics[width=0.12\linewidth]{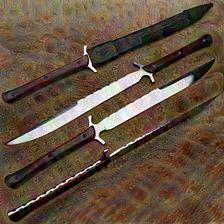} &
        \includegraphics[width=0.12\linewidth]{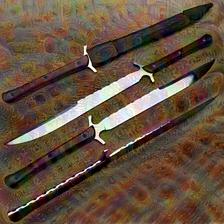} &

        \includegraphics[width=0.12\linewidth]{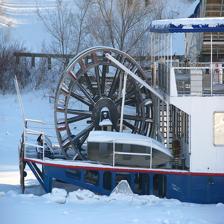} &
        \includegraphics[width=0.12\linewidth]{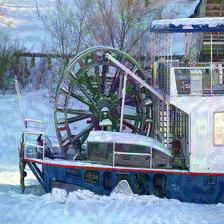} &
        \includegraphics[width=0.12\linewidth]{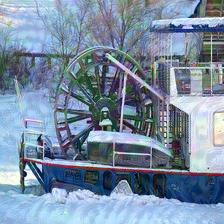} \\[1ex]
        \textbf{\textcolor{green!50!black}{\scriptsize Sheath}} & \textbf{\textcolor{gray}{\scriptsize Sheath}} & \textbf{\textcolor{red}{\scriptsize Rifle}} & \textbf{\textcolor{green!50!black}{\scriptsize Paddle wheel}} & \textbf{\textcolor{gray}{\scriptsize Paddle wheel}} & \textbf{\textcolor{red}{\scriptsize Harvester}} \\ [2ex]

        \includegraphics[width=0.12\linewidth]{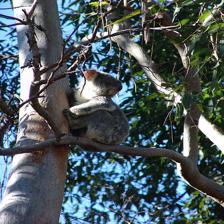} &
        \includegraphics[width=0.12\linewidth]{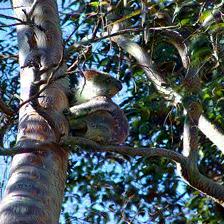} &
        \includegraphics[width=0.12\linewidth]{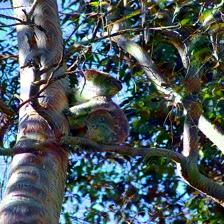} &
        \includegraphics[width=0.12\linewidth]{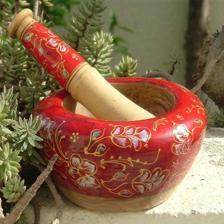} &
        \includegraphics[width=0.12\linewidth]{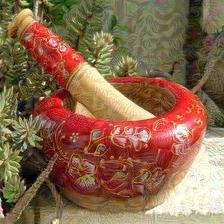} &
        \includegraphics[width=0.12\linewidth]{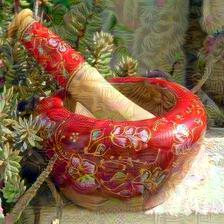} \\[1ex]
        \textbf{\textcolor{green!50!black}{\scriptsize Koala}} & \textbf{\textcolor{gray}{\scriptsize Koala}} & \textbf{\textcolor{red}{\scriptsize Monkey}} & \textbf{\textcolor{green!50!black}{\scriptsize Mortar}} & \textbf{\textcolor{gray}{\scriptsize Mortar}} & \textbf{\textcolor{red}{\scriptsize Shoes}} \\ [2ex]

    \end{tabular}

    \caption{Visualization of clean images and their adversarial counterparts across different attack methods against Alibaba vision API systems, and the surrogate model is RN-50. \textbf{\textcolor{green!50!black}{Green text}} represents the ground-truth labels. \textbf{\textcolor{gray}{Gray text}} denotes the results of failed adversarial examples. \textcolor{red}{\textbf{Red text}} indicates the misclassification of LI-Boost enhanced methods.}
    \label{fig:alibabavisionapi}
    \vspace{-.5cm}
\end{figure*}

Specifically, Fig.~\ref{fig:alibabavisionapi} demonstrates the attack effectiveness of LI-Boost against commercial image classification API system, \eg, Alibaba. For instance, baseline methods such as MI-FGSM and BSR, which preserve correct predictions, while our LI-Boost enhanced methods successfully induce misclassifications, \eg, a Dragonfly is recognized as a Turtle, and a Wolf is recognized as a Polar Bear. These examples confirm that LI-Boost successfully injects adversarial noise that disrupts the feature comprehension of the victim model, leading to misclassifications, which reveal significant security vulnerabilities within these deployed systems.

\begin{figure*}[htbp]
    \centering
    \footnotesize 
    \renewcommand{\arraystretch}{1.3}
    \setlength{\tabcolsep}{3pt}

    \begin{tabular}{ccc:ccc}
    
        \textbf{Clean} & \textbf{MI-FGSM} & \textbf{LI-Boost-MI} & \textbf{Clean} & \textbf{MI-FGSM} & \textbf{LI-Boost-MI} \\
        
        \includegraphics[width=0.155\textwidth]{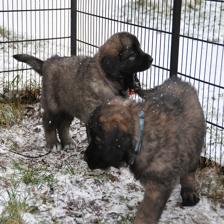} & 
        \includegraphics[width=0.155\textwidth]{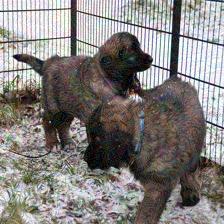} & 
        \includegraphics[width=0.155\textwidth]{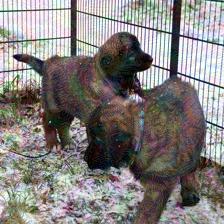} &
        \includegraphics[width=0.155\textwidth]{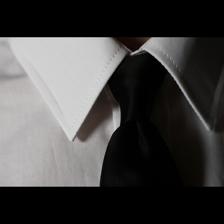} & 
        \includegraphics[width=0.155\textwidth]{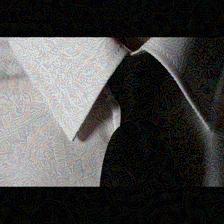} & 
        \includegraphics[width=0.155\textwidth]{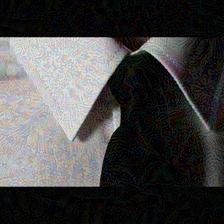} \\
        
        \begin{minipage}[t]{0.155\textwidth}\centering \textcolor{green!50!black}{\textbf{Two fluffy puppies play together in the snow near a metal fence.}}\end{minipage} & 
        \begin{minipage}[t]{0.155\textwidth}\centering \textcolor{gray}{\textbf{Two brown puppies play together on snowy ground near a metal fence in winter.}}\end{minipage} & 
        \begin{minipage}[t]{0.155\textwidth}\centering \textcolor{red}{\textbf{Two realistic baby dinosaur sculptures displayed outdoors behind a metal fence on snowy ground.}}\end{minipage} &
        
        \begin{minipage}[t]{0.155\textwidth}\centering \textcolor{green!50!black}{\textbf{A close-up of a black necktie knotted neatly against a crisp white dress shirt collar.}}\end{minipage} & 
        \begin{minipage}[t]{0.155\textwidth}\centering \textcolor{gray}{\textbf{A close-up of a white dress shirt collar and dark necktie against a patterned background.}}\end{minipage} & 
        \begin{minipage}[t]{0.155\textwidth}\centering \textcolor{red}{\textbf{A dark, grainy image showing a person's silhouette against a patterned light background with geometric shapes.}}\end{minipage} \\
    \end{tabular}
    \vspace{0.4em}
    \hrule
    \vspace{0.8em}
    \begin{tabular}{ccc:ccc}
    
        \textbf{Clean} & \textbf{BSR} & \textbf{LI-Boost-BSR} & \textbf{Clean} & \textbf{BSR} & \textbf{LI-Boost-BSR} \\
        
        \includegraphics[width=0.155\textwidth]{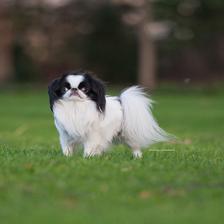} & 
        \includegraphics[width=0.155\textwidth]{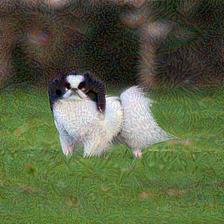} & 
        \includegraphics[width=0.155\textwidth]{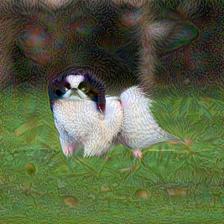} &
        \includegraphics[width=0.155\textwidth]{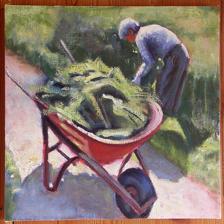} & 
        \includegraphics[width=0.155\textwidth]{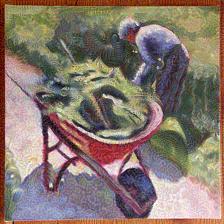} & 
        \includegraphics[width=0.155\textwidth]{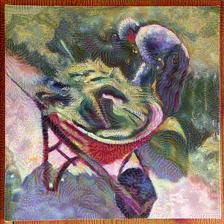} \\
        
        \begin{minipage}[t]{0.155\textwidth}\centering \textcolor{green!50!black}{\textbf{A black and white Japanese Chin dog standing on a lush green lawn.}}\end{minipage} & 
        \begin{minipage}[t]{0.155\textwidth}\centering \textcolor{gray}{\textbf{A small black and white Japanese Chin or similar toy breed dog standing on green grass.}}\end{minipage} & 
        \begin{minipage}[t]{0.155\textwidth}\centering \textcolor{red}{\textbf{A digitally altered image showing a bird with an owl's face walking on grass.}}\end{minipage} &
        
        \begin{minipage}[t]{0.155\textwidth}\centering \textcolor{green!50!black}{\textbf{A painting of a person loading green yard clippings into a red wheelbarrow.}}\end{minipage} & 
        \begin{minipage}[t]{0.155\textwidth}\centering \textcolor{gray}{\textbf{A person in blue clothing loads green plants into a red wheelbarrow in a garden.}}\end{minipage} & 
        \begin{minipage}[t]{0.155\textwidth}\centering \textcolor{red}{\textbf{A colorful, psychedelic artistic rendering of a pig eating watermelon with distorted, surreal imagery.}}\end{minipage} \\
    \end{tabular}

    \vspace{0.4em}
    \hrule
    \vspace{0.8em}

    \begin{tabular}{ccc:ccc}
        \textbf{Clean} & \textbf{ILA} & \textbf{LI-Boost-ILA} & \textbf{Clean} & \textbf{ILA} & \textbf{LI-Boost-ILA} \\
        
        \includegraphics[width=0.155\textwidth]{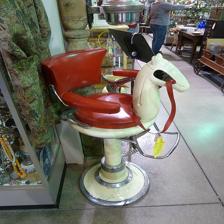} & 
        \includegraphics[width=0.155\textwidth]{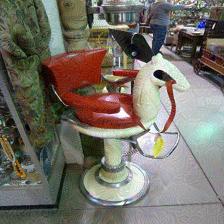} & 
        \includegraphics[width=0.155\textwidth]{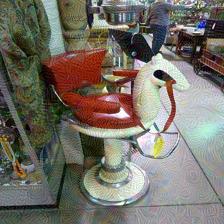} &
        \includegraphics[width=0.155\textwidth]{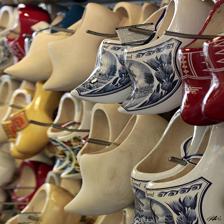} & 
        \includegraphics[width=0.155\textwidth]{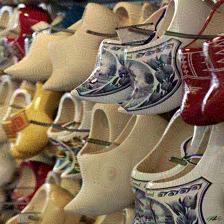} & 
        \includegraphics[width=0.155\textwidth]{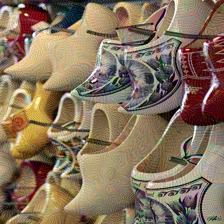} \\
        
        \begin{minipage}[t]{0.155\textwidth}\centering \textcolor{green!50!black}{\textbf{A vintage children's barber chair with a white horse head, featuring red leather upholstery.}}\end{minipage} & 
        \begin{minipage}[t]{0.155\textwidth}\centering \textcolor{gray}{\textbf{A vintage children's barber chair shaped like a horse in a cluttered antique shop.}}\end{minipage} & 
        \begin{minipage}[t]{0.155\textwidth}\centering \textcolor{red}{\textbf{A decorative cobra-shaped hookah pipe with red and white coloring displayed in a shop.}}\end{minipage} &
        
        \begin{minipage}[t]{0.155\textwidth}\centering \textcolor{green!50!black}{\textbf{Colorful Dutch wooden clogs displayed in rows on a wall, featuring traditional painted designs.}}\end{minipage} & 
        \begin{minipage}[t]{0.155\textwidth}\centering \textcolor{gray}{\textbf{Colorful traditional Dutch wooden clogs (klompen) displayed in rows, featuring various painted designs.}}\end{minipage} & 
        \begin{minipage}[t]{0.155\textwidth}\centering \textcolor{red}{\textbf{A surreal, distorted collection of colorful shoes and sneakers blended together in a dreamlike, AI-generated image.}}\end{minipage} \\
    \end{tabular}

    \vspace{0.4em}
    \hrule
    \vspace{0.8em}

    \begin{tabular}{ccc:ccc}
        \textbf{Clean} & \textbf{BPA} & \textbf{LI-Boost-BPA} & \textbf{Clean} & \textbf{BPA} & \textbf{LI-Boost-BPA} \\
        
        \includegraphics[width=0.155\textwidth]{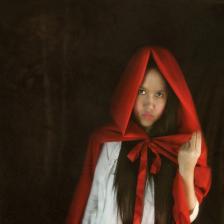} & 
        \includegraphics[width=0.155\textwidth]{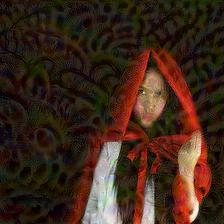} & 
        \includegraphics[width=0.155\textwidth]{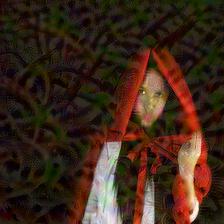} &
        \includegraphics[width=0.155\textwidth]{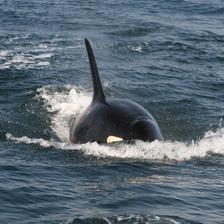} & 
        \includegraphics[width=0.155\textwidth]{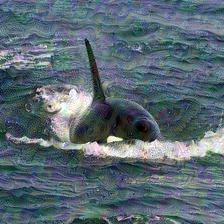} & 
        \includegraphics[width=0.155\textwidth]{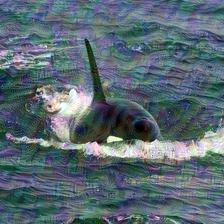} \\
        
        \begin{minipage}[t]{0.155\textwidth}\centering \textcolor{green!50!black}{\textbf{A young girl dressed as Little Red Riding Hood wears a red hooded cape, looking serious.}}\end{minipage} & 
        \begin{minipage}[t]{0.155\textwidth}\centering \textcolor{gray}{\textbf{A girl wearing a red hooded cloak against a psychedelic, swirling patterned background.}}\end{minipage} & 
        \begin{minipage}[t]{0.155\textwidth}\centering \textcolor{red}{\textbf{A person's face is partially visible through an abstract, psychedelic overlay of red swirling patterns.}}\end{minipage} &
        
        \begin{minipage}[t]{0.155\textwidth}\centering \textcolor{green!50!black}{\textbf{An orca (killer whale) surfaces in the ocean, its dorsal fin and back visible above water.}}\end{minipage} & 
        \begin{minipage}[t]{0.155\textwidth}\centering \textcolor{gray}{\textbf{An orca (killer whale) swimming near the surface of the ocean, showing its dorsal fin.}}\end{minipage} & 
        \begin{minipage}[t]{0.155\textwidth}\centering \textcolor{red}{\textbf{A seal swimming in greenish-purple water, with its head and body partially visible above the surface.}}\end{minipage} \\
    \end{tabular}

    \caption{Visualization of clean images and their adversarial counterparts across different attack methods against Claude-o, and the surrogate model is RN-50. \textbf{\textcolor{green!50!black}{Green text}} denotes the ground truth captions. \textbf{\textcolor{gray}{Gray text}} denotes the results of failed adversarial examples. \textbf{\textcolor{red}{Red text}} highlights the significant semantic deviation achieved by our LI-Boosted methods.}
    \label{fig:claude4opus}
\end{figure*}

Furthermore, Fig.~\ref{fig:claude4opus} provides a compelling qualitative demonstration of how LI-Boost transcends simple feature perturbation to profoundly degrade the high-level semantic reasoning capabilities of Claude4-o. The adversarial examples generated by baseline attacks without LI-Boost (\eg, ILA, BPA) exhibit only marginal semantic deviations. In these cases, the victim VLM largely retains the correct semantic understanding of the images, committing primarily minor attribute or background errors. Conversely, methods incorporating LI-Boost induces drastic semantic shifts and severe misalignment. For instance, LI-Boost-ILA successfully triggers a complete hallucination, causing the target model to misunderstand a static children's barber chair as a decorative cobra-shaped hookah pipe, and LI-Boost-BPA describes orca as a seal. These results demonstrate that LI-Boost not only induces severe misclassifications but also fundamentally disrupts the semantic reasoning ability of the victim VLM, thereby substantiating its superiority in generating highly transferable adversarial examples and exposing the fragility of semantic alignment in modern VLMs.



\vfill

\end{document}